\pdfoutput=1
\PassOptionsToPackage{table}{xcolor}

\documentclass[11pt]{article}

\usepackage[final]{acl}

\usepackage{times}
\usepackage{latexsym}

\usepackage[T1]{fontenc}
\usepackage[utf8]{inputenc}

\usepackage{microtype}

\usepackage{inconsolata}

\usepackage{graphicx}
\usepackage{latexsym}
\usepackage{caption}
\usepackage[font=small,labelfont=bf]{subcaption}
\usepackage{booktabs}
\usepackage{enumitem}
\usepackage{multirow}
\usepackage{color}
\usepackage{hyperref}
\usepackage{amsmath}
\usepackage{amsfonts}
\usepackage{amsthm}
\usepackage{scalefnt}
\usepackage{xspace}
\usepackage{fontawesome}
\usepackage{tcolorbox}
\usepackage{rotating}
\usepackage{subcaption}
\usepackage{dashrule}

\usepackage{pgfplots}
\usepackage{tikz}
\pgfplotsset{compat=1.17}
\usepackage{afterpage}

\usepackage{cleveref}
\crefname{section}{\S\!}{\S\S\!}
\crefname{table}{Tab.}{Tabs.}
\crefname{figure}{Fig.}{Figs.}
\crefname{algorithm}{Alg.}{Algs.}
\crefname{appendix}{App.}{Apps.}
\crefname{equation}{Eq.}{Eqs.}
\crefname{lemma}{Lemma}{}
\Crefname{theorem}{Theorem}{}
\crefname{proposition}{Proposition}{}
\crefname{hypothesis}{Hypothesis}{}
\crefname{deduction}{Deduction}{}
\crefname{intuition}{\textbf{Intuition}}{\textbf{Intuitions}}
\crefname{observation}{\textbf{Observation}}{\textbf{Observations}}
\crefname{finding}{\textbf{Finding}}{\textbf{Findings}}
\crefname{cor}{Corollary}{Corollaries}

\usepackage{xspace}
\newcommand{\modelfontstyle}{\textsf}
\newcommand{\modellogo}[3][0.8em]{%
    \raisebox{-0.1em}{%
        \includegraphics[height=#1]{#2}%
    }%
    \,%
    {\modelfontstyle{#3}}\xspace%
}
\newcommand{\palm}[1][0.8em]{%
    \modellogo[#1]{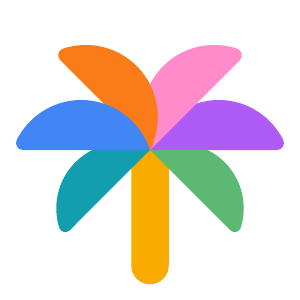}{PaLM}%
}
\newcommand{\llamaTwo}[1][0.8em]{%
    \modellogo[#1]{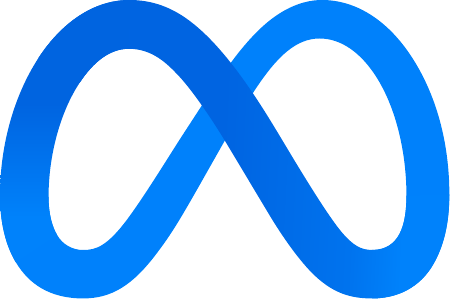}{Llama2}%
}
\newcommand{\llamaThree}[1][0.8em]{%
    \modellogo[#1]{logos/meta.pdf}{Llama3-8B-Instruct}%
}
\newcommand{\llamaThreeOne}[1][0.8em]{%
    \modellogo[#1]{logos/meta.pdf}{Llama3.1-8B-Instruct}%
}
\newcommand{\qwenTwo}[1][0.9em]{%
    \modellogo[#1]{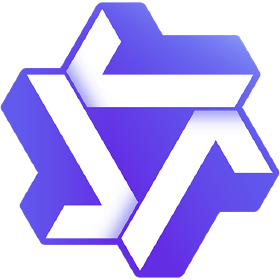}{Qwen2-7B-Instruct}%
}
\newcommand{\qwenTwoFive}[1][0.9em]{%
    \modellogo[#1]{logos/qwen.png}{Qwen2.5-7B-Instruct}%
}
\newcommand{\mistral}[1][0.8em]{%
    \modellogo[#1]{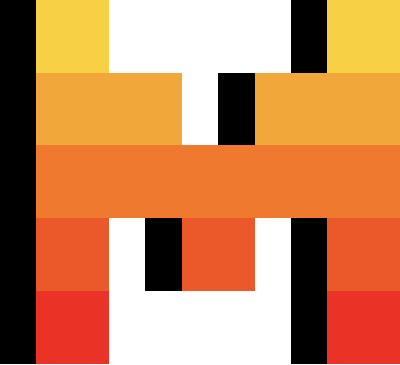}{NeMo-12B-Instruct}%
}
\newcommand{\aya}[1][0.9em]{%
    \modellogo[#1]{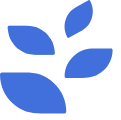}{Aya-Expanse-8B}%
}
\newcommand{\gptThreeFive}[1][0.9em]{%
    \modellogo[#1]{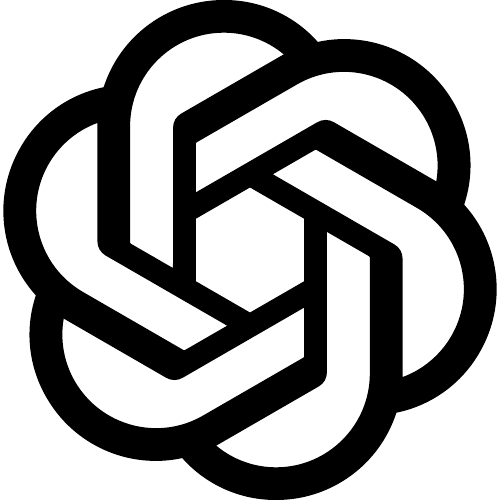}{GPT3.5-turbo}%
}

\newcommand{\gptFourO}[1][0.9em]{%
    \modellogo[#1]{logos/gpt.pdf}{GPT4o-mini}%
}

\newcommand{\xllamaThree}[1][0.8em]{%
    \modellogo[#1]{logos/meta.pdf}{3}%
}
\newcommand{\xllamaThreeOne}[1][0.8em]{%
    \modellogo[#1]{logos/meta.pdf}{3.1}%
}
\newcommand{\xqwenTwo}[1][0.9em]{%
    \modellogo[#1]{logos/qwen.png}{2}%
}
\newcommand{\xqwenTwoFive}[1][0.9em]{%
    \modellogo[#1]{logos/qwen.png}{2.5}%
}
\newcommand{\xmistral}[1][0.8em]{%
    \modellogo[#1]{logos/mistral.pdf}{}%
}
\newcommand{\xaya}[1][0.9em]{%
    \modellogo[#1]{logos/aya.png}{}%
}
\newcommand{\xgptThreeFive}[1][0.9em]{%
    \modellogo[#1]{logos/gpt.pdf}{3.5}%
}

\newcommand{\xgptFourO}[1][0.9em]{%
    \modellogo[#1]{logos/gpt.pdf}{4om}%
}

\newcommand{\worldemoji}[3][1em]{%
    \raisebox{-0.3em}{%
        \includegraphics[height=#1]{#2}%
    }%
    \,%
    {\modelfontstyle{#3}}%
}
\newcommand{\world}[1][1.2em]{%
    \worldemoji[#1]{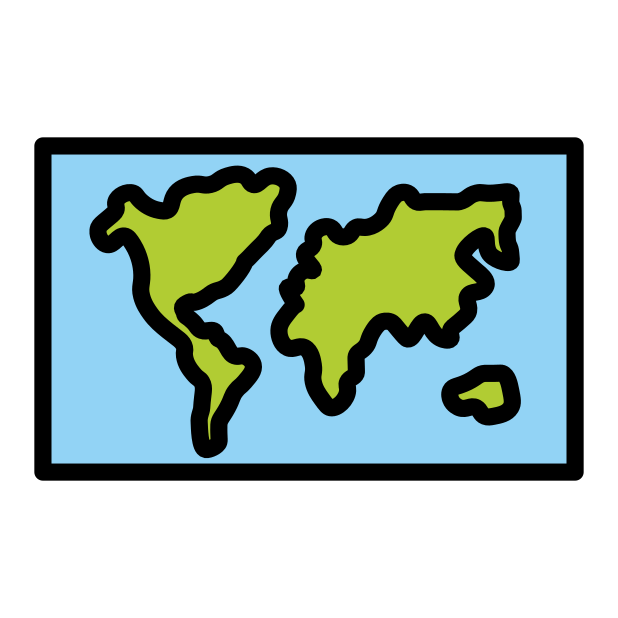}{}%
}

\usepackage[table]{xcolor}
\definecolor{tablerowcolor}{gray}{0.9}

\newcommand{\alternaterowcolors}[1][2]{
    \def\startrow{#1}  % The starting row is set by the optional parameter (default is 2)

    \rowcolors{\startrow}{tablerowcolor}{white}
}

\definecolor{ForestGreen}{HTML}{009B55}
\definecolor{OrangeRed}{HTML}{ED135A}
\definecolor{CadetBlue}{HTML}{74729A}
\definecolor{SkyBlue}{HTML}{46C5DD}
\definecolor{SeaGreen}{HTML}{3FBC9D}
\definecolor{Peach}{HTML}{F7965A}
\definecolor{Goldenrod}{HTML}{FFDF42}
\definecolor{NavyBlue}{HTML}{006EB8}
\definecolor{Periwinkle}{HTML}{7977B8}
\definecolor{Orchid}{HTML}{AF72B0}
\definecolor{BlueViolet}{HTML}{473992}
\definecolor{SpringGreen}{HTML}{C6DC67}
\definecolor{ETHGray}{RGB}{0, 94, 184}	% gray

\newcommand{\mgsm}{\textsf{MGSM}\xspace}
\newcommand{\xcopa}{\textsf{XCOPA}\xspace}
\newcommand{\xlwic}{\textsf{XL-WiC}\xspace}
\newcommand{\flores}{\textsf{FLORES-101}\xspace}
\newcommand{\combined}{\textsf{COMBINED}\xspace}
\newcommand{\xquad}{\textsf{XQuAD}\xspace}

\newcommand{\english}{\texttt{English}\xspace}
\newcommand{\multilingual}{\texttt{Multilingual}\xspace}

\newcommand{\native}{\texttt{Native}\xspace}
\newcommand{\monolingual}{\texttt{Monolingual}\xspace}
\newcommand{\chinese}{\texttt{Chinese}\xspace}
\newcommand{\japanese}{\texttt{Japanese}\xspace}
\newcommand{\french}{\texttt{French}\xspace}
\newcommand{\italian}{\texttt{Italian}\xspace}
\newcommand{\german}{\texttt{German}\xspace}

\newcommand{\transEn}{\texttt{Transl-Lang$\rightarrow$En}\xspace}
\newcommand{\transSource}{\texttt{Transl-En$\rightarrow$Lang}\xspace}

\newcommand{\cis}{CIS\xspace}
\newcommand{\cisEn}{\texttt{CIS-En}\xspace}
\newcommand{\cisZh}{\texttt{CIS-Zh}\xspace}
\newcommand{\cisFr}{\texttt{CIS-Fr}\xspace}
\newcommand{\cisJa}{\texttt{CIS-Ja}\xspace}
\newcommand{\cisMulti}{\texttt{CIS-Multi}\xspace}
\newcommand{\cisMultilingual}{\texttt{CIS-Multilingual}\xspace}
\newcommand{\cisWorld}{\texttt{CIS-}\world\xspace}

\usepackage{collcell,fp}
\usepackage{xcolor,colortbl}
\usepackage{pgfmath}
\newcommand{\dynamicCellColor}[1]{
    \pgfmathsetmacro{\colorValue}{(#1-55)/45} % Scale 59-100 to 0-1 for interpolation
    \pgfmathsetmacro{\redValue}{int(255 * \colorValue + 0 * (1 - \colorValue))} % Interpolated red component
    \pgfmathsetmacro{\greenValue}{60} % Interpolated green component
    \pgfmathsetmacro{\blueValue}{int(0 * (1 - \colorValue) + 255 * (1 - \colorValue))} % Interpolated blue component
    \definecolor{dynamiccolor}{RGB}{\redValue, \greenValue, \blueValue} % Define dynamic color using RGB
    \colorbox{dynamiccolor!30}{#1} % Adjust transparency
}

\newcommand{\vect}[1]{\boldsymbol{\mathbf{#1}}}
\DeclareMathOperator{\EOT}{\texttt{EOT}}
\newcommand{\user}{\texttt{user}\xspace}
\newcommand{\assistant}{\texttt{assistant}\xspace}

\usepackage{xparse}
\NewDocumentCommand{\increase}{m m o}{%
    \IfValueTF{#3}{%
        ${#1}_{\textcolor{ForestGreen}{#2\uparrow}}^{\boldsymbol{#3}}$%
    }{%
        $#1_{\textcolor{ForestGreen}{#2\uparrow}}$
    }%
}%
\NewDocumentCommand{\decrease}{m m o}{%
    \IfValueTF{#3}{%
        ${#1}_{\textcolor{OrangeRed}{#2\downarrow}}^{\boldsymbol{#3}}$%
    }{%
        $#1_{\textcolor{OrangeRed}{#2\downarrow}}$%
    }%
}%

\newcommand{\interalia}[1]{\citep[\textit{inter alia}]{#1}}

\usepackage{todonotes}
\usepackage{multirow}
\usepackage{array}
\usepackage{tcolorbox}
\definecolor{tticblue}{RGB}{0, 94, 184}  % a more visually pleasing blue

\title{Blessing of Multilinguality: A Systematic Analysis of Multilingual In-Context Learning}
\author{
Yilei Tu\textsuperscript{1} \quad\quad\quad\quad
Andrew Xue\textsuperscript{1} \quad\quad\quad\quad
Freda Shi\textsuperscript{1,2,3} \\
\textsuperscript{1}David R. Cheriton School of Computer Science, University of Waterloo \\
\textsuperscript{2}Vector Institute \\
\textsuperscript{3}Canada CIFAR AI Chair \\
\texttt{\href{mailto:yileitu.tt@gmail.com}{yileitu.tt@gmail.com}, \href{mailto:fhs@uwaterloo.ca}{fhs@uwaterloo.ca}}
}

\begin{document}
\maketitle
\begin{abstract}
    While multilingual large language models generally perform adequately, and sometimes even rival English performance on high-resource languages (HRLs), they often significantly underperform on low-resource languages \citep[LRLs; ][]{nllb}.
    Among several prompting strategies aiming at bridging the gap, multilingual in-context learning \citep[ICL; ][]{mgsm} has been particularly effective when demonstration in target languages is unavailable.
    However, there lacks a systematic understanding of when and why it works well.

    In this work, we systematically analyze multilingual ICL, using demonstrations in HRLs to enhance cross-lingual transfer.
    We show that demonstrations in mixed HRLs consistently outperform English-only ones across the board, particularly for tasks written in LRLs.
    Surprisingly, our ablation study shows that the presence of irrelevant non-English sentences in the prompt yields measurable gains, suggesting the effectiveness of multilingual exposure itself.
    Our results highlight the potential of strategically leveraging multilingual resources to bridge the performance gap for underrepresented languages.
    
    \faGithub \quad \url{https://github.com/yileitu/multilingual_icl}
\end{abstract}

\section{Introduction}

\begin{figure}[!t]
    \centering
    \begin{subfigure}[b]{\columnwidth} 
        \centering
        \includegraphics[width=\linewidth]{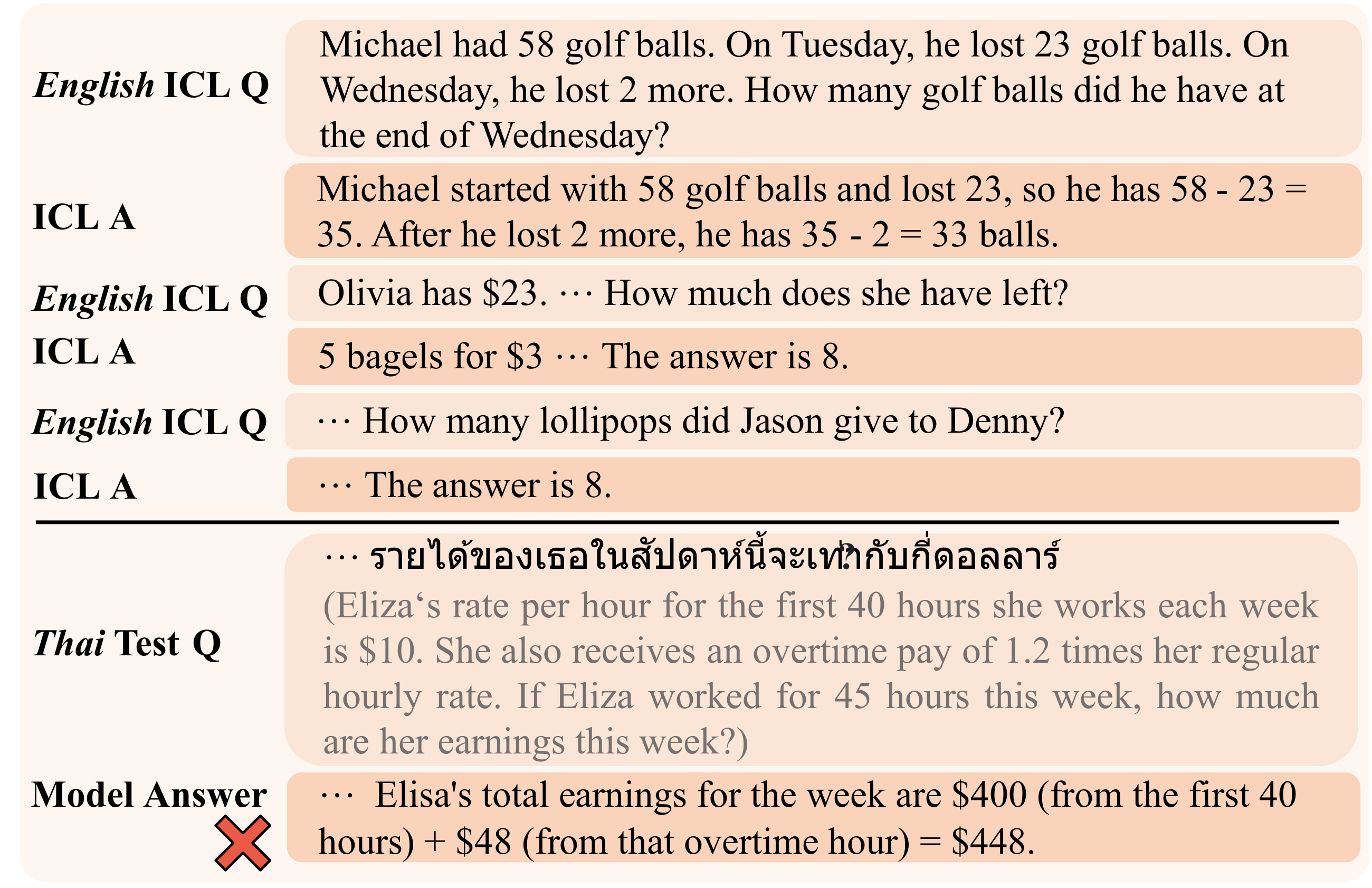}
        \captionsetup{skip=3pt}
        \caption{\english ICL Mode}
        \label{fig:pull:english}
    \end{subfigure}

    \vspace{0.0cm} 
    \begin{subfigure}[b]{\columnwidth}
        \centering
        \includegraphics[width=\linewidth]{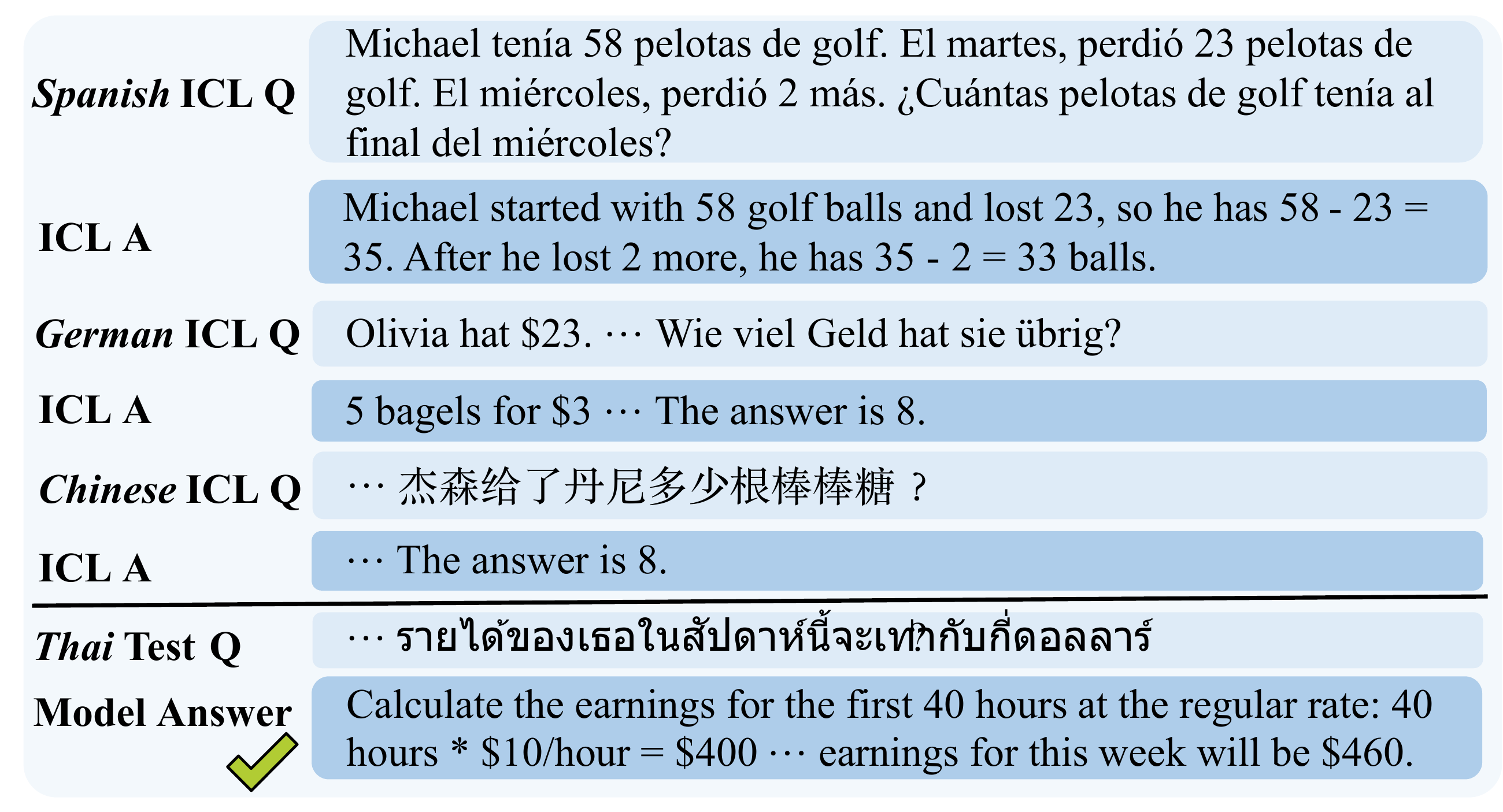}
        \captionsetup{skip=3pt}
        \caption{\multilingual ICL Mode}
        \label{fig:pull:multilingual}
    \end{subfigure}

    \caption{
        Illustration of two ICL modes. After providing a few-shot prompt, we evaluate LLM in the same domain in various languages. In a controlled experiment, each demonstration in (a) and (b) shares the same meaning, albeit in different languages. Contents and languages of demonstrations are randomly sampled from a training set and a preset high-resource language list, respectively. We find that the \multilingual ICL mode (b) is more effective in helping the LLM solve tasks in different languages compared to the \english ICL mode (a).
        \label{fig:pull}
    }
    \vspace{-5pt}
\end{figure}

\noindent In-context learning \citep[ICL;][]{lm_are_few_shot_learner} has become a widely adopted technique in natural language processing with large language models \cite[LLMs;][\textit{inter alia}]{llama2,palm,llama3,qwen2,qwen2.5}, which enables LLMs to learn to solve problems by analogy from a few input-output examples (i.e., demonstrations) without updating model parameters.
As a generic method, ICL has also been effective in improving the cross-lingual performance of multilingual LLMs \citep[MLLMs; ][]{lm_few_shot_multilingual_learner,mega,mgsm,few_shot_multilingual_nlu,buffet}.

Prior work has introduced two common ICL strategies for non-English languages: (i) translating the target question into English and performing English-only ICL \citep[\textit{inter alia}]{mgsm,mega,few_shot_multilingual_nlu}, and (ii) providing demonstrations in the target language \cite[\textit{in-language demonstrations;}][]{polyglot_prompt,cross_lingual_prompting,not_all_language_are_created_equal,plug}.
Both strategies have critical shortcomings: (i) may suffer from information loss in translation due to nuanced language gap \citep{dont_trust_when_question_not_in_english,roles_of_english} or the unavailability of high-quality translation systems for extremely low-resource languages (LRLs), whereas (ii) may become infeasible due to data scarcity in LRLs.
As alternatives, when presenting LLMs with problems in the target language, English demonstrations (\cref{fig:pull:english}) lead to poor performance on LRLs, whereas demonstrations in multiple high-resource languages (HRLs; \cref{fig:pull:multilingual}) can be more effective, even when there is little alphabetical overlap between the demonstration and target languages \citep{mgsm}; however, the underlying reasons on why it works remain unclear.

In this work, we systematically analyze multilingual ICL through a set of controlled experiments.
Each test question is paired with a set of semantically equivalent demonstrations, while the demonstration languages vary according to different ICL modes (\cref{sec:setup:prompts}).
We compare the performance differences across four ICL modes (\cref{sec:icl_mode_eval:multilingual_vs_english,sec:icl_mode_eval:monolingual}): English, individual-HRL, multilingual (i.e., mixed HRLs), and in-language demonstrations (\cref{fig:icl_modes}).
To disentangle the impact of demonstration language from other confounding factors (e.g., interactions between demonstration languages and in-domain demonstrations), we conduct additional control experiments by adding irrelevant sentences in various languages into English-only in-domain demonstrations (\cref{sec:icl_mode_eval:cis}).
% \freda{Yilei, could you cite each subsection here accordingly?}
We find that
\begin{itemize}[leftmargin=*,topsep=0pt,itemsep=-4pt]
    \item In-context demonstrations in HRLs, especially in languages with non-Latin writing systems such as Chinese and Japanese, can more effectively transfer knowledge compared to English ICL mode, leading to performance improvement in answering questions in all languages, especially in LRLs.
          This finding is generalizable enough across different domains and various LLMs.
    \item Demonstrations in mixed HRLs are more robust and effective compared to those in a single HRL, in terms of average accuracy boosting on different tasks. This strategy is favored.
    \item Surprisingly, simply introducing another non-English language (not necessarily in the demonstration) in the prompt could lead to performance improvement, albeit the improvement is not as significant as the aforementioned strategies.
\end{itemize}
\section{Background: Multilingual ICL}
\label{sec:multiling_icl}
In this section, we review the basic approaches of ICL with instruction-tuned LLMs (\cref{sec:multiling_icl:icl_instruction_tuning}) and introduce the multilingual prompting modes (\cref{sec:multiling_icl:modes}) that we evaluate in this work.
\subsection{ICL for Instruction-Tuned LLM} \label{sec:multiling_icl:icl_instruction_tuning}
Instruction-tuned LLMs \citep{instructGPT,cross_task_generalization, super_natural_instructions, ft_lm_are_zs_learners} generally possess the capability to follow task instructions (i.e., \textit{system prompt}), which are usually coupled with ICL to fully elicit their capability \interalia{cot}.
Formally, denote a training set by $\mathcal{D}_\text{train}=\left\{\left(q_i, a_i\right)\right\}_{i=1}^M$ and a test set $\mathcal{D}_\text{test}=\left\{\left(q_j, a_j\right)\right\}_{j=1}^{N}$ in the same domain, where $q_i$ is a task question (as model \textit{input}).
An ICL prompt for a test question $q_{\text{test}} \in \mathcal{D}_\text{test}$ has three core components:
(1) a system prompt $I_\text{sys}$ that describes the task and specifies the expected output format,
(2) $K$ sample input-output pairs ($K$-shot) from the training set $\left\{\left(q_k, a_k\right)\right\}_{k=1}^K \sim \mathcal{D}_\text{train}$ that provide in-context demonstrations, and
(3) a verbalizer $V$ mapping each ground truth label $a_i$ to a textual representation, which may also include reasoning steps \citep[i.e., chains of thoughts, or CoT in short;][]{cot}.
In summary, an ICL prompt for $q_{\text{test}}$ can be written as:
\begin{align}
  \text{prompt}_{q_\text{test}} & = I_\text{sys} \circ q_1 \circ V\left(a_1\right) \circ q_2 \circ V\left(a_2\right) \nonumber \\
                                & \quad \circ \cdots \circ q_K \circ V\left(a_K\right) \circ q_{\text{test}},
  \label{eq:icl_construction}
\end{align}
where $\circ$ is the string concatenation operator with a special end-of-turn $\left(\EOT\right)$ token as the delimiter.
The LLM with parameters $\vect{\theta}$, denoted as $p_{\vect{\theta}}$, then generates the response $\hat{a}_\text{test}$ given $\text{prompt}_{q_\text{test}}$:
$\hat{a}_\text{test} \sim p_{\vect{\theta}}\left(\text{prompt}_{q_\text{test}}\right)$.

\subsection{Multilingual Prompting Modes} \label{sec:multiling_icl:modes}
We extend our notations as follows to adapt to the multilingual settings.
A training set with $L$ languages is denoted by $\mathcal{D}_\text{train} = \left\{\mathcal{D}_\text{train}^{\text{lang}_1}, \ldots, \mathcal{D}_\text{train}^{\text{lang}_L} \right\}$, % e.g.,
% French split $\mathcal{D}_\text{train}^{\text{fr}}=\left\{\left(q_i^{\text{fr}}, a_i\right)\right\}_{i=1}^M$.
where the split $\mathcal{D}_\text{train}^{\text{lang}_\ell}$ consists of $M$ examples for any $\ell\in \{1, \ldots, L\}$.
The same applies to the test dataset $\mathcal{D}_\text{test}$, with each language-specific split consisting of $N$ examples.
Without further specification, we assume that the training examples at the same index are semantically equivalent across languages.

The ground-truth labels $a_i$ in quantitative LLM benchmarks \interalia{xcopa,gsm8k} are typically language-agnostic (such as numbers) or represented in a single word (such as Yes/No).
In such cases, the verbalizer is an identity.
For answers requiring reasoning steps, $V(a_i)$ is CoT in English, as there has been strong evidence that MLLMs perform better when generating English \interalia{mgsm, cross_lingual_prompting, not_all_language_are_created_equal}.
For the same reasons, $I_\text{sys}$ is always presented in English as well.

Following \citet{mega} and \citet{mgsm}, we evaluate MLLMs via several different prompting strategies (\textit{ICL modes}) in this work:

\begin{figure}[t]
  \centering
  \includegraphics[width=\linewidth]{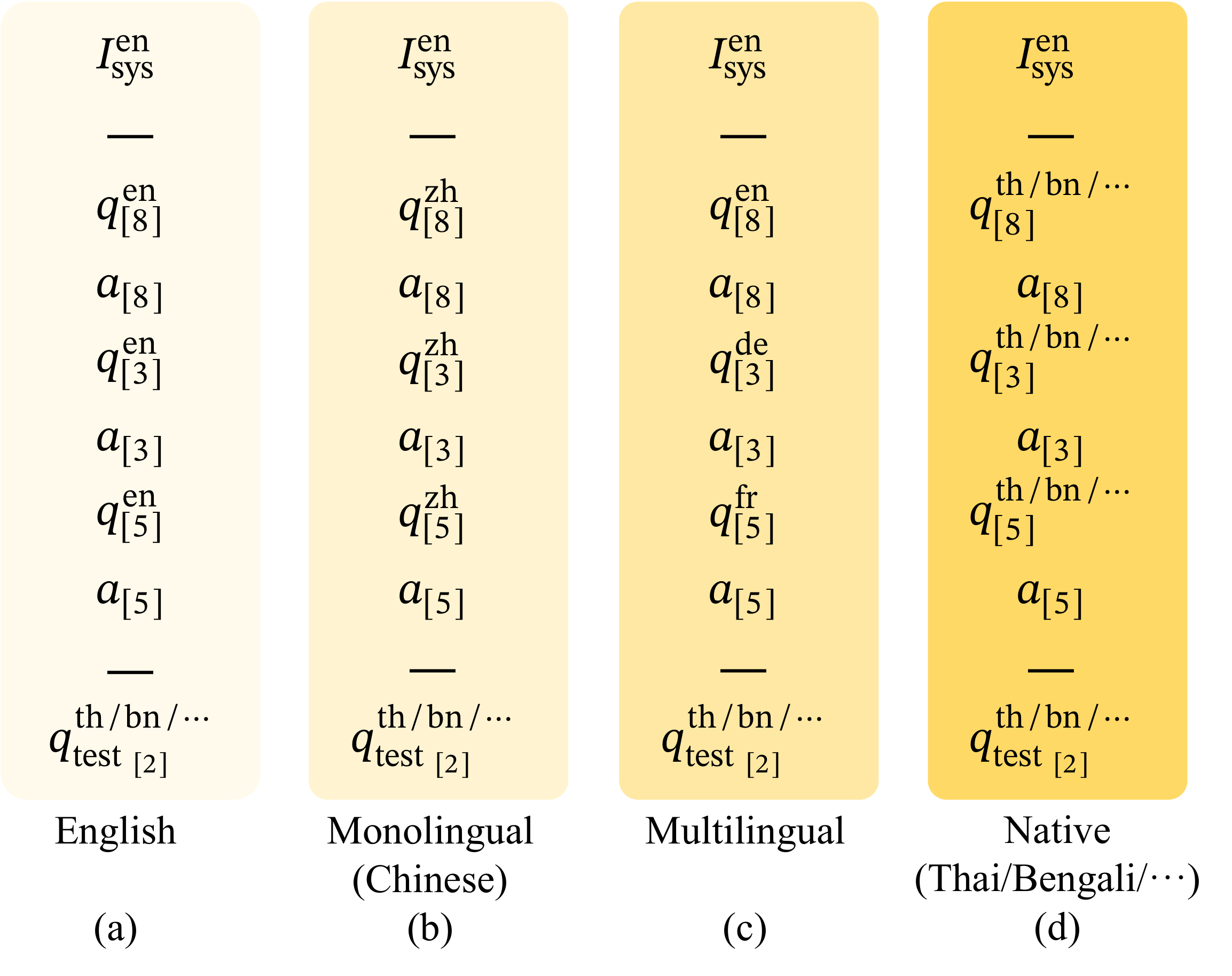}
  \caption{Illustration of ICL modes by \cref{eq:icl_construction}. Assume $K=3$ and $M=10$. For the second datapoint of the test set (regardless of its language split, e.g., $q_\text{test}$ could be in Thai, Bengali, etc.), we first randomly generate $K=3$ indices from $\left\{1, \cdots, 10\right\}$, say $\left\{8,3,5\right\}$.  Next, we determine the languages of the $K=3$ demonstrations. For modes (a), (b), and (d), the language is uniformly specified. For mode (c), we randomly select $K=3$ languages, say $\left\{\text{en, de, fr}\right\}$. Then $\left\{(8, \text{en}), (3, \text{de}), (5, \text{fr})\right\}$ determines each demonstration.
  }
  \vspace{-5pt}
  \label{fig:icl_modes}
\end{figure}

\vspace{2pt}\noindent\textbf{The \english mode.} The $K$ demonstrations are always in English (\cref{fig:pull:english,fig:icl_modes}a).

\vspace{2pt}\noindent\textbf{The \monolingual mode(s).} The $K$ demonstrations are always presented in a single non-English high-resource language such as Chinese (\cref{fig:icl_modes}b).

\vspace{2pt}\noindent\textbf{The \multilingual mode.} \label{sec:icl:multilingual_mode}
From a predefined list of high-resources languages $\mathcal{L}_H$, $K$ languages are randomly selected, which, together with the sampled $K$ indices, determine the contents and languages of the $K$ demonstrations (\cref{fig:pull:multilingual,fig:icl_modes}c).

\vspace{2pt}\noindent\textbf{The \native mode.} The $K$ demonstrations are in the same language as the test question (\cref{fig:icl_modes}d).

\section{Experiment Setups}
\label{sec:setup}
% \subsection{Model}
\noindent\textbf{Models}.
We evaluate state-of-the-art instruction-tuned LLMs with about 8 billion parameters, which have officially claimed multilingual capabilities in model release: \textsf{Llama3-8B-Instruct}, \textsf{Llama3.1-8B-Instruct} \citep{llama3}, \textsf{Qwen2-7B-Instruct} \citep{qwen2}, \textsf{Qwen2.5-7B-Instruct} \citep{qwen2.5}, \textsf{Mistral-NeMo-12B-Instruct} \citep{mistral} and \textsf{Aya-Expanse-8B} \citep{aya-expanse}.
For additional references, we evaluate OpenAI closed-sourced commercial models, including \textsf{GPT3.5-turbo} \citep{gpt3.5} and \textsf{GPT4o-mini} \citep{gpt4o-mini}.
Detailed model cards can be found in \cref{app:model}.

\begin{table*}[!t]
    %\resizebox{\columnwidth}{!}{
    \small
    \centering
    % \alternaterowcolors
    \setlength{\tabcolsep}{3.2pt}
    \begin{tabular}{llccrrc}
        \toprule
        \bfseries Dataset                              & \bfseries Domain       & \bfseries Expected          & \bfseries  \#Languages            & \bfseries  Language Split Size    & \bfseries En Avg Word & \bfseries \multirow{2}{*}{Parallel} \\
        \multicolumn{2}{l}{\textit{Datapoint Example}} & \bfseries Output       & \bfseries (\#HRL $+$ \#LRL) & \bfseries $M/N$ --- Training/Test & \bfseries Count$_{\pm\text{std}}$                                                               \\
        \midrule
        \midrule
        \mgsm                                          & Mathematical Reasoning & Numerals                    & $11 \ (7+4)$                      & $8/250$                           & $46.26_{\pm 18.29}$   & \faCheck                            \\
        \multicolumn{7}{l}{\textit{See \cref{fig:pull} for examples.}}                                                                                                                                                                              \\
        \midrule

        \xcopa                                         & Commonsense Reasoning  & ``1'' / ``2''               & $12 \ (5+7)$                      & $100/500$                         & $26.59_{\pm 3.41}$    & \faCheck                            \\
        \multicolumn{7}{l}{\textit{\textcolor{ETHGray}{Premise:} The man turned on the faucet. \quad \textcolor{ETHGray}{What happened as a RESULT?}}}                                                                                                    \\
        \multicolumn{7}{l}{\textit{\textcolor{ETHGray}{Hypothesis 1:} The toilet filled with water. \quad \textcolor{ETHGray}{Hypothesis 2:} Water flowed from the spout. \quad \textcolor{ETHGray}{Answer:} 2. }}                                           \\
        \midrule

        \xlwic                                         & Word Disambiguation    & ``Yes'' / ``No''            & $13 \ (9+4)$                      & $98/390$                          & $35.65_{\pm 5.36}$    & \faTimes                            \\
        \multicolumn{7}{l}{\textit{\textcolor{ETHGray}{Sentence 1:} What did you *get* at the toy store? \quad \textcolor{ETHGray}{Sentence 2:} She didn't *get* his name when they met the first time.}}                                                 \\
        \multicolumn{7}{l}{\textit{\textcolor{ETHGray}{Question: Is the word ``}get\textcolor{ETHGray}{'' (marked with *) used in the same way in both sentences above?} \quad \textcolor{ETHGray}{Answer:} No.}}                                            \\

        \midrule
        
        \xquad                                         & Extractive QA   & Text            & $12 \ (8+4)$                      & $190/1000$                          & $137.84_{\pm 59.35}$    & \faCheck                            \\

        \multicolumn{7}{l}{\textit{\textcolor{ETHGray}{Passage:} The Panthers defense gave up just 308 points, ranking sixth in the league, $\cdots$}}                                                \\
        \multicolumn{7}{l}{\textit{\textcolor{ETHGray}{Question:} How many points did the Panthers defense surrender? \quad \textcolor{ETHGray}{Answer:} 308.}}                                           \\
        \bottomrule
    \end{tabular}
    %}
    \caption{Dataset properties and examples. Each language split (for both training and test) is of the same size. In-context demonstrations are randomly drawn from the training dataset. Data source and languages are documented in \cref{tab:dataset_additional_info} in \cref{app:dataset}. \textcolor{ETHGray}{Texts in blue} represent \textit{interfaces} acting like reserved-words in programming languages \cite{roles_of_english}. The language of these interfaces changes accordingly with the input source language. ``En Avg Word Count$_{\pm\text{std}}$'' represents the average word count and standard deviation of the demonstration questions of the English split.}
    \vspace{-5pt}
    \label{tab:dataset}
\end{table*}

\vspace{3pt}
\noindent\textbf{Datasets.} \label{sec:setup:datasets}
We evaluate the models using multilingual benchmarks from $4$ distinct domains:
(1) \mgsm \cite{mgsm}, a benchmark of $250$ grade-school math problems sampled from the English GSM8K \citep{gsm8k} and translated into 10 additional languages by expert native speakers.
(2) \xcopa \cite{xcopa}, a commonsense reasoning benchmark that extends the COPA dataset in English \citep{copa} to $11$ additional languages.
% \footnote{In this work, we merge the English COPA and XCOPA datasets, which we still refer to as \xcopa.}  
(3) \xlwic \cite{xlwic}, a cross-lingual word-in-context understanding dataset spanning $13$ languages, where models are expected to tell whether a polysemous word retains the same meaning in two contexts.
(4) \xquad \cite{xquad} is designed to evaluate cross-lingual question answering performance, based on the English SQuAD \cite{squad} dataset and professionally translated into 10 languages.
% (5) \xlsum \cite{xlsum} is for evaluating cross-lingual abstractive summarization, consisting of professionally curated news article-summary pairs across $45$ languages.

\mgsm, \xcopa and \xquad are \textit{parallel} where each corresponding datapoint across different language splits contains semantically equivalent content, allowing us to minimize semantic confounders in our experimental design.  \xlwic is language-specific and translation-variant thus naturally non-parallel. Dataset properties and examples are in \cref{tab:dataset,tab:dataset_additional_info}. Details of data curation are in \cref{app:dataset}.

\vspace{2pt}\noindent\textbf{Languages.}
Languages with large-scale digitized data resources on the web are known as \textit{high-resource languages} \citep[HRLs;][]{bender_rule}, which are exemplified by English, Spanish, and Chinese, among others.
In contrast, \textit{low-resource languages} (LRL) have scarce accessible data \cite{nllb}.
However, a universal standard for dichotomizing languages as either high- or low-resource has not been set \citep{bender_rule,state_and_fate_of_linguistic_diversity,survey_low_resource_nlp}.
Moreover, none of the models we evaluate has disclosed the language distributions in their training corpora.
As a workaround, we define our preset HRL list as the union of the 20 most frequent languages in \textsf{Llama2} \cite{llama2} and \textsf{PaLM} \cite{palm}, and classify languages out of the HRL list as LRLs.
Details of the preset HRL list can be found in \cref{app:lang}.

\vspace{2pt}\noindent\textbf{Prompts.} \label{sec:setup:prompts}
Following \citet{mgsm}, we use $K=6$ examples for demonstration for any test question.
For each multilingual dataset (\cref{tab:dataset}), we first sample $N$ index lists of length $K=6$ all at once, where the index range is $\left\{1, 2, \cdots, M\right\}.$
We allocate the $i$-th index list to the set of test questions $q_{\text{test}_i} = \left\{q_{\text{test}_i}^{\text{lang}_1}, q_{\text{test}_i}^{\text{lang}_2}, \cdots, q_{\text{test}_i}^{\text{lang}_L} \right\}$ for the same index $i$ across all $L$ language splits. The training-set index list, together with the specified languages,\footnote{
    For the \multilingual mode, we apply the same sampling procedure to generate $N$ HRL code lists of length $K=6$, drawing from the available HRLs in the dataset.
} jointly determines the content and language of the demonstration for each testing example (\cref{fig:icl_modes}).
This approach both ensures linguistic diversity for multilingual prompting and, whenever applicable, mitigates confounding factors that come with semantic inconsistency across examples.
All  \textit{interface} words (see \cref{tab:dataset}) are in the same language as the examples rather than in English.

\vspace{2pt}\noindent\textbf{Inference and metrics.}
Throughout this work, we use greedy decoding for inference, selecting the token with the highest probability at each step.
For \mgsm in need of CoT, we set the maximum token length to $500$; for \xcopa and \xlwic, we set it to 10, as we expect the answers to be short; for \xquad, it is $30$.
We use exact match accuracy as our evaluation metric: for \mgsm, we extract the last numeral in the response; for \xcopa and \xlwic, we extract the label (\textit{expected output}) in the response (\cref{tab:dataset}); for \xquad, a response of an MLLM is considered correct if it contains the gold answer as a substring after normalization.
\section{ICL Mode Evaluation}
\subsection{\multilingual Prompts Surpass \english} \label{sec:icl_mode_eval:multilingual_vs_english}

\begin{figure}[!t]
	\centering
	\begin{subfigure}{.5\textwidth}
        \raggedright
        \begin{tikzpicture}
        \scalefont{0.7}
    	\begin{axis}[
            ybar, 
            bar width=0.12cm, 
            legend columns=-1,
            width=8.5cm, 
            height=4cm,
            % xtick={$\xllamaThree$, $\xllamaThreeOne$, $\xqwenTwo$, $\xqwenTwoFive$, $\xmistral$, $\xaya$, $\xgptThreeFive$, $\xgptFourO$},
            % symbolic x coords={$\xllamaThree$, $\xllamaThreeOne$, $\xqwenTwo$, $\xqwenTwoFive$, $\xmistral$, $\xaya$, $\xgptThreeFive$, $\xgptFourO$}, 
            xtick=data,
            enlarge x limits=0.1,
            symbolic x coords={\xllamaThree, \xllamaThreeOne, \xqwenTwo, \xqwenTwoFive, \xmistral, \xaya, \xgptThreeFive, \xgptFourO},
            % x tick label style={rotate=50, anchor=east, align=left},
            x tick label style={rotate=0, anchor=center, align=center},
            xlabel near ticks,
            ylabel near ticks,
            xmajorgrids=true,
            ymajorgrids=true,
            grid style=dashed,
            legend style={at={(0.5,1.0)}, anchor=center, nodes={scale=1.0, transform shape}},
            legend columns=3, 
            %legend pos=outer north east, 
            ]
            \addplot+[line width=0.36mm, color=OrangeRed] coordinates {(\xllamaThree, 0.642) (\xllamaThreeOne, 0.571) (\xqwenTwo, 0.437) (\xqwenTwoFive, 0.594) (\xmistral, 0.57) 
            (\xaya, 0.294) 
            (\xgptThreeFive, 0.4420) (\xgptFourO, 0.8580)};
            \addlegendentry{\english}
            
            \addplot+[line width=0.36mm, color=Orchid] coordinates {(\xllamaThree, 0.602) (\xllamaThreeOne, 0.66) (\xqwenTwo, 0.475) (\xqwenTwoFive, 0.595) (\xmistral, 0.633) 
            (\xaya, 0.301) 
            (\xgptThreeFive, 0.5100) (\xgptFourO, 0.8470)};
            \addlegendentry{\multilingual}
            
            \addplot+[line width=0.36mm, color=BlueViolet] coordinates {(\xllamaThree, 0.621) (\xllamaThreeOne, 0.685) (\xqwenTwo, 0.554) (\xqwenTwoFive, 0.603) (\xmistral, 0.658) 
            (\xaya, 0.326) 
            (\xgptThreeFive, 0.5490) (\xgptFourO, 0.8520)};
            \addlegendentry{\native}
            
    	\end{axis}
    	\end{tikzpicture}
        \vspace{-8pt}
        \caption{\mgsm}
        \label{fig:vanilla_eval:mgsm}
	\end{subfigure}

	\begin{subfigure}{.5\textwidth}
        \raggedright
    	\begin{tikzpicture}
        \scalefont{0.7}
    	\begin{axis}[
            ybar, bar width=0.12cm, 
            enlarge x limits=.1, 
            legend columns=-1,
            width=8.5cm, 
            height=4cm,
            xtick=data,
            symbolic x coords={\xllamaThree, \xllamaThreeOne, \xqwenTwo, \xqwenTwoFive, \xmistral, \xaya, \xgptThreeFive, \xgptFourO}, 
            x tick label style={rotate=0, anchor=center, align=center},
            xlabel near ticks,
            ylabel near ticks,
            xmajorgrids=true,
            ymajorgrids=true,
            grid style=dashed,
            % legend style={at={(0.5,1.1)}, anchor=south},
            % legend columns=3, 
            % %legend pos=outer north east, 
            % legend style={nodes={scale=1.0, transform shape}},
            ]
            \addplot+[line width=0.36mm, color=OrangeRed,] coordinates {(\xllamaThree, 0.5737) (\xllamaThreeOne, 0.4442) (\xqwenTwo, 0.4846) (\xqwenTwoFive, 0.5782) (\xmistral, 0.5154) 
            (\xaya, 0.5878) 
            (\xgptThreeFive, 0.5372) (\xgptFourO, 0.5635)};
            % \addlegendentry{\english}
            \addplot+[line width=0.36mm, color=Orchid] coordinates {(\xllamaThree, 0.5929) (\xllamaThreeOne, 0.5705) (\xqwenTwo, 0.5628) (\xqwenTwoFive, 0.5596) (\xmistral, 0.5115) 
            (\xaya, 0.5987) 
            (\xgptThreeFive, 0.5590) (\xgptFourO, 0.6526)};
            % \addlegendentry{\multilingual}
            \addplot+[line width=0.36mm, color=BlueViolet] coordinates {(\xllamaThree, 0.6314) (\xllamaThreeOne, 0.6288) (\xqwenTwo, 0.5776) (\xqwenTwoFive, 0.5872) (\xmistral, 0.6019) 
            (\xaya, 0.6173) 
            (\xgptThreeFive, 0.5821) (\xgptFourO, 0.7186)};
            % \addlegendentry{\native}
    	\end{axis}
    	\end{tikzpicture}
        \vspace{-8pt}
        \caption{\xlwic}
        \label{fig:vanilla_eval:xlwic}
	\end{subfigure}

	\begin{subfigure}{.5\textwidth}
        \raggedright
    	\begin{tikzpicture}
        \scalefont{0.7}
    	\begin{axis}[
            ybar, bar width=0.12cm, 
            enlarge x limits=.1, 
            legend columns=-1,
            width=8.5cm, 
            height=4cm,
            xtick=data,
            symbolic x coords={\xllamaThree, \xllamaThreeOne, \xqwenTwo, \xqwenTwoFive, \xmistral, \xaya, \xgptThreeFive, \xgptFourO}, 
            x tick label style={rotate=0, anchor=center, align=center},
            xlabel near ticks,
            ylabel near ticks,
            xmajorgrids=true,
            ymajorgrids=true,
            grid style=dashed,
            % legend style={at={(0.5,1.1)}, anchor=south, nodes={scale=1.0, transform shape}},
            % legend columns=3, 
            % legend pos=outer north east, 
            ]
            \addplot+[line width=0.36mm, color=OrangeRed,] coordinates {(\xllamaThree, 0.4623) (\xllamaThreeOne, 0.5591) (\xqwenTwo, 0.6229) (\xqwenTwoFive, 0.6469) (\xmistral, 0.6226) 
            (\xaya, 0.2654) 
            (\xgptThreeFive, 0.6343) (\xgptFourO, 0.8166)};
            % \addlegendentry{\english}
            \addplot+[line width=0.36mm, color=Orchid] coordinates {(\xllamaThree, 0.624) (\xllamaThreeOne, 0.6611) (\xqwenTwo, 0.6383) (\xqwenTwoFive, 0.6463) (\xmistral, 0.6294) 
            (\xaya, 0.4869) 
            (\xgptThreeFive, 0.6271) (\xgptFourO, 0.8314)};
            % \addlegendentry{\multilingual}
            \addplot+[line width=0.36mm, color=BlueViolet] coordinates {(\xllamaThree, 0.6729) (\xllamaThreeOne, 0.7163) (\xqwenTwo, 0.6763) (\xqwenTwoFive, 0.6769) (\xmistral, 0.7091) 
            (\xaya, 0.6214) 
            (\xgptThreeFive, 0.6717) (\xgptFourO, 0.8597)};
            % \addlegendentry{\native}
    	\end{axis}
    	\end{tikzpicture}
        \vspace{-8pt}
        \caption{\xcopa}
        \label{fig:vanilla_eval:xcopa}
	\end{subfigure}

        \begin{subfigure}{.5\textwidth}
        \raggedright
    	\begin{tikzpicture}
        \scalefont{0.7}
    	\begin{axis}[
            ybar, bar width=0.12cm, 
            enlarge x limits=.1, 
            legend columns=-1,
            width=8.5cm, 
            height=4cm,
            xtick=data,
            symbolic x coords={\xllamaThree, \xllamaThreeOne, \xqwenTwo, \xqwenTwoFive, \xmistral, \xaya}, 
            x tick label style={rotate=0, anchor=center, align=center},
            xlabel near ticks,
            ylabel near ticks,
            xmajorgrids=true,
            ymajorgrids=true,
            grid style=dashed,
            % legend style={at={(0.5,1.1)}, anchor=south, nodes={scale=1.0, transform shape}},
            % legend columns=3, 
            % legend pos=outer north east, 
            ]
            \addplot+[line width=0.36mm, color=OrangeRed,] coordinates {(\xllamaThree, 0.7202) (\xllamaThreeOne, 0.6825) (\xqwenTwo, 0.534) (\xqwenTwoFive, 0.6898) (\xmistral, 0.6093) 
            (\xaya, 0.7048)};
            % \addlegendentry{\english}
            \addplot+[line width=0.36mm, color=Orchid] coordinates {(\xllamaThree, 0.743) (\xllamaThreeOne, 0.7112) (\xqwenTwo, 0.6018) (\xqwenTwoFive, 0.6975) (\xmistral, 0.701) 
            (\xaya, 0.7025)};
            % \addlegendentry{\multilingual}
            \addplot+[line width=0.36mm, color=BlueViolet] coordinates {(\xllamaThree, 0.7522) (\xllamaThreeOne, 0.7128) (\xqwenTwo, 0.649) (\xqwenTwoFive, 0.7008) (\xmistral, 0.706) 
            (\xaya, 0.6812)};
            % \addlegendentry{\native}
    	\end{axis}
    	\end{tikzpicture}
        \vspace{-8pt}
        \caption{\xquad}
        \label{fig:vanilla_eval:xquad}
	\end{subfigure}

    \vspace{-10pt}
    \caption{Average accuracies of LRLs across three ICL modes on our evaluated $4$ datasets and $7$ MLLMs. 
    % In general, \native mode outperforms \multilingual mode, which in turn outperforms \english mode. 
    Raw accuracies of all language splits are in \cref{tab:vanilla_eval:mgsm,tab:vanilla_eval:xcopa,tab:vanilla_eval:xlwic,tab:vanilla_eval:xquad} in \cref{app:expt:vanilla}. For simplicity, on the $x$-axis, only the model logos are labeled -- \xllamaThree: \textsf{Llama3-8B-Instruct}; \xllamaThreeOne: \textsf{Llama3.1-8B-Instruct}; \xqwenTwo: \textsf{Qwen2-7B-Instruct}; \xqwenTwoFive: \textsf{Qwen2.5-7B-Instruct}; \xmistral: \textsf{Mistral-NeMo-12B-Instruct}; \xaya: \textsf{Aya-Expanse-8b}; \xgptThreeFive: \textsf{GPT3.5-turbo}; \xgptFourO: \textsf{GPT4o-mini}.  }
    \vspace{-20pt}
    \label{fig:vanilla_eval_three_modes}
\end{figure}

\begin{table*}[!htbp]
\setlength{\tabcolsep}{4pt}
    \small
    \centering
    \alternaterowcolors
\begin{tabular}{l||cccc|l||cccc|l||l}
\toprule
\textbf{Acc $(\%)_{\Delta{\color{ForestGreen}\uparrow}{\color{OrangeRed}\downarrow}}$}                                       & \multicolumn{5}{c||}{\textbf{\mgsm}}         & \multicolumn{5}{c||}{\textbf{\xlwic}}        & \multicolumn{1}{c}{\textbf{\xcopa}} \\ 
 &
\textbf{bn} &
\textbf{sw} &
\textbf{te} &
\textbf{th} &
  \multicolumn{1}{l||}{\textbf{LRL Avg}} &
\textbf{bg} &
\textbf{et} &
\textbf{fa} &
\textbf{hr} &
  \multicolumn{1}{l||}{\textbf{LRL Avg}} &
  \multicolumn{1}{l}{\textbf{LRL Avg}} \\ \midrule

\multicolumn{12}{l}{\textbf{\llamaThreeOne}} \\

\english                                         & 52.40     & 67.20 & 39.60 & 69.20 & 57.10       & 51.54 & 44.62 & 29.74 & 51.79 & 44.42 & 55.91           \\
\multilingual                                    & 68.00     & 68.80 & 55.60 & 71.60 & \increase{66.00}{8.90}[***] & 57.69 & 55.13 & 57.95 & 56.92 & \increase{57.05}{12.63}[***] & \increase{66.11}{10.20}[***]          \\
\native                                          & 67.20     & 72.40 & 58.00 & 76.40 & \increase{68.50}{11.40}[***] & 61.79 & 62.05 & 62.31 & 57.18 & \increase{62.88}{18.46}[***] & \increase{71.63}{15.72}[***]      \\ \midrule

\multicolumn{12}{l}{\textbf{\qwenTwo}} \\
\english                                         & 57.20     & 20.80 & 23.20 & 73.60 & 43.70          & 28.21  & 56.92 & 63.33 & 54.87 & 48.46 & 62.29           \\
\multilingual                                     & 64.80     & 28.40 & 23.60 & 73.20 & \increase{47.50}{3.80}[**] & 53.59 & 58.46 & 60.26 & 54.36 & \increase{56.28}{7.82}[***] & \increase{63.83}{1.54}[*]          \\
\native                                          & 72.00     & 32.40 & 40.40 & 76.80 & \increase{55.40}{11.70}[***] & 55.13 & 59.23 & 63.08 & 54.62 & \increase{57.76}{9.30}[***] & \increase{67.63}{5.34}[***]          \\ \midrule

\multicolumn{12}{l}{\textbf{\gptThreeFive}} \\
\english                                         & 39.60     & 63.60 & 12.80 & 60.80 & 44.20 & 54.36     & 54.62 & 54.10 & 51.79 & 53.72 & 63.43           \\
\multilingual                                       & 54.40     & 68.00 & 24.40 & 57.20 & \increase{51.00}{6.80}[***] & 52.82     & 60.00 & 54.36 & 56.41 & \increase{55.90}{2.18}[**] & \decrease{62.71}{0.72}        \\
\native                                          & 57.20     & 73.60 & 30.00 & 58.80 & \increase{54.90}{10.70}[***] & 54.62     & 59.49 & 58.46 & 60.26 & \increase{58.21}{4.49}[***] & \increase{67.17}{3.74}[***]          \\

\bottomrule
\end{tabular}
    \caption{Accuracies on LRLs of \english, \multilingual and \native modes across $3$ MLLMs of $3$ datasets. Please refer to \cref{tab:lang25} for language code-to-name mapping.  Avg represents the average accuracy of the LRLs. Subscript indicate the performance \textcolor{ForestGreen}{increase$\uparrow$} (or \textcolor{OrangeRed}{decrease$\downarrow$}) of \multilingual and \native compared to \english. Superscripts are significance levels (in terms of $p$-value) of the same comparison --- *: $p<0.05$; **: $p<0.01$; ***: $p<0.001$. Raw evaluation accuracies and hypothesis test results for all MLLMs and all languages are in \cref{tab:vanilla_eval:mgsm,tab:vanilla_eval:xcopa,tab:vanilla_eval:xlwic} and \cref{tab:hyp_test:vanilla_eval:mgsm,tab:hyp_test:vanilla_eval:xcopa,tab:hyp_test:vanilla_eval:xlwic} in \cref{app:expt:vanilla}, respectively.}
    \vspace{-5pt}
    \label{tab:lrl_vanilla_eval}
\end{table*}

\noindent\textbf{Results.}
We first compare the $6$-shot performance with \english, \multilingual and \native ICL modes on $6$ open-sourced MLLMs and $2$ commercial OpenAI models (\cref{fig:vanilla_eval_three_modes}), with \cref{tab:lrl_vanilla_eval} presenting the detailed performance of three selected MLLMs across various LRLs.
Overall, \multilingual mode outperforms \english mode, both for individual LRLs and on average.
In $23$ out of $30$ cases (\cref{fig:vanilla_eval_three_modes}), \multilingual mode achieves higher accuracy than \english one.
This phenomenon is evident even for \textsf{GPT4o-mini}, one of the currently strongest LLMs \cite{chatbot_arena}, and for HRLs as well (see \cref{tab:vanilla_eval:mgsm,tab:vanilla_eval:xcopa,tab:vanilla_eval:xlwic} in \cref{app:expt:vanilla} for HRL accuracies).
Extending the results of \citet{mgsm} that \multilingual mode generally outperforms \english mode for PaLM \citep{palm} and Codex \citep{codex}, our results confirm this trend is a general phenomenon across various MLLMs and datasets.

\vspace{2pt}\noindent\textbf{Outstanding performance of the \native mode and the practical unfeasibility.}
Admittedly, in $27$ out of $30$ comparisons, \native mode outperforms \multilingual mode (\cref{fig:vanilla_eval_three_modes}), which aligns with the machine-learning intuition that in-domain data, in terms of both genre and language, are more promising for model performance \citep{is_translation_all_you_need}. However, domain-specific datasets for LRLs are often difficult to obtain due to the scarcity of native speakers or professional translators \citep{nllb, nllb-200}; therefore, in practice, it is usually challenging to provide high-quality demonstrations in the same language and domain as the test question.
In contrast, annotations make the HRL-\multilingual mode more feasible in many scenarios.

\vspace{2pt}\noindent\textbf{Hypothesis Tests.} \label{sec:icl_mode_eval:hyp_test}
To verify whether the improvement is statistically significant, we conduct McNemar's test \cite{mcnemar}---the null hypothesis means no significant accuracy difference between the baseline (\english) and the compared mode.
Let $b$ denote the number of cases where the baseline is correct while the compared mode is incorrect, and $c$ denote the number of cases where the baseline mode is erroneous while the compared mode is correct.
We calculate the corrected version \cite{corrected_mcnemar} of the McNemar's statistic:
\begin{equation}
    \begin{aligned}
        \chi^2 = \frac{\left( |b-c| -1 \right)^2}{b+c},
    \end{aligned}
    \label{eq:mcnemar}
\end{equation}
which has a chi-squared distribution with one degree of freedom.
Significant $\chi^2$-test results provide strong evidence to reject the null hypothesis of no accuracy improvement.
Our results results (\cref{tab:lrl_vanilla_eval} and \cref{tab:hyp_test:vanilla_eval:mgsm,tab:hyp_test:vanilla_eval:xcopa,tab:hyp_test:vanilla_eval:xlwic,tab:hyp_test:vanilla_eval:xquad} in \cref{app:expt:vanilla}) indicate that both \multilingual and \native modes significantly outperform the \english mode.

\begin{figure}[!t]
	\centering
	\begin{subfigure}{.5\textwidth}
        \raggedright
    	\begin{tikzpicture}
        \scalefont{0.7}
    	\begin{axis}[
            ybar, bar width=0.06cm, 
            enlarge x limits=.1, 
            legend columns=-1,
            width=8.5cm, height=4cm,
            xtick=data,
            symbolic x coords={\xllamaThree, \xllamaThreeOne, \xqwenTwo, \xqwenTwoFive, \xmistral, \xaya, \xgptThreeFive, \xgptFourO}, 
            x tick label style={rotate=0, anchor=center, align=center},
            xlabel near ticks,
            ylabel near ticks,
            xmajorgrids=true,
            ymajorgrids=true,
            grid style=dashed,
            legend style={at={(0.5,1.1)}, anchor=center, nodes={scale=1.0, transform shape}, font=\tiny},
            legend columns=5, 
            %legend pos=outer north east, 
            % legend style={nodes={scale=1.0, transform shape}},
            ]
            \addplot+[line width=0.36mm, color=OrangeRed,] coordinates {(\xllamaThree, 0.642) (\xllamaThreeOne, 0.571) (\xqwenTwo, 0.437) (\xqwenTwoFive, 0.594) (\xmistral, 0.57) 
            (\xaya, 0.294) 
            (\xgptThreeFive, 0.4420) (\xgptFourO, 0.8580)};
            \addlegendentry{\english}
            
            \addplot+[line width=0.36mm, color=Peach] coordinates {(\xllamaThree, 0.629) (\xllamaThreeOne, 0.619) (\xqwenTwo, 0.429) (\xqwenTwoFive, 0.603) (\xmistral, 0.643) 
            (\xaya, 0.307) 
            (\xgptThreeFive, 0.4200) (\xgptFourO, 0.8630)};
            \addlegendentry{\french}
            
            \addplot+[line width=0.36mm, color=SkyBlue] coordinates {(\xllamaThree, 0.63) (\xllamaThreeOne, 0.634) (\xqwenTwo, 0.447) (\xqwenTwoFive, 0.6) (\xmistral, 0.646) 
            (\xaya, 0.317) 
            (\xgptThreeFive, 0.4390) (\xgptFourO, 0.8660)};
            \addlegendentry{\chinese}

            \addplot+[line width=0.36mm, color=NavyBlue] coordinates {(\xllamaThree, 0.622) (\xllamaThreeOne, 0.681) (\xqwenTwo, 0.429) (\xqwenTwoFive, 0.604) (\xmistral, 0.636) 
            (\xaya, 0.304) 
            (\xgptThreeFive, 0.4310) (\xgptFourO, 0.8450)};
            \addlegendentry{\japanese}

            \addplot+[line width=0.36mm, color=Orchid] coordinates {(\xllamaThree, 0.602) (\xllamaThreeOne, 0.66) (\xqwenTwo, 0.475) (\xqwenTwoFive, 0.595) (\xmistral, 0.633) 
            (\xaya, 0.301) 
            (\xgptThreeFive, 0.5100) (\xgptFourO, 0.8470)};
            \addlegendentry{\multilingual}
            
    	\end{axis}
    	\end{tikzpicture}
        \vspace{-8pt}
        \caption{\mgsm}
        \label{fig:mono_vs_multi:mgsm}
	\end{subfigure}

	\begin{subfigure}{.5\textwidth}
        \raggedright
    	\begin{tikzpicture}
        \scalefont{0.7}
    	\begin{axis}[
            ybar, bar width=0.06cm, 
            enlarge x limits=.1, 
            legend columns=-1,
            width=8.5cm, height=4cm,
            xtick=data,
            symbolic x coords={\xllamaThree, \xllamaThreeOne, \xqwenTwo, \xqwenTwoFive, \xmistral, \xaya, \xgptThreeFive, \xgptFourO}, 
            x tick label style={rotate=0, anchor=center, align=center},
            xlabel near ticks,
            ylabel near ticks,
            xmajorgrids=true,
            ymajorgrids=true,
            grid style=dashed,
            legend style={at={(0.5,1.1)}, anchor=center, nodes={scale=1.0, transform shape}, font=\tiny},
            legend columns=5, 
            % %legend pos=outer north east, 
            % legend style={nodes={scale=1.0, transform shape}},
            ]
            \addplot+[line width=0.36mm, color=OrangeRed,] coordinates {(\xllamaThree, 0.5737) (\xllamaThreeOne, 0.4442) (\xqwenTwo, 0.4846) (\xqwenTwoFive, 0.5782) (\xmistral, 0.5154) 
            (\xaya, 0.5878) 
            (\xgptThreeFive, 0.5372) (\xgptFourO, 0.5635)};
            \addlegendentry{\english}
            
            \addplot+[line width=0.36mm, color=Peach] coordinates {(\xllamaThree, 0.5904) (\xllamaThreeOne, 0.4609) (\xqwenTwo, 0.5551) (\xqwenTwoFive, 0.5545) (\xmistral, 0.5231) 
            (\xaya, 0.5994) 
            (\xgptThreeFive, 0.5564) (\xgptFourO, 0.5878)};
            \addlegendentry{\french}
            
            \addplot+[line width=0.36mm, color=SkyBlue] coordinates {(\xllamaThree, 0.575) (\xllamaThreeOne, 0.5538) (\xqwenTwo, 0.5468) (\xqwenTwoFive, 0.5417) (\xmistral, 0.4949) 
            (\xaya, 0.5968) 
            (\xgptThreeFive, 0.5500) (\xgptFourO, 0.6135)};
            \addlegendentry{\chinese}

            \addplot+[line width=0.36mm, color=NavyBlue] coordinates {(\xllamaThree, 0.5994) (\xllamaThreeOne, 0.5436) (\xqwenTwo, 0.5731) (\xqwenTwoFive, 0.5365) (\xmistral, 0.4981) 
            (\xaya, 0.6109) 
            (\xgptThreeFive, 0.5532) (\xgptFourO, 0.5782)};
            \addlegendentry{\japanese}

            \addplot+[line width=0.36mm, color=Orchid] coordinates {(\xllamaThree, 0.5929) (\xllamaThreeOne, 0.5705) (\xqwenTwo, 0.5628) (\xqwenTwoFive, 0.5596) (\xmistral, 0.5115) 
            (\xaya, 0.5987) 
            (\xgptThreeFive, 0.5590) (\xgptFourO, 0.6526)};
            \addlegendentry{\multilingual}
            
    	\end{axis}
    	\end{tikzpicture}
        \vspace{-8pt}
        \caption{\xlwic}
        \label{fig:mono_vs_multi:xlwic}
	\end{subfigure}

	\begin{subfigure}{.5\textwidth}
        \raggedright
    	\begin{tikzpicture}
        \scalefont{0.7}
    	\begin{axis}[
            ybar, bar width=0.08cm, 
            enlarge x limits=.1, 
            legend columns=-1,
            width=8.5cm,
            height=4cm,
            xtick=data,
            symbolic x coords={\xllamaThree, \xllamaThreeOne, \xqwenTwo, \xqwenTwoFive, \xmistral, \xaya, \xgptThreeFive, \xgptFourO}, 
            x tick label style={rotate=0, anchor=center, align=center},
            xlabel near ticks,
            ylabel near ticks,
            xmajorgrids=true,
            ymajorgrids=true,
            grid style=dashed,
            legend style={at={(0.5,1.1)}, anchor=center, nodes={scale=1.0, transform shape}, font=\tiny},
            legend columns=4, 
            ]
            \addplot+[line width=0.36mm, color=OrangeRed,] coordinates {(\xllamaThree, 0.4623) (\xllamaThreeOne, 0.5591) (\xqwenTwo, 0.6229) (\xqwenTwoFive, 0.6469) (\xmistral, 0.6226) 
            (\xaya, 0.2654) 
            (\xgptThreeFive, 0.6593) (\xgptFourO, 0.8697) };
            \addlegendentry{\english}
            
            \addplot+[line width=0.36mm, color=Peach] coordinates {(\xllamaThree, 0.5966) (\xllamaThreeOne, 0.5951) (\xqwenTwo, 0.64) (\xqwenTwoFive, 0.6454) (\xmistral, 0.6209) 
            (\xaya, 0.2931) 
            (\xgptThreeFive, 0.6577) (\xgptFourO, 0.8580) };
            \addlegendentry{\italian}
            
            \addplot+[line width=0.36mm, color=SkyBlue] coordinates {(\xllamaThree, 0.6189) (\xllamaThreeOne, 0.6226) (\xqwenTwo, 0.6323) (\xqwenTwoFive, 0.6534) (\xmistral, 0.6157) 
            (\xaya, 0.3311) 
            (\xgptThreeFive, 0.6547) (\xgptFourO, 0.8750) };
            \addlegendentry{\chinese}

            \addplot+[line width=0.36mm, color=Orchid] coordinates {(\xllamaThree, 0.624) (\xllamaThreeOne, 0.6611) (\xqwenTwo, 0.6383) (\xqwenTwoFive, 0.6463) (\xmistral, 0.6294) 
            (\xaya, 0.4869) 
            (\xgptThreeFive, 0.6497) (\xgptFourO, 0.8863) };
            \addlegendentry{\multilingual}
            
    	\end{axis}
    	\end{tikzpicture}
        \vspace{-8pt}
        \caption{\xcopa}
        \label{fig:mono_vs_multi:xcopa}
	\end{subfigure}

	\begin{subfigure}{.5\textwidth}
        \raggedright
    	\begin{tikzpicture}
        \scalefont{0.7}
    	\begin{axis}[
            ybar, bar width=0.08cm, 
            enlarge x limits=.1, 
            legend columns=-1,
            width=8.5cm,
            height=4cm,
            xtick=data,
            symbolic x coords={\xllamaThree, \xllamaThreeOne, \xqwenTwo, \xqwenTwoFive, \xmistral, \xaya}, 
            x tick label style={rotate=0, anchor=center, align=center},
            xlabel near ticks,
            ylabel near ticks,
            xmajorgrids=true,
            ymajorgrids=true,
            grid style=dashed,
            legend style={at={(0.5,1.1)}, anchor=center, nodes={scale=1.0, transform shape}, font=\tiny},
            legend columns=4, 
            ]
            \addplot+[line width=0.36mm, color=OrangeRed,] coordinates {(\xllamaThree, 0.7202) (\xllamaThreeOne, 0.6825) (\xqwenTwo, 0.534) (\xqwenTwoFive, 0.6898) (\xmistral, 0.6093) 
            (\xaya, 0.7048) };
            \addlegendentry{\english}
            
            \addplot+[line width=0.36mm, color=Peach] coordinates {(\xllamaThree, 0.7293) (\xllamaThreeOne, 0.692) (\xqwenTwo, 0.5967) (\xqwenTwoFive, 0.6895) (\xmistral, 0.6862) 
            (\xaya, 0.6742) };
            \addlegendentry{\german}
            
            \addplot+[line width=0.36mm, color=SkyBlue] coordinates {(\xllamaThree, 0.7353) (\xllamaThreeOne, 0.6798) (\xqwenTwo, 0.524) (\xqwenTwoFive, 0.6725) (\xmistral, 0.6888) 
            (\xaya, 0.693) };
            \addlegendentry{\chinese}

            \addplot+[line width=0.36mm, color=Orchid] coordinates {(\xllamaThree, 0.743) (\xllamaThreeOne, 0.7112) (\xqwenTwo, 0.6018) (\xqwenTwoFive, 0.6975) (\xmistral, 0.7008) 
            (\xaya, 0.7025) };
            \addlegendentry{\multilingual}
            
    	\end{axis}
    	\end{tikzpicture}
        \vspace{-8pt}
        \caption{\xquad}
        \label{fig:mono_vs_multi:xquad}
	\end{subfigure}

    \vspace{-10pt}
    \caption{\monolingual modes vs \multilingual on average accuracies of LRLs. The $x$-axis is the same as in \cref{fig:vanilla_eval_three_modes}. 
    % In the considerable comparisons, each \monolingual mode outperforms \english mode. Multilingual mode is the most robust among ICL modes in terms of performance boosting. 
    Raw evaluation accuracies are in \cref{tab:vanilla_eval:mgsm,tab:vanilla_eval:xcopa,tab:vanilla_eval:xlwic,tab:vanilla_eval:xquad} in \cref{app:expt:vanilla}.}
    \vspace{-5pt}
    \label{fig:mono_vs_multi}
    
\end{figure}

\subsection{Ablation Study: Non-English \monolingual Prompts Are Effective} \label{sec:icl_mode_eval:monolingual}

With the success of the \multilingual mode, we investigate whether the improvement comes from the introduction of multiple languages or simply from a single non-English HRL.
% We conduct further experiments to verify whether \monolingual in a single non-English HRL could achieve accuracy improvements over \english.
Specifically, we compare several \monolingual modes including \chinese (for all $4$ datasets), \french, \japanese (both for \mgsm and \xlwic), \italian (for \xcopa) and \german (for \xquad).
We select French, Italian and German because they are European languages and thus share considerable subword overlap with English; in contrast, Chinese and Japanese exhibit little subword overlap with Latin-script languages, but there is a substantial overlap between the two due to their shared use of Chinese characters (or \textit{Hanzi, Kanji}). This language selection allows us to analyze simultaneously the impact of the writing system (or \textit{subword overlap}) on ICL performance.

\vspace{2pt}\noindent\textbf{Results.} \label{sec:mono_vs_multilingual_results}
All HRL-\monolingual modes outperform \english in a considerable number of comparisons (\cref{fig:mono_vs_multi}).
This finding also holds for HRL evaluations (see \cref{tab:vanilla_eval:mgsm,tab:vanilla_eval:xcopa,tab:vanilla_eval:xlwic,tab:vanilla_eval:xquad} in \cref{app:expt:vanilla} for raw accuracies of each language). Among all \monolingual modes, \chinese performs the best, with $20$ out of $30$ comparisons when accuracy surpasses that of \english.
\japanese also frequently outperforms \english.
Extending the findings of \citet{revisit_primacy_of_english} that non-English languages are more effective than English in pretraining and fine-tuning based cross-lingual transfer, our results suggest that non-English languages, particularly those with non-Latin scripts, may be more effective under the prompting scheme as well.

However, the \multilingual mode exhibits stronger robustness, outperforming \chinese in $23$ out of $30$ LRL cases.
The same trend applies to HRLs.
The results of the hypothesis test further confirm the robustness: for both LRL and HRL splits, the \multilingual mode exhibits the highest number of significant results relative to other \monolingual modes (\cref{tab:hyp_test:vanilla_eval:mgsm,tab:hyp_test:vanilla_eval:xcopa,tab:hyp_test:vanilla_eval:xlwic,tab:hyp_test:vanilla_eval:xquad} in \cref{app:expt:vanilla}).
Intuitively, we hypothesize that \multilingual mode functions like an ``average'' of individual HRL-\monolingual modes, making it most robust, and thus it outperforms \english most frequently and achieves the highest overall average accuracy. 

Following \citet{language_specific_neuron_msra}, we further identify ICL-mode-specific neurons and find that the neurons activated by \multilingual overlap most with those activated by \native among other modes. This further explains why \multilingual could achieve performance comparable to \native. See \cref{app:neuron} for details.

\subsection{Ablation Study: Merely Introducing New Language(s) Enhances ICL Performance} \label{sec:icl_mode_eval:cis}

\begin{figure}[!t]
    \centering
    \includegraphics[width=\linewidth]{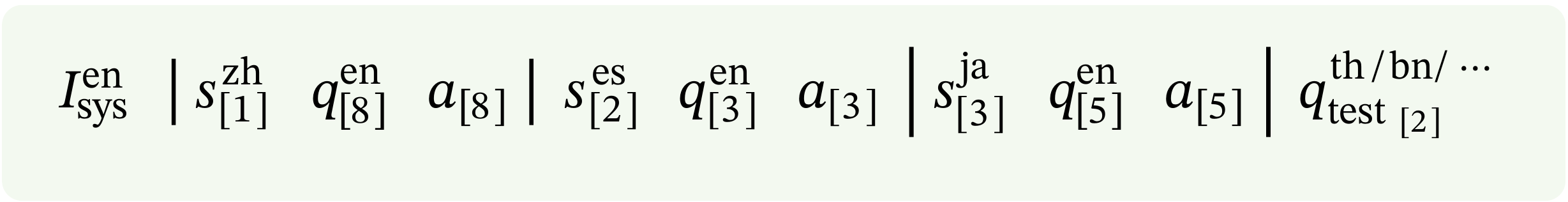}
    % 添加居中文本框
    \centering
    {\scriptsize
        \parbox{\linewidth}{%
            \centering
            English ICL Mode $+$ Multilingual Context-Irrelevant Sentences\\
            (\english$+\ $\cisMulti)
        }
    }
    \vspace{-18pt}
    \caption{Prepending multilingual \cis (\cisMulti) $\{s_i^{\text{lang}}\}_{i=1}^K \sim \mathcal{S}^{\text{lang}}$ to demonstrations $\{(q_i^{\text{en}}, a_i)\}_{i=1}^K \sim \mathcal{D}_\text{train}^{\text{en}}$ of \english ICL template illustrated in \cref{fig:icl_modes}a.}
    \label{fig:irr_sent_illustration}
    \vspace{-5pt}
\end{figure}

After showing that including non-English in the prompts can improve ICL performance, a natural follow-up question arises: does the gain come from the mere presence of non-English languages, or the interaction between the target language and the in-topic examples? 
To distinguish these two setups, we prepend a \textit{context-irrelevant sentence} (\cis) $s^{\text{lang}}$ before each ICL demonstration, which is unrelated to the current domain and can be in any language.
Analogous to the ICL modes introduced in \cref{sec:multiling_icl:modes}, CIS resembles the construction of those modes with the same sampling strategy (\cref{sec:setup}).
For example, based on the \english mode, we could prepend a set of multilingual \cis, which augments \cref{fig:icl_modes}a into \cref{fig:irr_sent_illustration}. We denote such setting by ``\english$+\ $\cisMulti''. This naming convention applies to other settings accordingly.

We use sentences from \flores \cite{flores101} as the source of our CIS, which provides parallel Wikipedia sentences in multiple languages, a fairly distant genre to all evaluation datasets.
We filter sentence-parallel sets with English word counts ranging from 10 to 15 as our sampling pool $\mathcal{S}^\text{lang}$, a small range compared to target datasets (\cref{tab:dataset}), to mitigate the risk of introducing too much noise.
Details of \flores can be found at \cref{app:dataset:flores}.

\vspace{2pt}\noindent\textbf{Results and analysis.}
We perform hypothesis testing as in \cref{sec:icl_mode_eval:hyp_test} to compare different \cis settings with \english$+\ $\cisEn (\cref{tab:irr_sent})---\english$+\ $\cisEn exhibit a small drop compared to \english only, indicating that our filtration effectively controls the negative impact of noise to a tolerable level.
We find that introducing a single non-English language generally improves ICL performance in most cases $\left(\frac{51}{72} \approx 71\%\right)$; however, the improvement is only statistically significant in $\frac{22}{51} \approx 43\%$ cases.
This reveals that simply introducing a language can lead to a modest improvement in MLLM performance, but it is more pronounced when in-topic demonstrations in another language (\monolingual modes) are incorporated.
More hypothesis test results for both LRL and HRL splits can be found in \cref{tab:hyp_test:cis:mgsm,tab:hyp_test:cis:xcopa,tab:hyp_test:cis:xlwic} in \cref{app:expt:cis}.

Introducing multiple languages (\cisMulti, \cref{fig:irr_sent_illustration}) is slightly more promising than \cis in a single language, with $\frac{18}{24}\approx 75\%$ of cases showing improvement, of which $\frac{11}{18}\approx 61\%$ are significant.
We further conduct experiments by prepending multilingual \cis to the \multilingual mode (\multilingual$+\ $\cisMulti, \cref{tab:irr_sent}). The performance of \multilingual$+\ $\cisMulti is not significantly lower than that of \english$+\ $\cisMulti; in $\frac{16}{24}\approx 67\%$ cases, it significantly outperforms \english$+\ $\cisMulti. This concludes that the significant improvement from \english mode to \multilingual mode is attributed to the incorporation of multiple languages with in-topic demonstrations.

\begin{table}[!htbp]
    \setlength{\tabcolsep}{0.5pt}
    \small
    \centering
    \alternaterowcolors
    \begin{tabular}{l|llll}
        \toprule
        % {\textbf{LRL Avg Acc}$_{\Delta{\color{ForestGreen}\uparrow}{\color{OrangeRed}\downarrow}}$} $(\%)$
        \textbf{LRL }$\overline{\textbf{Acc}}$ & \textbf{\mgsm}              & \textbf{\xlwic}             & \textbf{\xcopa}     & \textbf{\xquad}          \\ \midrule
        \multicolumn{5}{l}{\textbf{\llamaThree}}                                                                                                                                                  \\
        \english                                                                                       & 64.20                       & 57.37                       & 46.23  & 72.02                      \\
        $\quad +\ $\cisEn                                                                            & 61.00                       & 56.73                       & 48.37  & 71.43                       \\
        $\quad +\ $\cisFr                                                                            & \increase{61.60}{0.30}      & \increase{57.50}{0.77}      & \increase{58.49}{10.12}[***] &  \decrease{70.98}{0.45}\\
        $\quad +\ $\cisJa                                                                            & \increase{61.30}{0.30}      & \increase{58.14}{1.41}      & \increase{58.69}{10.32}[***] & \decrease{70.85}{0.58} \\
        $\quad +\ $\cisZh                                                                            & \decrease{59.30}{1.70}      & \increase{58.27}{1.54}      & \increase{58.03}{9.66}[***]  & \decrease{71.28}{0.15} \\
        $\quad +\ $\cisWorld                                                                         & \decrease{60.50}{0.50}      & \increase{57.82}{1.09}      & \increase{61.14}{12.77}[***] & \decrease{70.90}{0.53} \\
        \cmidrule(lr){1-5}
        \world+\cisWorld                                                       & \decrease{60.10}{0.40}      & \increase{58.33}{0.51}      & \increase{61.54}{0.40}  & \increase{74.08}{3.18}[***]      \\

        \midrule
        \multicolumn{5}{l}{\textbf{\llamaThreeOne}}                                                                                                                                               \\
        \english                                                                                       & 57.10                       & 44.42                       & 55.91    &68.25                    \\
        $\quad+\ $\cisEn                                                                            & 55.90                       & 47.88                       & 55.46 & 68.45                        \\
        $\quad+\ $\cisFr                                                                            & \decrease{52.10}{3.80}[*]   & \increase{52.82}{4.94}[***] & \increase{59.40}{3.94}[***]  & \increase{68.95}{0.50}\\
        $\quad+\ $\cisJa                                                                            & \increase{58.80}{2.90}      & \increase{55.96}{8.08}[***] & \increase{59.86}{4.40}[***]  & \increase{68.90}{0.45}\\
        $\quad+\ $\cisZh                                                                            & \decrease{55.00}{0.90}      & \increase{54.68}{6.80}[***] & \increase{64.66}{9.20}[***]  & \increase{69.10}{0.65} \\
        $\quad+\ $\cisWorld                                                                         & \increase{62.50}{6.60}[***] & \increase{56.03}{8.15}[***] & \increase{64.74}{9.28}[***]  & \increase{69.60}{1.15}[***] \\
        \cmidrule(lr){1-5}
        \world+\cisWorld                                                     & \increase{68.60}{6.10}[***] & \increase{57.24}{1.21}      & \increase{66.20}{1.46}[*]    & \increase{71.13}{1.53}[***]  \\

        \midrule
        \multicolumn{5}{l}{\textbf{\qwenTwo}}                                                                                                                                                     \\
        \english                                                                                       & 43.70                       & 48.46                       & 62.29     &53.40                   \\
        $\quad+\ $\cisEn                                                                            & 43.00                       & 50.90                       & 62.31  & 58.55                       \\
        $\quad+\ $\cisFr                                                                            & \increase{43.50}{0.50}      & \increase{53.65}{2.75}[***] & $62.31_{0.00-}$       & \increase{60.90}{2.35}[***]      \\
        $\quad+\ $\cisJa                                                                            & \increase{43.80}{0.80}      & \increase{56.22}{5.32}[**]  & \increase{63.09}{0.78}   & \increase{60.58}{2.03}[***]     \\
        $\quad+\ $\cisZh                                                                            & \decrease{42.60}{0.40}      & \increase{56.79}{5.89}[***] & \increase{62.49}{0.18}   & \increase{58.95}{0.40}     \\
        $\quad+\ $\cisWorld                                                                         & \decrease{42.70}{0.30}      & \increase{54.94}{4.04}[***] & \increase{62.51}{0.20}   & \increase{60.33}{1.77}[***]     \\
        \cmidrule(lr){1-5}
        \world+\cisWorld                                                     & \increase{47.30}{4.70}[***] & \increase{55.83}{0.89}      & \increase{63.51}{1.00}  & \increase{61.93}{1.60}[***]      \\

        \midrule
        \multicolumn{5}{l}{\textbf{\qwenTwoFive}}                                                                                                                                                 \\
        \english                                                                                       & 59.40                       & 57.82                       & 64.69            & 68.98            \\
        $\quad+\ $\cisEn                                                                            & 59.40                       & 59.04                       & 64.20 &69.05                        \\
        $\quad+\ $\cisFr                                                                            & $59.40_{0.00-}$             & $59.04_{0.00-}$             & \decrease{64.11}{0.09}   & \increase{69.13}{0.08}    \\
        $\quad+\ $\cisJa                                                                            & \increase{60.10}{0.70}      & \increase{59.36}{0.32}      & $64.20_{0.00-}$         & \decrease{68.65}{0.40}       \\
        $\quad+\ $\cisZh                                                                            & \increase{60.90}{1.50}      & \increase{59.23}{0.19}      & \increase{65.20}{1.00}[*]    & \decrease{68.25}{0.80}[*]  \\
        $\quad+\ $\cisWorld                                                                         & \increase{59.50}{0.10}      & \increase{59.36}{0.32}      & \increase{65.06}{0.86}   & \decrease{68.63}{0.42}     \\
        \cmidrule(lr){1-5}
        \world+\cisWorld                                                       & \decrease{59.20}{0.30}      & \decrease{56.79}{2.57}[*]   & \decrease{64.14}{0.92}       & \increase{69.63}{1.00}[**] \\

        \midrule
        \multicolumn{5}{l}{\textbf{\mistral}}                                                                                                                                                     \\
        \english                                                                                       & 57.00                       & 51.54                       & 62.26             &60.93          \\
        $\quad+\ $\cisEn                                                                            & 60.70                       & 49.62                       & 61.00  & 56.48                      \\
        $\quad+\ $\cisFr                                                                            & \increase{61.40}{0.70}      & \increase{50.58}{0.96}[*]   & \increase{61.23}{0.23}   & \decrease{51.15}{5.33}[***]    \\
        $\quad+\ $\cisJa                                                                            & \decrease{60.20}{0.50}      & \increase{49.87}{0.25}      & \increase{61.80}{0.80}  & \decrease{49.90}{6.58}[***]     \\
        $\quad+\ $\cisZh                                                                            & \decrease{60.50}{0.20}      & \increase{50.06}{0.44}      & \increase{61.14}{0.14}  & \decrease{50.05}{6.43}[***]     \\
        $\quad+\ $\cisWorld                                                                         & \increase{64.90}{4.20}[***] & \increase{50.19}{0.57}      & \increase{62.23}{1.23}[*]  & \decrease{51.85}{4.63}[***]   \\
        \cmidrule(lr){1-5}
        \world+\cisWorld                                                      & \decrease{62.90}{2.00}      & \increase{52.76}{2.57}[**]  & \decrease{61.97}{0.26}      & \increase{69.38}{17.53}[***] \\

        \midrule
        \multicolumn{5}{l}{\textbf{\aya}}                                                                                                                                                         \\
        \english                                                                                       & 29.40                       & 58.78                       & 26.54         & 70.48              \\
        $\quad+\ $\cisEn                                                                            & 27.80                       & 57.82                       & 23.20  & 70.30                       \\
        $\quad+\ $\cisFr                                                                            & \increase{28.70}{0.90}      & \increase{61.54}{3.72}[***] & \increase{28.71}{5.51}[***] & \decrease{69.68}{0.62}[*]   \\
        $\quad+\ $\cisJa                                                                            & \increase{29.30}{1.50}      & \increase{61.03}{3.21}[**]  & \increase{33.06}{9.86}[***] & \decrease{70.00}{0.30} \\
        $\quad+\ $\cisZh                                                                            & \increase{28.60}{0.80}      & \increase{60.58}{2.76}[**]  & \increase{26.94}{3.74}[***] & \decrease{70.25}{0.05} \\
        $\quad+\ $\cisWorld                                                                         & \increase{27.90}{0.10}      & \increase{62.24}{4.42}[***] & \increase{33.60}{10.40}[***] & \decrease{70.10}{0.20} \\
        \cmidrule(lr){1-5}
        \world+\cisWorld                                                       & \increase{31.50}{3.60}[**]  & \decrease{61.09}{1.15}      & \increase{45.91}{12.31}[***] & \decrease{69.70}{0.40} \\
        \bottomrule
    \end{tabular}

    \vspace{-.2cm}
    \caption{Average accuracies (\%) on LRLs after prepending English, monolingual or multilingual \cis to the original \english mode. Subscripts denote the accuracy delta between the current value and that of \english$+\ $\cisEn. Except for \multilingual$+\ $\cisMultilingual (i.e., \world+\cisWorld where \world denotes \multilingual), subscripts represent the difference of the current value and \english$+\ $\cisWorld. The asterisk superscript indicates the significance level, which we compare in the same way as the accuracy delta . Raw evaluation accuracies for all languages are in \cref{tab:cis:mgsm,tab:cis:xcopa,tab:cis:xlwic,tab:cis:xquad} in \cref{app:expt:cis}. Raw hypothesis test results for \cis mode comparisons are in \cref{tab:hyp_test:cis:mgsm,tab:hyp_test:cis:xcopa,tab:hyp_test:cis:xlwic,tab:hyp_test:cis:xquad} in \cref{app:expt:cis}.}
    % \vspace{-.2cm}
    \label{tab:irr_sent}
\end{table}

\subsection{Translation-Based Performance} \label{sec:icl_mode_eval:translation}
Mirroring the translation-training \interalia{xtreme} setup, existing work suggest that translating from LRLs into English and prompting with the translation results may yield better results \interalia{mega}.
While translation is not the main focus of our work, we conduct experiment to compare the performance of translation-based strategies for reference.

\vspace{2pt}\noindent\textbf{Strategies.}
We test two translation strategies for baseline comparison.
(1) \transEn: Translating test questions in other languages in the \english ICL mode (\cref{fig:icl_modes}a) into English.
(2) \transSource: Translating demonstrations in \english ICL mode (\cref{fig:icl_modes}a) into the source language of the current test question, which mirrors the \native mode (\cref{fig:icl_modes}d).
We use the Google Cloud Translation API for translation.\footnote{\url{https://cloud.google.com/translate/}}

\begin{table}[!t]
    \setlength{\tabcolsep}{4pt}
    \small
    \centering
    \alternaterowcolors
\begin{tabular}{l|cc|cc}
\toprule
\textbf{Avg Acc} $(\%)$ & \multicolumn{2}{c|}{\mgsm} & \multicolumn{2}{c}{\xcopa} \\
~ & \textbf{LRL}   & \textbf{HRL}    & \textbf{LRL}   & \textbf{HRL}      \\
\midrule
\multicolumn{5}{l}{\textbf{\llamaThreeOne}}                                    \\
\multilingual         & $\dynamicCellColor{66.00}$ & $\dynamicCellColor{79.20}$ & $\dynamicCellColor{66.11}$ & $\dynamicCellColor{89.64}$ \\
\native               & $\dynamicCellColor{68.50}$ & $\dynamicCellColor{79.43}$ & $\dynamicCellColor{71.63}$ & $\dynamicCellColor{90.80}$ \\
\transEn              & $\dynamicCellColor{60.00}$ & $\dynamicCellColor{74.80}$ & $\dynamicCellColor{76.74}$ & $\dynamicCellColor{89.68}$ \\
\transSource          & $\dynamicCellColor{68.60}$ & $\dynamicCellColor{80.63}$ & $\dynamicCellColor{70.49}$ & $\dynamicCellColor{90.04}$ \\
\midrule

\multicolumn{5}{l}{\textbf{\qwenTwoFive}}                                      \\
\multilingual         & $\dynamicCellColor{59.50}$ & $\dynamicCellColor{86.97}$ & $\dynamicCellColor{64.63}$ & $\dynamicCellColor{91.84}$ \\
\native               & $\dynamicCellColor{60.30}$ & $\dynamicCellColor{87.31}$ & $\dynamicCellColor{67.69}$ & $\dynamicCellColor{92.64}$  \\
\transEn              & $\dynamicCellColor{63.40}$ & $\dynamicCellColor{78.63}$ & $\dynamicCellColor{80.49}$ & $\dynamicCellColor{91.52}$  \\
\transSource          & $\dynamicCellColor{59.90}$ & $\dynamicCellColor{86.97}$ & $\dynamicCellColor{66.23}$ & $\dynamicCellColor{92.48}$  \\
\midrule

\multicolumn{5}{l}{\textbf{\mistral}}                                          \\
\multilingual         & $\dynamicCellColor{63.30}$ & $\dynamicCellColor{81.09}$ & $\dynamicCellColor{62.94}$ & $\dynamicCellColor{88.16}$  \\
\native               & $\dynamicCellColor{65.80}$ & $\dynamicCellColor{81.26}$ & $\dynamicCellColor{70.91}$ & $\dynamicCellColor{90.28}$ \\
\transEn              & $\dynamicCellColor{61.60}$ & $\dynamicCellColor{76.00}$ & $\dynamicCellColor{77.83}$ & $\dynamicCellColor{89.32}$ \\
\transSource          & $\dynamicCellColor{66.60}$ & $\dynamicCellColor{82.00}$ & $\dynamicCellColor{68.46}$ & $\dynamicCellColor{90.52}$ \\
\bottomrule
\end{tabular}
    \caption{Average accuracies (\%) on \mgsm and \xcopa datasets of translation strategies. The comparison includes two translation strategies, the \native mode and our proposed \multilingual mode, evaluated across low- and high-resource language splits. Raw accuracies of individual languages and more models are recorded in \cref{tab:vanilla_eval:mgsm,tab:vanilla_eval:xcopa} in \cref{app:expt:vanilla}.}
    \vspace{-5pt}
    \label{tab:translation}
\end{table}

\vspace{2pt}\noindent\textbf{Analysis.}
For LRLs, \transEn sometimes outperform the \native mode, while \transSource performs comparably to the \multilingual mode, but falls short of the \native mode performance (\cref{tab:translation}).
These results resonate with the phenomena that the translation quality of LRL$\rightarrow$En is generally higher than that of the reversed direction \cite{beyond_english_centric_multilingual_mt,flores101,nllb}.

For HRLs, however, \transEn underperforms to \native and, sometimes, even \multilingual (e.g., on \mgsm), suggestting that if a language is sufficiently well-trained, generating responses directly in that language is more effective than translating into English before inference.
These two findings highlight that MLLMs approach the ideal of being equally capable in HRLs \cite{is_translation_all_you_need}, but are still undertrained on LRLs, making translation into English a favored strategy.

In agreement with \citet{roles_of_english}, we would like to note that even if the task performance of translation-based strategies are the best, the ultimate goal of multilingualism is not just about optimizing task-specific performances.
A universal language model should be able to understand and generate text in all languages, instead of relying on specific language(s) as an intermediary.
On the other hand, due to the loss of semantic nuances, grammatical structures and cultural context, translation-based strategies may not be the best choice for tasks heavily reliant on language-specific nuances \cite{is_translation_all_you_need}.

\section{Related Work}
\noindent\textbf{Prompt engineering.} Instruction tuning aligns LLMs more closely with human instructions \cite{instructGPT,cross_task_generalization, ft_lm_are_zs_learners,general_language_assistant,super_natural_instructions, self_instruct}. Concurrently, numerous prompting strategies have been developed \cite{prompt_survey} and shown to consistently enhance the performance of instruction-tuned LLMs, such as in-context learning \cite{lm_are_few_shot_learner,rethink_demonstration} and chain-of-thought \cite{cot,lm_zero_shot_reasoner}. These prompting strategies are proven effective in multilingual tasks as well \cite{lm_few_shot_multilingual_learner,few_shot_learning_multilingual_lm, mgsm}. Concurrent work \cite{multilingual_empower_reasoning} shows that multilingual prompting has improved the general reasoning capabilities of multilingual LLMs. 

\vspace{2pt}\noindent\textbf{Multilingual ICL.} 
For languages of templates, demonstrations and sample questions in native languages are conventionally inserted into a predefined English template \cite{few_shot_learning_multilingual_lm,polyglot_prompt,cross_lingual_prompting,not_all_language_are_created_equal,mega,plug}.  
\citet{roles_of_english} critiques this widespread misuse of English as the \textit{interface} language. \citet{cross_lingual_prompting,not_all_language_are_created_equal,plug} guide models to ``think'' and generate CoT in English, regardless of the input language, leading to improved performance for generation tasks compared to ``thinking'' in other language(s). \citet{sensitivity_prompt_design,impact_demonstration_multilingual_icl} highlight that models are sensitive to those templates.
For languages of demonstrations and test questions, \citet{mgsm,mega} conclude that in-language demonstrations outperform English demonstrations. \citet{do_llm_think_better_in_english,is_translation_all_you_need} suggest translating questions from LRLs into English can improve performance. 

\section{Conclusion and Discussion}
This work systematically analyzes multiple ICL strategies for MLLMs, and confirms that the presence of multiple languages is an effective strategy across multiple MLLMs.
This improvement is partially due to the inclusion of non-English languages in the prompting, and partially due to the in-topic demonstrations in non-English languages, which together strengthen the models' cross-lingual transfer capabilities, particularly the capability to process LRLs.
Our work echoes with \citet{revisit_primacy_of_english}---who suggest that HRLs other than English excel in the pretraining-finetuning framework---in the in-context learning framework, highlighting the importance of language inclusivity.

We are in agreement with \citet{nllb} and \citet{is_translation_all_you_need} that an ideal language-universal LLM should be equally capable in all languages. 
Beyond this belief, we found that non-English languages may better elicit the potential of MLLMs.
Although these observations remain in using HRLs for LRL processing, our results strongly support the call for greater research investment in enhancing MLLM capabilities for a broader range of languages.

\section*{Limitations}
This paper treats multilingual LLMs as black-box models, drawing the findings and conclusions based solely on their input-output behavior. Hence, we have not interpreted the internal mechanism of how multilingualism could affect MLLM's ``thinking'' process and its manifested performance. We have briefly touched on the impact of demonstrations in different languages on the MLLM performance. However, we do not conduct a thorough empirical analysis to identify which specific linguistic characteristics (e.g., writing systems, grammatical structures, or linguistic relatedness) contribute to the observed performance differences.

% \section*{Acknowledgment}
% We thank xxx. \yilei{to be filled...}

% \section*{Reproducibility}

\bibliography{custom}

\clearpage
\appendix

\section{Experiment Setup} \label{app:setup}
\subsection{Model} \label{app:model}
\begin{table*}[!htbp]
    \small
    \centering
    \alternaterowcolors
    \begin{tabular}{lcccl}
        \toprule
        \bfseries Model Name & \bfseries  Scale & \bfseries Instruct? & \bfseries Open-Source? & \bfseries Checkpoint \\
        \midrule
        \llamaThree & 8B & \faCheck & \faCheck & \href{https://huggingface.co/meta-llama/Meta-Llama-3-8B-Instruct}{meta-llama/Meta-Llama-3-8B-Instruct} \\
        \llamaThreeOne & 8B & \faCheck & \faCheck & \href{https://huggingface.co/meta-llama/Meta-Llama-3-8B-Instruct}{meta-llama/Meta-Llama-3.1-8B-Instruct} \\
        \qwenTwo & 7B & \faCheck & \faCheck & \href{https://huggingface.co/Qwen/Qwen2-7B-Instruct}{Qwen/Qwen2-7B-Instruct} \\
        \qwenTwoFive & 7B & \faCheck & \faCheck & \href{https://huggingface.co/Qwen/Qwen2.5-7B-Instruct}{Qwen/Qwen2.5-7B-Instruct} \\
        \mistral & 12B & \faCheck & \faCheck & \href{https://huggingface.co/mistralai/Mistral-Nemo-Instruct-2407}{mistralai/Mistral-Nemo-Instruct-2407} \\
        \aya & 8B & \faCheck & \faCheck & \href{https://huggingface.co/CohereForAI/aya-expanse-8b}{CohereForAI/aya-expanse-8b} \\
        \gptThreeFive & NA & \faCheck & \faTimes & \href{https://platform.openai.com/docs/models/gpt-4o#gpt-3-5-turbo}{gpt-3.5-turbo-0125} \\
        % \gptFour & NA & \faCheck & \faTimes & \href{https://platform.openai.com/docs/models/gpt-4o#gpt-4-turbo-and-gpt-4}{gpt-4-turbo-2024-04-09} \\
        \gptFourO & NA & \faCheck & \faTimes & \href{https://platform.openai.com/docs/models/gpt-4o#gpt-4o-mini}{gpt-4o-mini-2024-07-18} \\
        \bottomrule
    \end{tabular}
    \caption{Model details. Checkpoints are either from Hugging Face or OpenAI API.}
    \label{tab:model_card}
\end{table*}
All model checkpoints we use and their properties are listed in \cref{tab:model_card}.

\subsection{Dataset} \label{app:dataset}
\begin{table*}[!htbp]
    %\resizebox{\columnwidth}{!}{
    \small
    \centering
    \alternaterowcolors
    \setlength{\tabcolsep}{6pt}
    \begin{tabular}{llllr}
        \toprule
        \bfseries Dataset & \bfseries HRL & \bfseries LRL & \bfseries Source \\
        \midrule
        \mgsm & de, en, es, fr, ja, ru, zh & bn, sw, te, th & \href{https://huggingface.co/datasets/juletxara/mgsm}{juletxara/mgsm} \\
        \xcopa & en, id, it, tr, zh & et, ht, 
        qu, sw, ta, th, vi & \href{https://autonlp.ai/datasets/choice-of-plausible-alternatives-(copa)}{English COPA} \& \href{https://huggingface.co/datasets/cambridgeltl/xcopa}{cambridgeltl/xcopa}   \\
        \xlwic & da, de, en, fr, it, ja, ko, nl, zh & bg, et, fa, hr& \href{https://pilehvar.github.io/xlwic/}{pilehvar.github.io/xlwic/} \\
        \xquad & ar, de, en, es, ru, tr, vi, zh & el, hi, ro, th & \href{https://huggingface.co/datasets/google/xquad}{google/xquad} \\
        
        \bottomrule
    \end{tabular}
    %}
    \caption{Additional information for the three datasets we evaluate. The correspondence between language codes and names can be found in \cref{tab:lang25}. ``Source'' indicates where to download the dataset.}
    \label{tab:dataset_additional_info}
\end{table*}
After preprocessing, datapoints for each language split are stored in a single JSON file. \cref{tab:dataset_additional_info} summarizes the supported languages for each dataset and the sources from which they are obtained.

\begin{figure}[!ht]
    \centering
    \includegraphics[width=\linewidth]{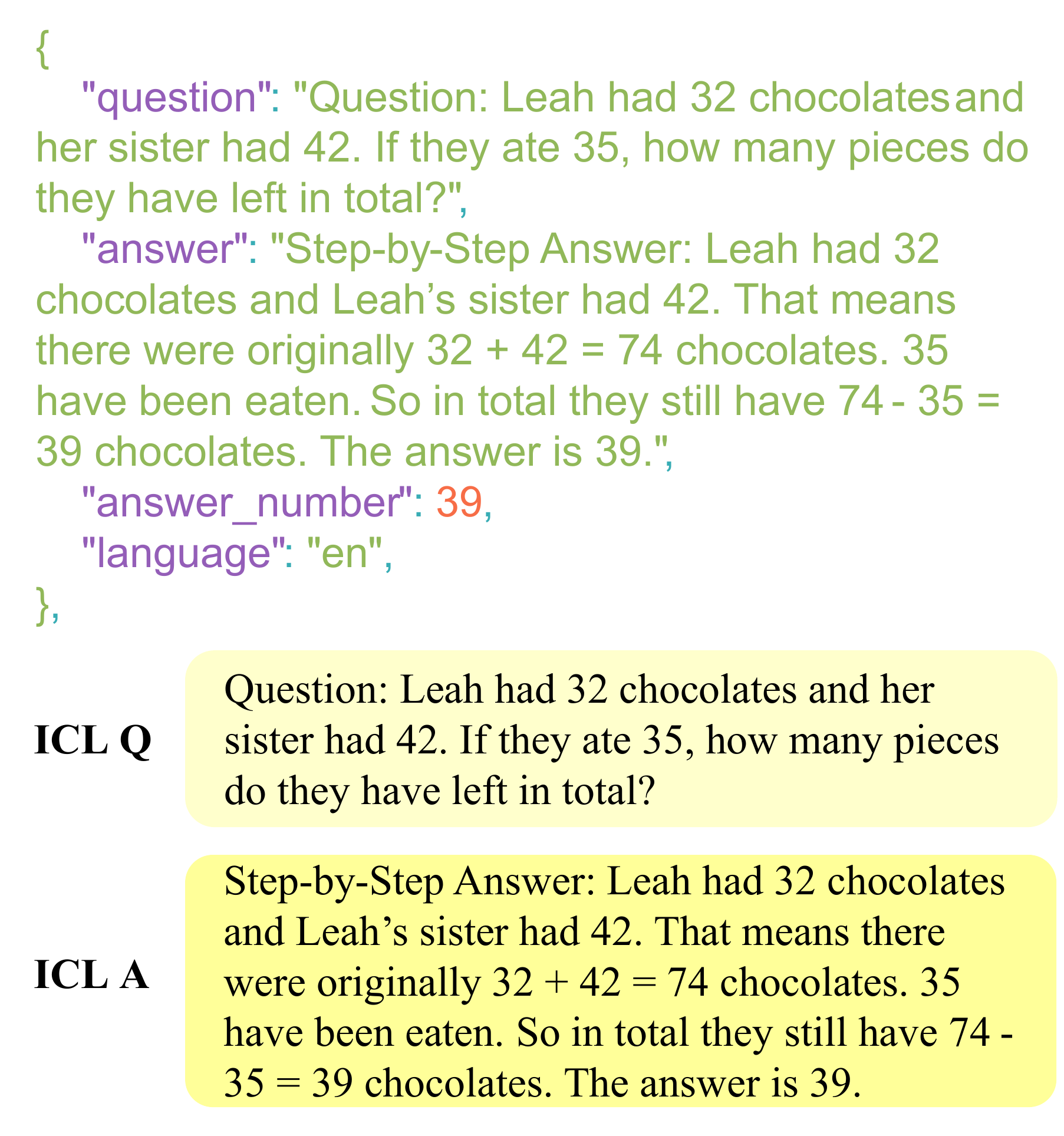}
    \caption{An example of an English datapoint from \mgsm training set. When calling Chat Template API, \user role message is the ``question'' value, while \assistant role message is the ``answer'' value. Note that in the test set, the answer is null without exemplar CoT response. The correct numerical answer is stored in ``answer\_number''.}
    \label{fig:template:mgsm}
\end{figure}

\paragraph{\mgsm} The original dataset consists of parallel datapoints across all language splits and training/test splits of the same size. Example datapoint and the Chat Template for few-shot demonstrations can be found in \cref{fig:template:mgsm}.

\begin{figure}[!t]
    \centering
    \begin{subfigure}[b]{\columnwidth}
        \centering
        \includegraphics[width=\linewidth]{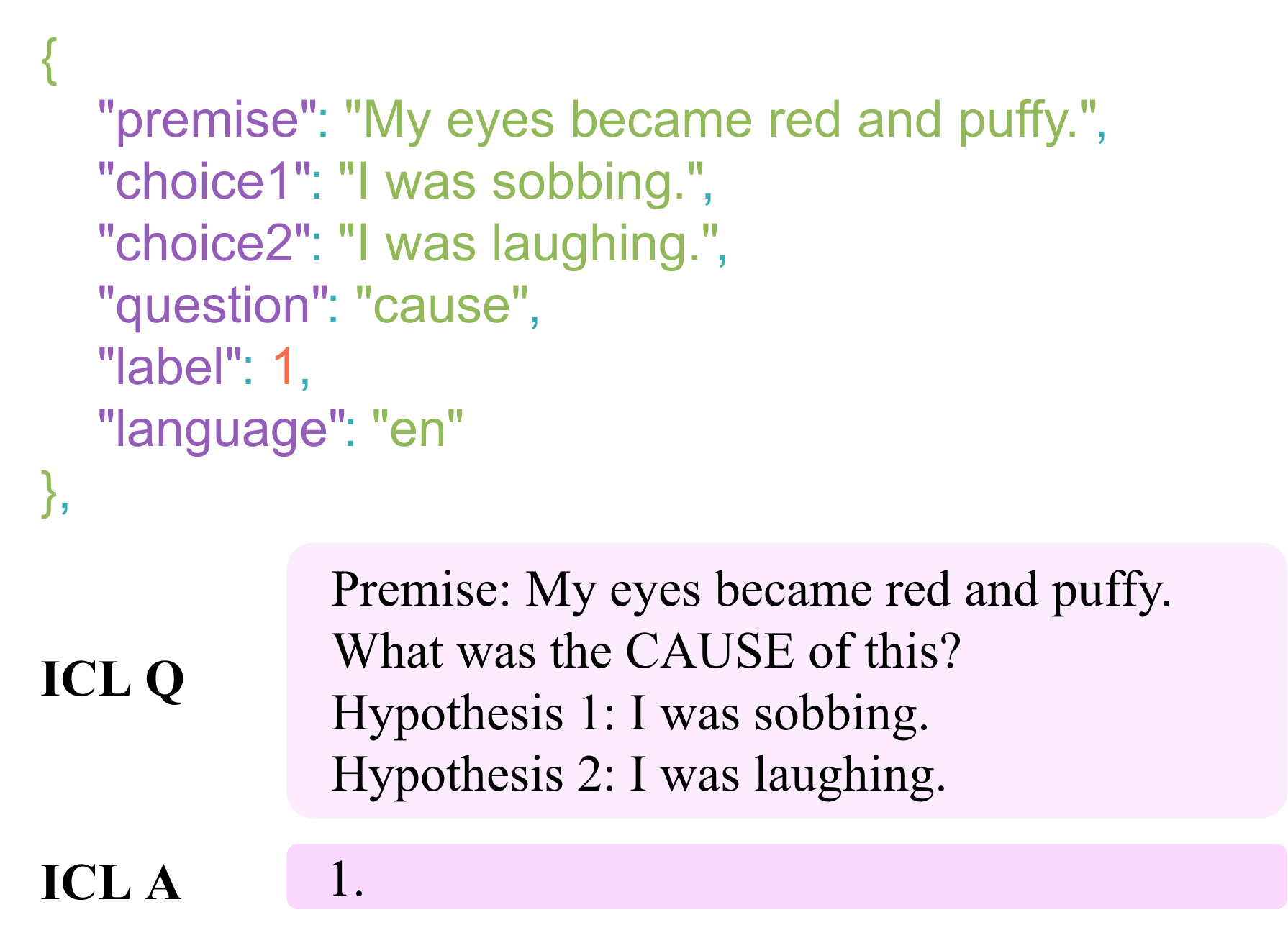}
        \captionsetup{skip=1pt}
        \caption{An example of ``cause'' datapoint.}
        \label{fig:template:xcopa:cause}
    \end{subfigure}

    \begin{subfigure}[b]{\columnwidth}
        \centering
        \includegraphics[width=\linewidth]{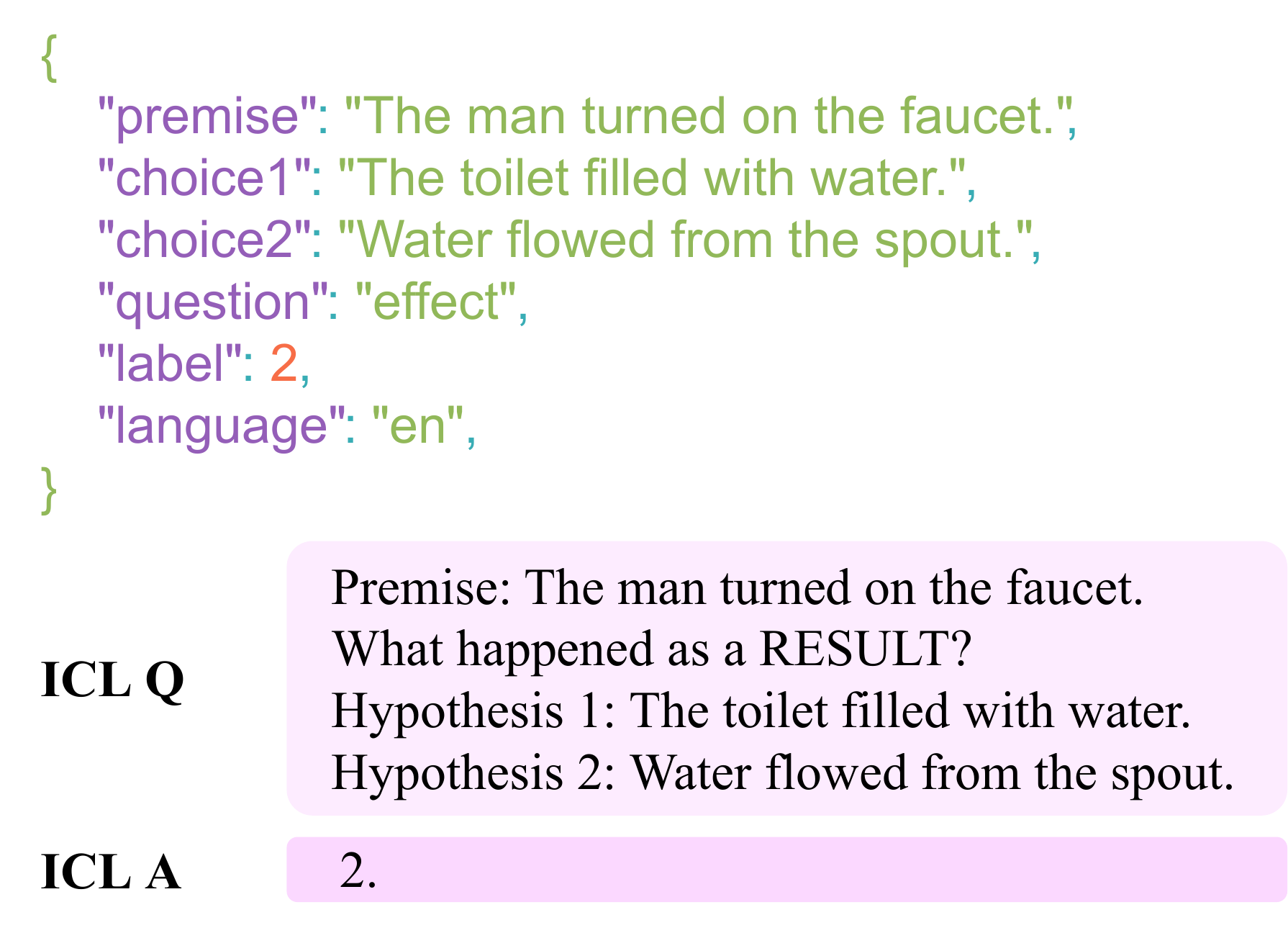}
        \captionsetup{skip=1pt}
        \caption{An example of ``effect'' datapoint.}
        \label{fig:template:xcopa:effect}
    \end{subfigure}

    % \captionsetup{skip=1pt}
    \caption{Examples of English datapoints from \xcopa training set. First, based on whether ``question'' is  ``cause'' or ``effect'', we fill the ``premise'', ``choice1'', and ``choice2'' values into one of the two predefined templates. The template's language is changeable as per the language split of the datapoint. Then we call the Chat Template API, \user role message is the filled template, while \assistant role message is the ``label'' value.}
    \label{fig:template:xcopa}
\end{figure}

\paragraph{\xcopa} The $100$ datapoints in the training split of XCOPA are parallel to the last 100 datapoints in the English COPA development split. Therefore, we exclude the first $400$ datapoints from the development split of English COPA. The test splits of both datasets are parallel and contain the same number of datapoints. We then merge both into our \xcopa dataset. In \xcopa, there are two types of questions: ``cause'' and ``effect'', each corresponding to a distinct template, as shown in \cref{fig:template:xcopa:cause,fig:template:xcopa:effect}.

\begin{figure}[!t]
    \centering
    \includegraphics[width=\linewidth]{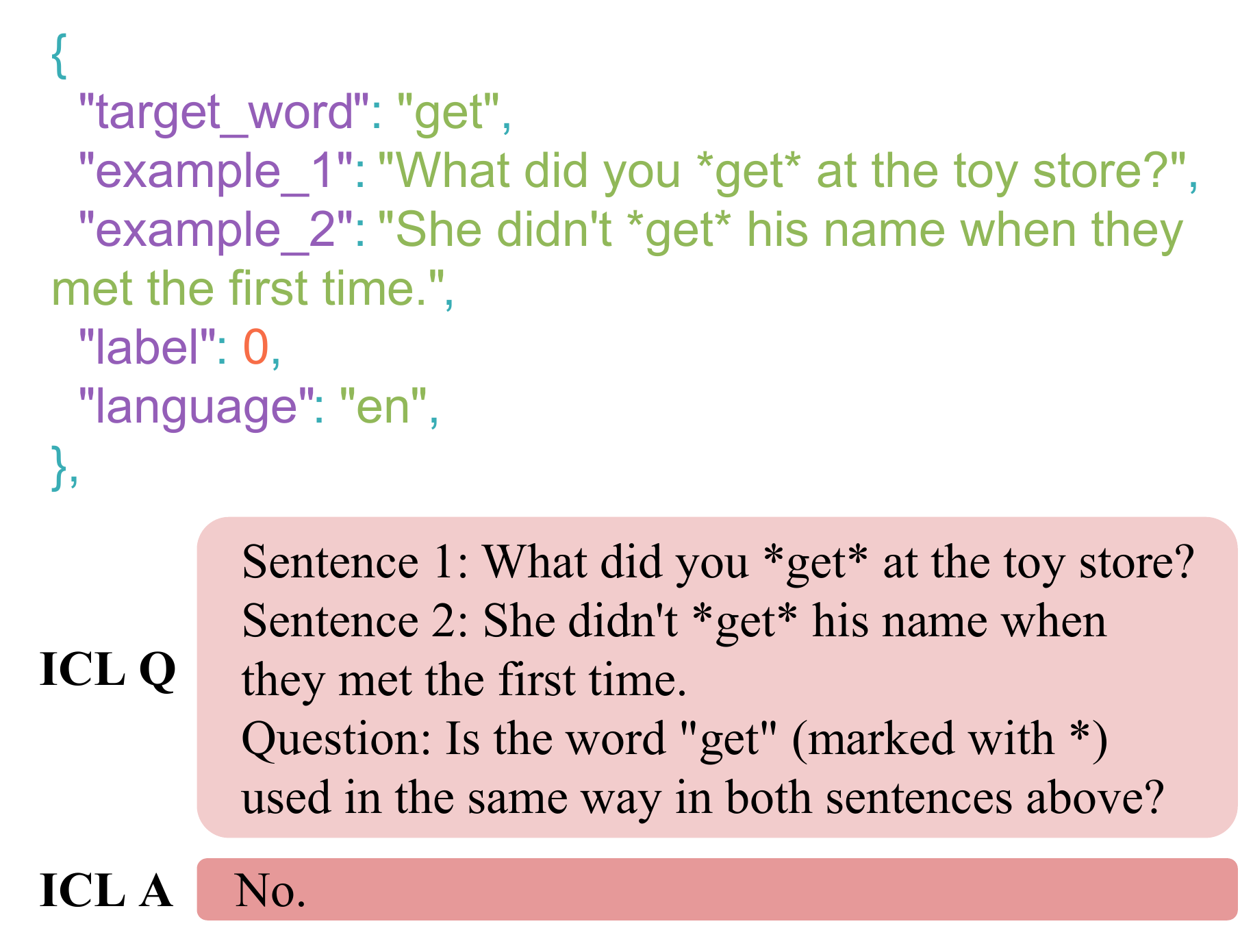}
    \caption{An example of English datapoint from \xlwic training set. We first fill the ``example\_1'', ``example\_2'', and ``target\_word'' values into the predefined templates. Asterisks * are surrounded around ``target\_word'' to draw the LLM's attention. The template's language is changeable as per the language split of the datapoint. Then we call the Chat Template API, \user role message is the filled template, while \assistant role message is ``Yes'' or ``No'' (``label'' is $1$ or $0$).}
    \label{fig:template:xlwic}
\end{figure}

\paragraph{\xlwic} This benchmark is designed to determine whether a specific word in a given language has the same meaning in two different sentences. As a result, the dataset is inherently non-parallel. Among all language splits, Estonian (et) contains the fewest datapoints, with $98$ in the training split and $390$ in the test split. For all other languages, we randomly subsample to match the size of the Estonian split to satisfy the demonstration sampling requirements outlined in \cref{sec:icl:multilingual_mode}. To leverage the attention mechanism of transformers \cite{transformer}, we add asterisks around the target word in both sentences to indicate that the LLM needs to disambiguate the meaning of that specific word. An example datapoint is shown in \cref{fig:template:xlwic}.

\paragraph{\combined} \label{app:dataset:combined} Identifying specific neurons depends on the nature of the input corpus. Since language is inherently conjugate with the task, and our focus is on language-specific neurons rather than task-specific neurons, it is necessary to input all three datasets into the LLM to eliminate the confounding factor of the task domain. To balance the three datasets, we subsample their test splits to only $250$ datapoints each. To balance the number of language splits across datasets, we excluded two HRL splits, Korean (ko) and Dutch (nl), from the \xlwic dataset. This ensures that all three (sub-)datasets have $11$ languages for combination. Additionally, for \mgsm, we limit the answers to only include the final numeric result without the CoT reasoning. This approach ensures that the total number of datapoints and the overall token count are roughly the same across the original three datasets.

\begin{figure}[!t]
    \centering
    \includegraphics[width=\linewidth]{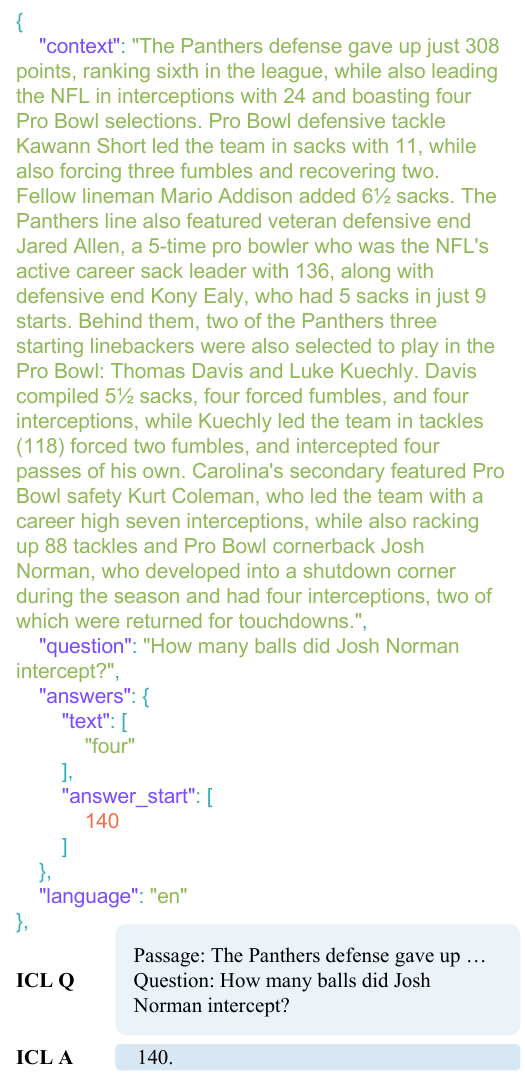}
    \caption{An example of English datapoint from \xquad validation set. We fill the ``context'' and ``question'' values into the predefined templates. The template's language is changeable as per the language split of the datapoint. Then we call the Chat Template API, \user role message is the filled template, while \assistant role message is ``answers['text']''.}
    \label{fig:template:xquad}
\end{figure}
 
\paragraph{\xquad} The dataset consists of parallel question-answering examples across $12$ languages, derived from the same subset of the English SQuAD \cite{squad} v1.1 development set. For each language, the context passages, questions, and answers are professionally translated while preserving semantic alignment across languages. Each example includes a context paragraph, a question, and a span-based answer. An example and the input prompt format used in our experiments are illustrated in \cref{fig:template:xquad}.

\begin{figure}[!t]
    \centering
    \includegraphics[width=\linewidth]{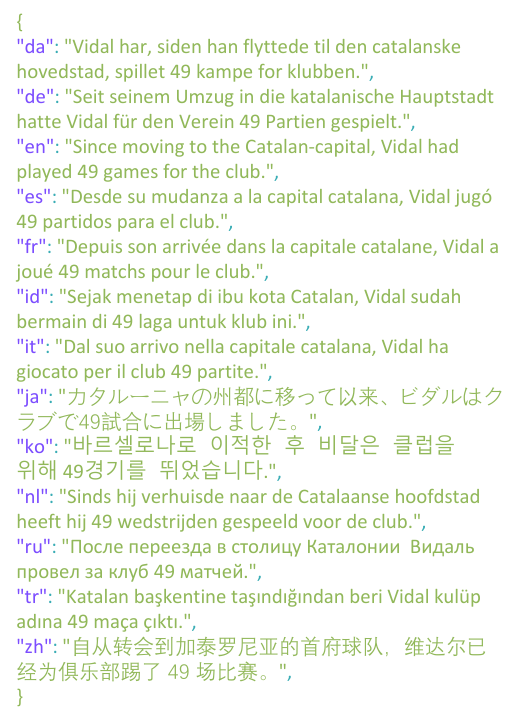}
    \captionsetup{skip=1pt}
    \caption{A datapoint example from \flores of semantic-equivalent context-irrelevant sentences in all high resource languages we study in this work.}
    \label{fig:flores}
\end{figure}

\paragraph{\flores} \label{app:dataset:flores} This machine translation benchmark comprises $3,001$ sentences extracted from English Wikipedia, spanning diverse topics and domains. These sentences were translated into $101$ languages by professional translators via a carefully controlled process. The Wikipedia domain is largely unrelated to the domains of the three datasets we evaluate (math, commonsense reasoning, and word disambiguation). Therefore, we choose \flores as our source pool of irrelevant sentences. Note that in this benchmark, each datapoint carries the same semantic meaning across all $101$ language splits. We do not want irrelevant sentences to affect the LLM's understanding of the original task excessively, thus we select sentences with word counts between $10$ and $15$ in the English split, introducing limited noise. \cis are drawn form the filtered \flores dataset. An example is provided in \cref{fig:flores}.

\subsection{Language} \label{app:lang}
\subsubsection{High-Resource Language List} \label{app:lang:hrl_list}
\begin{table}[!htbp]
    \small
    \centering
    \alternaterowcolors
    \begin{tabular}{llr}
        \toprule
            \bfseries Code & \bfseries Language & \bfseries Percentage \\
        \midrule
            en & English & $89.70 \%$ \\
            de & German & $0.17 \%$ \\ 
            fr & French & $0.16 \%$ \\
            sv & Swedish & $0.15 \%$ \\
            zh & Chinese & $0.13 \%$ \\
            es & Spanish & $0.13 \%$ \\
            ru & Russian & $0.13 \%$ \\
            nl & Dutch & $0.12 \%$ \\
            it & Italian & $0.11 \%$ \\
            ja & Japanese & $0.10 \%$ \\
            pl & Polish & $0.09 \%$ \\
            pt & Portuguese & $0.09 \%$ \\
            vi & Vietnamese & $0.08 \%$ \\
            uk & Ukrainian & $0.07 \%$ \\
            ko & Korean & $0.06 \%$ \\
            ca & Catalan & $0.04 \%$ \\
            sr & Serbian & $0.04 \%$ \\
            id & Indonesian & $0.03 \%$ \\
            cs & Czech & $0.03 \%$ \\
            fi & Finnish & $0.03 \%$ \\
        \bottomrule
    \end{tabular}
    \caption{Top 20 language distribution of the training data for \llamaTwo, excluding code and unknown data. Adopted from Table 10 in \citet{llama2}.}
    \vspace{-.2cm}
    \label{tab:llama2_top_20_lang}
\end{table}

\begin{table}[!htbp]
    \small
    \centering
    \alternaterowcolors
    \begin{tabular}{llrr}
        \toprule
        \bfseries Code & \bfseries Language & \bfseries Tokens (B) & \bfseries Percentage \\
        \midrule
            en & English & 578.064 & $77.984 \%$ \\
            de & German & 25.954 & $3.501 \%$ \\
            fr & French & 24.094 & $3.250 \%$ \\
            es & Spanish & 15.654 & $2.112 \%$ \\
            pl & Polish & 10.764 & $1.452 \%$ \\
            it & Italian & 9.699 & $1.308 \%$ \\
            nl & Dutch & 7.690 & $1.037 \%$ \\
            sv & Swedish & 5.218 & $0.704 \%$ \\
            tr & Turkish & 4.855 & $0.655 \%$ \\
            pt & Portuguese & 4.701 & $0.634 \%$ \\
            ru & Russian & 3.932 & $0.530 \%$ \\
            fi & Finnish & 3.101 & $0.418 \%$ \\
            cs & Czech & 2.991 & $0.404 \%$ \\
            zh & Chinese & 2.977 & $0.402 \%$ \\
            ja & Japanese & 2.832 & $0.382 \%$ \\
            no & Norwegian & 2.695 & $0.364 \%$ \\
            ko & Korean & 1.444 & $0.195 \%$ \\
            da & Danish & 1.387 & $0.187 \%$ \\
            id & Indonesian & 1.175 & $0.159 \%$ \\
            ar & Arabic & 1.091 & $0.147\%$ \\
        \bottomrule
    \end{tabular}
    \caption{Top 20 language distribution of the training data for \palm, excluding code and unknown data. Adopted from Table 28 in \citet{palm}.}
    % \vspace{-.2cm}
    \label{tab:palm_top_20_lang}
\end{table}

To the best of our knowledge, we find two multilingual LLMs—\llamaTwo \cite{llama2} and \palm \cite{palm}—that publicly report the language distribution used during pretraining. The top 20 languages and their percentages for each model are listed in \cref{tab:llama2_top_20_lang} and \cref{tab:palm_top_20_lang}, respectively. We take the union of these languages to form what we consider a \textit{high-resource language list} (in terms of the richness in the LLM pretraining dataset), including the following $24$ languages:
\begin{equation}
    \begin{aligned}
        \mathcal{L}_{H} = & \{ \text{ar, ca, cs, da, de, en, es, fi, fr, id, it, ja, } \\
                          & \text{ko, nl, no, pl, pt, ru, sr, sv, tr, uk, vi, zh} \}.
    \end{aligned}
    \label{eq:high_lang_list}
\end{equation}

The high overlap between the top languages of the two LLMs further supports the rationale for applying this list to other LLMs.

\subsubsection{Languages We Evaluate}
\cref{tab:lang25} presents the union of languages supported by the datasets we evaluated (\cref{sec:setup:datasets}). Based on whether a language appears in the high-resource language list, we categorized the $29$ languages into  HRL and LRL groups, with $14$ classified as HRL and $15$ as LRL. Among them, only English and Chinese are present in all three datasets.

It is worth highlighting that the $11$ LRLs span $7$ distinct writing systems and $6$ language families. This diversity suggests that when tokenizing inputs in these LRLs, they are unlikely to share common tokens with HRLs ($9$ out of $13$ use the Latin script). Consequently, this limits the MLLMs' ability to leverage shared subwords or similar syntax structures for cross-lingual transfer across LRLs.
\begin{table*}[!htbp]
    \small
    \centering
    \alternaterowcolors
    \begin{tabular}{llclll}
        \toprule
        \bfseries Code & \bfseries Name in English & \bfseries HRL/LRL & \bfseries In Which Dataset(s) & \bfseries Writing System & \bfseries Family\\
        \midrule
        ar & Arabic & H &\xquad & Arabic & Afro-Asiatic \\
        bg & Bulgarian & L & \xlwic & Cyrillic & Indo-European \\
        bn & Bengali & L & \mgsm & Bengali–Assamese & Indo-European \\
        da & Danish & H & \xlwic & Latin & Indo-European \\
        de & German & H & \mgsm, \xlwic, \xquad & Latin & Indo-European \\
        el & Greek & L & \xquad & Greek & Indo-European \\
        en & English & H & \mgsm, \xcopa, \xlwic \xquad  & Latin & Indo-European \\
        es & Spanish & H & \mgsm, \xquad & Latin & Indo-European \\
        et & Estonian & L & \xcopa & Latin & Indo-European \\
        fa & Persian & L & \xlwic & Arabic & Indo-European \\
        fr & French & H & \mgsm, \xlwic & Latin & Indo-European \\
        hi & Hindi & L & \xquad & Devanagari & Indo-European \\
        hr & Croatian & L & \xlwic & Latin & Indo-European \\
        ht & Haitian & L & \xcopa & Latin & French Creole \\
        id & Indonesian & H & \xcopa & Latin & Austronesian \\
        it & Italian & H & \xlwic, \xcopa & Latin & Indo-European \\
        ja & Japanese & H & \mgsm, \xlwic & Kana \& Hanzi & Japonic \\
        ko & Korean & H & \xlwic & Hangul & Koreanic \\
        nl & Dutch & H & \xlwic & Latin & Indo-European \\
        qu & Quechua & L & \xcopa & Latin & Quechumaran \\
        ro & Romanian & L & \xquad & Latin & Indo-European \\
        ru & Russian & H & \mgsm, \xquad & Cyrillic & Indo-European \\
        sw & Swahili & L & \mgsm & Latin & Niger–Congo \\
        ta & Tamil & L & \xcopa & Tamil & Dravidian \\
        te & Telegu & L & \mgsm & Telegu & Dravidian \\
        th & Thai & L & \mgsm, \xcopa, \xquad & Thai & Kra–Dai \\
        tr & Turkish & H & \xcopa, \xquad & Latin & 	Turkic \\
        vi & Vietnamese & L & \xcopa, \xquad & Latin & Austroasiatic \\
        zh & Chinese & H & \mgsm, \xcopa, \xlwic \xquad & Hanzi & Sino-Tibetan \\
        \bottomrule
    \end{tabular}
    \caption{List of $29$ languages we study and their properties, in ascending order of ISO 639-1 codes \cite{iso639}.}
    \vspace{-.2cm}
    \label{tab:lang25}
\end{table*}

\section{Experiment Raw Results} \label{app:expt}
\subsection{ICL Modes Evaluation} \label{app:expt:vanilla}

The raw data for vanilla evaluation is recorded in \cref{tab:vanilla_eval:mgsm,tab:vanilla_eval:xcopa,tab:vanilla_eval:xlwic,tab:vanilla_eval:xquad}. McNemar's test results for ICL modes are in \cref{tab:hyp_test:vanilla_eval:mgsm,tab:hyp_test:vanilla_eval:xcopa,tab:hyp_test:vanilla_eval:xlwic,tab:hyp_test:vanilla_eval:xquad}. Significance (sig.) levels -- *: $p<0.05$; **: $p<0.01$; ***: $p<0.001$.

\subsection{Context-Irrelevant Sentence} \label{app:expt:cis}

The raw data for \cis is recorded in \cref{tab:cis:mgsm,tab:cis:xcopa,tab:cis:xlwic,tab:cis:xquad}. McNemar's test results for \cis modes are in \cref{tab:hyp_test:cis:mgsm,tab:hyp_test:cis:xcopa,tab:hyp_test:cis:xlwic,tab:hyp_test:cis:xquad}.

\section{ICL Behavioral Analysis} \label{app:neuron}
\subsection{Specilized Neuron}
Inspired by the universal concept space \cite{llama_english_epfl, neuron_plnd, semantic_hub}, we hypothesize that MLLMs could activate more cross-lingual capabilities by aligning different linguistic representations. To validate the above hypothesis, we seek to find patterns in neuron behavior between ICL modes. Following \citet{neuron_specialization_translation,language_specific_neuron_msra,language_specific_neuron_tokyo,neuron_pim,neuron_plnd,neuron_sft}, we look at the activations of neurons in the multilayer perceptron (MLP, or \textit{Feedforward Network, FFN}) modules of the MLLMs.

\subsubsection{Identifying Top-Activated Neurons}
Each neuron in every MLP layer of the model is assigned a dedicated counter, initialized to $0$. During vanilla evaluation, we monitor the activation of every neuron in the forward pass. Since LLMs typically employ ReLU-like \cite{relu} activation functions (e.g., SwiGLU \cite{glu} for \textsf{Llama} series), a positive activation value can be interpreted as the neuron being ``activated''. If a neuron is ``activated'', we increment the corresponding counter by $1$, otherwise no action. To ensure balanced input across our three datasets, we curated a \combined dataset, see \cref{app:dataset} for details. After processing the inputs of a single ICL mode, each neuron accumulates an  ``activated'' count. The neurons with the highest counts are identified as specialized neurons attributed to this ICL mode.

We employ two methods for selecting the most activated neurons. Top-$k$ selects neurons in the top $k$ percentile \cite{language_specific_neuron_msra}, while top-$p$ selects neurons progressively until their cumulative activation counts reach $p\ (\%)$ of the sum of all activation values \cite{neuron_specialization_translation}.

\subsection{\multilingual-specific Neurons Overlap Most with \native-specific Neurons}
\begin{table}[!htbp]
\setlength{\tabcolsep}{4pt}
    \small
    \centering
    \alternaterowcolors
\begin{tabular}{lccc}
\toprule
\midrule
\bfseries IoU $(\%)$&
\textbf{All Langs} &
\textbf{LRL} &
\textbf{HRL} \\
\midrule

\multicolumn{4}{l}{\textbf{\english -- \multilingual -- \native}} \\
\midrule

\multicolumn{4}{l}{\textbf{\llamaThreeOne}} \\
\native - \english & $\dynamicCellColor{61.82}$ & $\dynamicCellColor{56.81}$ & $\dynamicCellColor{68.50}$ \\
\native - \multilingual & $\dynamicCellColor{78.84}$ & $\dynamicCellColor{66.19}$ & $\dynamicCellColor{85.10}$ \\
\english - \multilingual & $\dynamicCellColor{65.84}$ & $\dynamicCellColor{66.89}$ & $\dynamicCellColor{64.60}$ \\

\multicolumn{4}{l}{\textbf{\qwenTwo}} \\
\native - \english & $\dynamicCellColor{70.61}$ & $\dynamicCellColor{66.05}$ & $\dynamicCellColor{81.22}$ \\
\native - \multilingual & $\dynamicCellColor{85.13}$ & $\dynamicCellColor{77.40}$ & $\dynamicCellColor{91.31}$ \\
\english - \multilingual & $\dynamicCellColor{78.24}$ & $\dynamicCellColor{79.83}$ & $\dynamicCellColor{77.46}$ \\

\midrule
\multicolumn{4}{l}{\textbf{\english -- \chinese -- \native}} \\
\midrule

\multicolumn{4}{l}{\textbf{\llamaThreeOne}} \\
\native - \english & $\dynamicCellColor{61.82}$ & $\dynamicCellColor{56.81}$ & $\dynamicCellColor{68.50}$ \\
\native - \chinese & $\dynamicCellColor{60.58}$ & $\dynamicCellColor{55.35}$ & $\dynamicCellColor{65.73}$ \\
\english - \chinese & $\dynamicCellColor{56.52}$ & $\dynamicCellColor{57.94}$ & $\dynamicCellColor{55.72}$ \\

\multicolumn{4}{l}{\textbf{\qwenTwo}} \\
\native - \english & $\dynamicCellColor{70.61}$ & $\dynamicCellColor{66.05}$ & $\dynamicCellColor{81.22}$ \\
\native - \chinese & $\dynamicCellColor{73.41}$ & $\dynamicCellColor{69.39}$ & $\dynamicCellColor{82.18}$ \\
\english - \chinese & $\dynamicCellColor{77.07}$ & $\dynamicCellColor{78.34}$ & $\dynamicCellColor{76.51}$ \\

\midrule
\bottomrule
\end{tabular}
    \caption{The IoU score of most-activated neurons between every pair of ICL modes in triplets \english -- \multilingual -- \native and \english -- \chinese -- \native. Neurons were selected by first filtering out neurons outside the first 8 and last 8 MLP layers, and applying top-$k$ method with $k = 0.7$.}
    \vspace{-.2cm}
    \label{tab:neuron_statistics}
\end{table}

We examined the overlaps among the most-activated neurons (\textit{specialized neurons}) across different ICL modes by calculating the Intersection over Union (IoU) scores. For ICL modes $M_1$ and $M_2$, with specialized neurons denoted as sets $S_1$ and $S_2$, their overlap is quantified by \cref{eq:iou}:

\begin{equation}
    \begin{aligned}
    \operatorname{IoU}\left(S_1, S_2\right)=\frac{\left|S_1 \cap S_2\right|}{\left|S_1 \cup S_2\right|}.
    \end{aligned}
    \label{eq:iou}
\end{equation}

Prior research \cite{language_specific_neuron_msra,neuron_plnd} has discovered that \textit{language-specific neurons} are located primarily in the models' top and bottom layers. Because we want to explain the multilingual reasoning capabilities of a model, we only consider neurons that belong to a certain prefix or suffix of the models' layers in an effort to restrict our selected neuron sets to be mainly language-specific neurons. 

The similarity in performance between \multilingual and \native can be explained by the high overlap in the sets of most-activated neurons. On the other hand, the poorer performance of \english can be explained by the low overlap between \english most-activated neurons and other ICL modes . 

The same experiment was also performed but with  \multilingual replaced by HRL \chinese. In this case, the patterns were less obvious and model-specific. This result aligns with our findings in vanilla evaluation that \english $\le$ Non-\english HRL \monolingual $\le$ \multilingual (\cref{sec:mono_vs_multilingual_results}).

We further split this neuron experiment to either use only LRL or HRL ICL modes when recording neuron activations. We observed that HRL tends to activate similar neurons between \native and \multilingual, whereas LRL tends to activate similar neurons between \english and \multilingual.

\begin{table*}[!htbp]
    \setlength{\tabcolsep}{3.7pt}
    \scriptsize
    \centering
    \alternaterowcolors
\begin{tabular}{l|ccccccccccc|lll}
\toprule
    \textbf{\mgsm} &
  \textbf{\underline{bn}} &
  \textbf{de} &
  \textbf{en} &
  \textbf{es} &
  \textbf{fr} &
  \textbf{ja} &
  \textbf{ru} &
  \textbf{\underline{sw}} &
  \textbf{\underline{te}} &
  \textbf{\underline{th}} &
  \textbf{zh} &
  \textbf{\underline{LRL Avg}} &
  \textbf{HRL Avg} &
  \textbf{ALL Avg} \\ 
  \midrule

\multicolumn{15}{l}{\textbf{\llamaThree}} \\
\english      & 66.40 & 76.40 & 86.40 & 78.80 & 77.60 & 66.40 & 77.20 & 59.20 & 60.00 & 71.20 & 75.60 & 64.20     & 76.91      & 72.29     \\
\french       & 67.60 & 77.60 & 86.00 & 77.60 & 75.20 & 70.00 & 79.60 & 56.40 & 59.60 & 68.00 & 70.40 & \decrease{62.90}{1.30} & \decrease{76.63}{0.28} & \decrease{71.64}{0.65} \\
\chinese      & 66.80 & 77.20 & 82.80 & 79.60 & 74.80 & 68.00 & 76.40 & 55.60 & 59.20 & 70.40 & 72.80 & \decrease{63.00}{1.20} & \decrease{75.94}{0.97} & \decrease{71.24}{1.05} \\
\japanese     & 66.80 & 78.80 & 83.20 & 76.00 & 74.80 & 70.80 & 76.40 & 56.40 & 56.80 & 68.80 & 68.00 & \decrease{62.20}{2.00} & \decrease{75.43}{1.48} & \decrease{70.62}{1.67} \\
\multilingual & 65.20 & 79.60 & 84.40 & 77.60 & 74.00 & 69.60 & 76.80 & 52.40 & 53.60 & 69.60 & 70.00 & \decrease{60.20}{4.00} & \decrease{76.00}{0.91} & \decrease{70.25}{2.04} \\
\native       & 64.80 & 76.80 & 86.40 & 80.00 & 75.20 & 70.80 & 76.80 & 58.40 & 54.80 & 70.40 & 72.80 & \decrease{62.10}{2.10} & \increase{76.97}{0.06}  & \decrease{71.56}{0.73} \\ 
\transEn & 63.60	 & 73.20	 &86.40	 &78.00	 &74.00	 &60.40	 &77.60	 &69.60	 &60.00	 &46.40	 &58.80	 &\decrease{59.90}{4.30}	 &\decrease{72.63}{4.28}	 &\decrease{68.00}{4.29} \\
\transSource &65.20	&78.80	&86.40	&79.60	&76.00	&68.40	&76.40	&58.80	&52.80	&69.20	&73.60	&\decrease{61.50}{2.70}	&\increase{77.03}{0.12}	&\decrease{71.38}{0.91}\\

\midrule

\multicolumn{15}{l}{\textbf{\llamaThreeOne}} \\
\english      & 52.40 & 69.20 & 87.20 & 63.60 & 76.80 & 68.00 & 78.80 & 67.20 & 39.60 & 69.20 & 75.20 & 57.10 & 74.11 & 67.93 \\
\french & 62.80 & 74.40 & 89.20 & 82.40 & 82.40 & 69.60 & 79.20 & 69.60 & 45.20 & 70.00 & 76.80 &  \increase{61.90}{4.80} &  \increase{79.14}{5.03} &  \increase{72.87}{4.94} \\
\chinese & 66.00 & 76.80 & 87.20 & 83.20 & 78.00 & 67.60 & 79.60 & 70.00 & 48.40 & 69.20 & 77.60 &  \increase{63.40}{6.30} &  \increase{78.57}{4.46} &  \increase{73.05}{5.12} \\
\japanese & 66.40 & 76.80 & 86.00 & 80.80 & 78.00 & 67.20 & 79.60 & 72.80 & 58.00 & 75.20 & 77.60 &  \increase{68.10}{11.00} &  \increase{78.00}{3.89} &  \increase{74.40}{6.47} \\
\multilingual & 68.00 & 76.40 & 88.00 & 84.80 & 80.40 & 69.20 & 80.40 & 68.80 & 55.60 & 71.60 & 75.20 &  \increase{66.00}{8.90} &  \increase{79.20}{5.09} &  \increase{74.40}{6.47} \\
\native & 67.20 & 80.40 & 87.20 & 81.20 & 82.40 & 67.20 & 80.00 & 72.40 & 58.00 & 76.40 & 77.60 &  \increase{68.50}{11.40} &  \increase{79.43}{5.32} &  \increase{75.45}{7.52} \\
\transEn & 62.80	&74.80	&87.20	&82.00	&77.60	&62.80	&81.60	&68.00	&62.00	&47.20	&57.60	&\increase{60.00}{2.90}	&\increase{74.80}{0.69}	&\increase{69.42}{1.49} \\
\transSource & 67.60	&80.00	&87.20	&82.80	&83.60	&70.80	&80.80	&70.80	&59.60	&76.40	&79.20	&\increase{68.60}{11.50}	&\increase{80.63}{6.52}	&\increase{76.25}{8.32} \\
\midrule

\multicolumn{15}{l}{\textbf{\qwenTwo}} \\
\english      & 57.20 & 74.00 & 90.40 & 82.00 & 80.00 & 67.60 & 80.40 & 20.80 & 23.20 & 73.60 & 79.20 & 43.70 & 79.09 & 66.22 \\
\french & 54.80 & 74.00 & 92.00 & 82.00 & 79.60 & 66.80 & 79.20 & 22.40 & 20.80 & 73.60 & 80.40 &  \decrease{42.90}{0.80} &  \increase{79.14}{0.05} &  \decrease{65.96}{0.26} \\
\chinese & 56.00 & 76.40 & 92.00 & 81.20 & 77.60 & 71.20 & 79.60 & 28.80 & 20.40 & 73.60 & 85.20 &  \increase{44.70}{1.00} &  \increase{80.46}{1.37} &  \increase{67.45}{1.23} \\
\japanese & 51.20 & 72.80 & 88.80 & 78.80 & 76.00 & 75.20 & 79.20 & 27.20 & 20.80 & 72.40 & 80.80 &  \decrease{42.90}{0.80} &  \decrease{78.80}{0.29} &  \decrease{65.75}{0.47} \\
\multilingual & 64.80 & 77.20 & 89.20 & 82.80 & 81.60 & 70.40 & 82.00 & 28.40 & 23.60 & 73.20 & 79.60 &  \increase{47.50}{3.80} &  \increase{80.40}{1.31} &  \increase{68.44}{2.22} \\
\native & 72.00 & 83.60 & 90.40 & 82.80 & 79.60 & 75.20 & 83.20 & 32.40 & 40.40 & 76.80 & 85.20 &  \increase{55.40}{11.70} &  \increase{82.86}{3.77} &  \increase{72.87}{6.65} \\
\transEn & 66.00	&76.00	&90.40	&79.60	&79.20	&64.00	&77.60	&72.40	&61.20	&47.60	&58.40	&\increase{61.80}{18.10}	&\decrease{75.03}{4.06}	&\increase{70.22}{4.00} \\
\transSource & 72.40	&81.20	&90.40	&82.40	&80.00	&69.20	&81.20	&32.80	&43.20	&74.80	&80.80	&\increase{55.80}{12.10}	&\increase{80.74}{1.65}	&\increase{71.67}{5.45} \\
\midrule

\multicolumn{15}{l}{\textbf{\qwenTwoFive}} \\
\english      & 75.20 & 82.40 & 94.40 & 88.80 & 87.60 & 76.80 & 88.00 & 30.80 & 48.80 & 82.80 & 86.80 & 59.40 & 86.40 & 76.58 \\
\french & 74.00 & 85.20 & 92.40 & 90.00 & 85.60 & 78.80 & 86.80 & 33.60 & 51.20 & 82.40 & 84.00 &  \increase{60.30}{0.90} &  \decrease{86.11}{0.29} &  \increase{76.73}{0.15} \\
\chinese & 77.20 & 85.20 & 94.00 & 88.80 & 86.00 & 81.60 & 88.00 & 31.20 & 50.40 & 81.20 & 86.40 &  \increase{60.00}{0.60} &  \increase{87.14}{0.74} &  \increase{77.27}{0.69} \\
\japanese & 74.80 & 84.80 & 94.80 & 90.80 & 88.00 & 81.20 & 88.40 & 32.80 & 50.80 & 83.20 & 85.60 &  \increase{60.40}{1.00} &  \increase{87.66}{1.26} &  \increase{77.75}{1.17} \\
\multilingual & 76.40 & 86.80 & 92.80 & 89.60 & 88.00 & 78.40 & 87.20 & 30.40 & 50.00 & 81.20 & 86.00 &  \increase{59.50}{0.10} &  \increase{86.97}{0.57} &  \increase{76.98}{0.40} \\
\native & 75.20 & 86.80 & 94.40 & 89.20 & 85.60 & 81.20 & 87.60 & 35.60 & 46.00 & 84.40 & 86.40 &  \increase{60.30}{0.90} &  \increase{87.31}{0.91} &  \increase{77.49}{0.91} \\
\transEn & 70.40	&79.60	&94.40	&83.60	&82.40	&64.00	&81.60	&72.80	&62.40	&48.00	&64.80	&\increase{63.40}{4.60}	&\decrease{78.63}{7.77}	&\decrease{73.09}{3.49} \\
\transSource & 74.80	&85.60	&94.40	&88.80	&84.40	&80.40	&87.20	&35.20	&46.80	&82.80	&88.00	&\increase{59.90}{0.50}	&\increase{86.97}{0.57}	&\increase{77.13}{0.55} \\
\midrule

\multicolumn{15}{l}{\textbf{\mistral}} \\
\english      & 66.80 & 75.60 & 90.80 & 81.20 & 80.00 & 64.40 & 83.60 & 42.00 & 54.00 & 65.20 & 76.80 & 57.00 & 78.91 & 70.95 \\
\french & 66.00 & 81.20 & 92.40 & 82.00 & 81.60 & 74.80 & 83.20 & 46.40 & 70.40 & 74.40 & 76.40 &  \increase{64.30}{7.30} &  \increase{81.66}{2.75} &  \increase{75.35}{4.40} \\
\chinese & 70.40 & 81.60 & 91.60 & 81.60 & 82.00 & 73.20 & 84.80 & 49.20 & 69.60 & 69.20 & 77.20 &  \increase{64.60}{7.60} &  \increase{81.71}{2.80} &  \increase{75.49}{4.54} \\
\japanese & 67.20 & 84.40 & 90.00 & 85.60 & 83.20 & 74.40 & 84.80 & 45.60 & 69.60 & 72.00 & 76.00 &  \increase{63.60}{6.60} &  \increase{82.63}{3.72} &  \increase{75.71}{4.76} \\
\multilingual & 67.20 & 82.00 & 90.40 & 82.80 & 77.60 & 71.60 & 83.20 & 50.80 & 66.40 & 68.80 & 80.00 &  \increase{63.30}{6.30} &  \increase{81.09}{2.18} &  \increase{74.62}{3.67} \\
\native & 66.40 & 81.20 & 90.80 & 82.00 & 81.60 & 74.40 & 81.60 & 58.00 & 67.60 & 71.20 & 77.20 &  \increase{65.80}{8.80} &  \increase{81.26}{2.35} &  \increase{75.64}{4.69} \\
\transEn & 65.20	&76.80	&90.80	&82.00	&78.80	&62.00	&80.00	&71.60	&62.40	&47.20	&61.60	&\increase{61.60}{4.60}	&\decrease{76.00}{2.91}	&\decrease{70.76}{0.19} \\
\transSource & 68.80	&83.60	&90.80	&80.40	&82.00	&74.00	&85.60	&57.60	&66.80	&73.20	&77.60	&\increase{66.60}{9.60}	&\increase{82.00}{3.09}	&\increase{76.40}{5.45} \\
\midrule

\multicolumn{15}{l}{\textbf{\aya}} \\
\english      & 39.60 & 75.60 & 82.40 & 81.60 & 72.40 & 67.20 & 77.20 & 20.00 & 17.20 & 40.80 & 71.20 & 29.40 & 75.37 & 58.65 \\
\french & 43.60 & 74.40 & 81.60 & 80.00 & 75.20 & 67.60 & 77.60 & 19.60 & 23.20 & 36.40 & 74.00 &  \increase{30.70}{1.30} &  \increase{75.77}{0.40} &  \increase{59.38}{0.73} \\
\chinese & 44.00 & 74.40 & 80.40 & 77.60 & 73.20 & 66.00 & 77.20 & 20.00 & 22.40 & 40.40 & 76.40 &  \increase{31.70}{2.30} &  \decrease{75.03}{0.34} &  \increase{59.27}{0.62} \\
\japanese & 42.80 & 72.40 & 83.20 & 79.20 & 72.80 & 66.80 & 77.20 & 20.40 & 20.00 & 38.40 & 72.00 &  \increase{30.40}{1.00} &  \decrease{74.80}{0.57} & $58.65_{0.00-}$ \\
\multilingual & 46.80 & 74.00 & 83.60 & 78.00 & 72.80 & 67.60 & 79.60 & 19.20 & 17.60 & 36.80 & 72.00 &  \increase{30.10}{0.70} & $75.37_{0.00-}$ &  \increase{58.91}{0.26} \\
\native & 44.80 & 77.60 & 82.40 & 79.60 & 75.20 & 66.80 & 79.20 & 19.60 & 22.00 & 44.00 & 76.40 &  \increase{32.60}{3.20} &  \increase{76.74}{1.37} &  \increase{60.69}{2.04} \\
\transEn & 64.40	&71.20	&82.40	&76.40	&74.40	&57.60	&69.20	&65.20	&62.80	&48.00	&59.60	&\increase{60.10}{30.70}	&\decrease{70.11}{5.26}	&\increase{66.47}{7.82}\\
\transSource & 46.40	&74.80	&82.40	&80.00	&74.80	&66.00	&78.80	&19.20	&20.80	&42.40	&76.00	&\increase{32.20}{2.80}	&\increase{76.11}{0.74}	&\increase{60.15}{1.50}\\
\midrule

\multicolumn{15}{l}{\textbf{\gptThreeFive}} \\
\english      & 39.60 & 75.20 & 86.00 & 76.80 & 62.80 & 59.20 & 66.40 & 63.60 & 12.80 & 60.80 & 67.20 & 44.20 & 70.51 & 60.95 \\
\french & 34.00 & 74.80 & 86.00 & 84.80 & 79.20 & 66.00 & 67.60 & 63.20 & 15.20 & 55.60 & 74.40 &  \decrease{42.00}{2.20} &  \increase{76.11}{5.60} &  \increase{63.71}{2.76} \\
\chinese & 30.40 & 78.80 & 83.20 & 78.40 & 77.20 & 69.20 & 77.60 & 67.20 & 15.60 & 62.40 & 73.20 &  \decrease{43.90}{0.30} &  \increase{76.80}{6.29} &  \increase{64.84}{3.89} \\
\japanese & 27.60 & 78.80 & 85.20 & 82.00 & 72.40 & 73.20 & 74.40 & 69.60 & 14.00 & 61.20 & 76.00 &  \decrease{43.10}{1.10} &  \increase{77.43}{6.92} &  \increase{64.95}{4.00} \\
\multilingual & 54.40 & 79.60 & 83.60 & 79.20 & 73.60 & 68.00 & 75.20 & 68.00 & 24.40 & 57.20 & 74.40 &  \increase{51.00}{6.80} &  \increase{76.23}{5.72} &  \increase{67.05}{6.10} \\
\native & 57.20 & 79.60 & 86.00 & 80.40 & 81.60 & 75.20 & 77.60 & 73.60 & 30.00 & 58.80 & 74.00 &  \increase{54.90}{10.70} &  \increase{79.26}{8.75} &  \increase{70.40}{9.45} \\
\midrule

\multicolumn{15}{l}{\textbf{\gptFourO}} \\
\english      & 87.20 & 90.80 & 94.80 & 92.00 & 87.20 & 84.80 & 92.00 & 83.20 & 84.00 & 88.80 & 90.00 & 85.80 & 90.23 & 88.62 \\
\french & 87.20 & 90.80 & 94.40 & 92.80 & 89.20 & 82.40 & 90.80 & 85.20 & 84.00 & 88.80 & 87.60 &  \increase{86.30}{0.50} &  \decrease{89.71}{0.52} &  \decrease{88.47}{0.15} \\
\chinese & 86.80 & 90.80 & 93.60 & 93.20 & 90.00 & 83.20 & 91.20 & 85.20 & 84.40 & 90.00 & 90.80 &  \increase{86.60}{0.80} &  \increase{90.40}{0.17} &  \increase{89.02}{0.40} \\
\japanese & 85.20 & 89.60 & 94.80 & 92.80 & 86.80 & 86.00 & 92.00 & 82.80 & 81.60 & 88.40 & 87.60 &  \decrease{84.50}{1.30} &  \decrease{89.94}{0.29} &  \decrease{87.96}{0.66} \\
\multilingual & 86.40 & 90.00 & 92.80 & 94.00 & 89.20 & 84.00 & 92.40 & 84.00 & 80.40 & 88.00 & 89.20 &  \decrease{84.70}{1.10} & $90.23_{0.00-}$ &  \decrease{88.22}{0.40} \\
\native & 85.20 & 88.00 & 94.80 & 91.20 & 89.20 & 85.20 & 90.40 & 85.60 & 81.60 & 88.40 & 90.00 &  \decrease{85.20}{0.60} &  \decrease{89.66}{0.57} &  \decrease{88.04}{0.58} \\
\bottomrule

    \end{tabular}
    \caption{Accuracies ($\%$) of \english, \multilingual, \native, all \monolingual ICL modes (\french, \chinese and \japanese) and two translation strategies (\transEn, \transSource) across $11$ languages of the \mgsm dataset. AVG represents the average accuracy of the language set (LRLs, HRLs or All languages). The \underline{underlined languages} in the table header are \underline{LRLs}, otherwise HRLs. The subscript indicates the performance \textcolor{ForestGreen}{increase$\uparrow$} (or \textcolor{OrangeRed}{decrease$\downarrow$}) of all other modes compared to the \english ICL mode.}
    % \vspace{-.2cm}
    \label{tab:vanilla_eval:mgsm}
\end{table*}

\begin{table*}[!htbp]
    \setlength{\tabcolsep}{3pt}
    \scriptsize
    \centering
    \alternaterowcolors
\begin{tabular}{l|cccccccccccc|lll}
\toprule
\textbf{\xcopa} &
  \multicolumn{1}{c}{\textbf{en}} &
  \multicolumn{1}{c}{{\textbf{\underline{et}}}} &
  \multicolumn{1}{c}{{\textbf{\underline{ht}}}} &
  \multicolumn{1}{c}{\textbf{id}} &
  \multicolumn{1}{c}{\textbf{it}} &
  \multicolumn{1}{c}{{\textbf{\underline{qu}}}} &
  \multicolumn{1}{c}{{\textbf{\underline{sw}}}} &
  \multicolumn{1}{c}{{\textbf{\underline{ta}}}} &
  \multicolumn{1}{c}{{\textbf{\underline{th}}}} &
  \multicolumn{1}{c}{\textbf{tr}} &
  \multicolumn{1}{c}{\textbf{\underline{vi}}} &
  \textbf{zh} &
  \textbf{\underline{LRL AVG}} &
  \textbf{HRL AVG} &
  \textbf{ALL AVG} \\
\midrule

\multicolumn{16}{l}{\textbf{\llamaThree}} \\
\english      & 95.20 & 55.80 & 10.20 & 79.60 & 86.60 & 7.60  & 40.00 & 59.40 & 69.60 & 72.00 & 81.00 & 87.00 & 46.23 & 84.08 & 62.00 \\
\italian & 93.80 & 59.80 & 46.80 & 81.00 & 89.60 & 41.20 & 57.40 & 58.80 & 73.40 & 76.00 & 80.20 & 88.20 &  \increase{59.66}{13.43} &  \increase{85.72}{1.64} &  \increase{70.52}{8.52} \\
\chinese & 93.80 & 59.80 & 52.80 & 82.40 & 85.60 & 47.60 & 55.40 & 59.00 & 79.00 & 74.60 & 79.60 & 90.80 &  \increase{61.89}{15.66} &  \increase{85.44}{1.36} &  \increase{71.70}{9.70} \\
\multilingual & 94.60 & 57.80 & 51.80 & 81.60 & 87.80 & 46.40 & 60.40 & 60.80 & 79.20 & 78.80 & 80.40 & 88.80 &  \increase{62.40}{16.17} &  \increase{86.32}{2.24} &  \increase{72.37}{10.37} \\
\native & 95.20 & 68.60 & 61.60 & 85.00 & 89.60 & 50.40 & 62.00 & 65.80 & 77.80 & 79.80 & 84.80 & 90.80 &  \increase{67.29}{21.06} &  \increase{88.08}{4.00} &  \increase{75.95}{13.95} \\
\transEn & 95.20	&84.00	&69.80	&83.00	&88.40	&61.00	&69.00	&70.40	&66.00	&86.20	&85.20	&85.20	&\increase{72.20}{25.97}	&\increase{87.60}{3.52}	&\increase{78.62}{16.62} \\
\transSource & 95.20	&69.60	&62.00	&87.40	&89.20	&52.00	&62.00	&63.80	&78.60	&79.80	&84.00	&89.20	&\increase{67.43}{21.20}	&\increase{88.16}{4.08}	&\increase{76.07}{14.07} \\
\midrule

\multicolumn{16}{l}{\textbf{\llamaThreeOne}} \\
\english      & 95.60 & 65.20 & 27.00 & 84.20 & 88.40 & 26.00 & 52.60 & 62.40 & 73.80 & 76.80 & 84.40 & 89.80 & 55.91 & 86.96 & 68.85 \\
\italian & 95.00 & 62.20 & 28.40 & 86.00 & 92.80 & 32.60 & 61.60 & 70.20 & 76.60 & 78.20 & 85.00 & 90.80 &  \increase{59.51}{3.60} &  \increase{88.56}{1.60} &  \increase{71.62}{2.77} \\
\chinese & 95.00 & 65.20 & 48.40 & 83.80 & 88.20 & 32.20 & 58.80 & 68.80 & 76.40 & 78.60 & 86.00 & 91.40 &  \increase{62.26}{6.35} &  \increase{87.40}{0.44} &  \increase{72.73}{3.88} \\
\multilingual & 96.00 & 60.40 & 57.20 & 87.60 & 90.80 & 46.80 & 61.60 & 70.80 & 78.60 & 83.00 & 87.40 & 90.80 &  \increase{66.11}{10.20} &  \increase{89.64}{2.68} &  \increase{75.92}{7.07} \\
\native & 95.60 & 72.40 & 66.20 & 89.80 & 92.80 & 52.60 & 66.60 & 75.20 & 80.80 & 84.40 & 87.60 & 91.40 &  \increase{71.63}{15.72} &  \increase{90.80}{3.84} &  \increase{79.62}{10.77} \\
\transEn & 95.60	&87.60	&74.60	&88.20	&89.60	&68.20	&73.20	&74.20	&72.60	&88.00	&86.80	&87.00	&\increase{76.74}{20.83}	&\increase{89.68}{2.72}	&\increase{82.13}{13.28} \\
\transSource & 95.60	&74.40	&63.40	&89.00	&90.00	&52.20	&65.20	&74.00	&78.40	&83.60	&85.80	&92.00	&\increase{70.49}{14.58}	&\increase{90.04}{3.08}	&\increase{78.63}{9.78} \\
\midrule

\multicolumn{16}{l}{\textbf{\qwenTwo}} \\
\english      & 97.00 & 61.80 & 50.60 & 88.00 & 90.20 & 49.80 & 53.20 & 58.40 & 77.80 & 75.40 & 84.40 & 91.00 & 62.29 & 88.32 & 73.13 \\
\italian & 92.20 & 66.40 & 51.60 & 88.40 & 95.40 & 52.20 & 54.40 & 59.80 & 79.80 & 78.20 & 83.80 & 87.20 &  \increase{64.00}{1.71} &  \decrease{88.28}{0.04} &  \increase{74.12}{0.99} \\
\chinese & 88.60 & 65.00 & 51.80 & 81.20 & 86.00 & 50.20 & 53.00 & 60.00 & 79.00 & 77.60 & 83.60 & 93.60 &  \increase{63.23}{0.94} &  \decrease{85.40}{2.92} &  \decrease{72.47}{0.66} \\
\multilingual & 96.00 & 63.20 & 53.80 & 90.60 & 94.00 & 53.00 & 53.40 & 59.60 & 80.40 & 77.60 & 83.40 & 93.00 &  \increase{63.83}{1.54} &  \increase{90.24}{1.92} &  \increase{74.83}{1.70} \\
\native & 97.00 & 71.20 & 54.80 & 93.00 & 95.40 & 51.40 & 60.20 & 63.20 & 83.60 & 81.20 & 89.00 & 93.60 &  \increase{67.63}{5.34} &  \increase{92.04}{3.72} &  \increase{77.80}{4.67} \\
\transEn & 97.00	&88.40	&80.00	&91.00	&92.00	&75.60	&79.80	&82.80	&76.20	&90.20	&88.40	&89.40	&\increase{81.60}{19.31}	&\increase{91.92}{3.60}	&\increase{85.90}{12.77} \\
\transSource & 97.00	&67.20	&54.60	&91.20	&94.80	&51.00	&56.80	&61.60	&83.00	&80.20	&86.40	&91.80	&\increase{65.80}{3.51}	&\increase{91.00}{2.68}	&\increase{76.30}{3.17} \\
\midrule

\multicolumn{16}{l}{\textbf{\qwenTwoFive}} \\
\english      & 97.40 & 62.20 & 56.40 & 89.40 & 93.20 & 50.80 & 53.40 & 58.20 & 83.40 & 80.20 & 88.40 & 93.60 & 64.69 & 90.76 & 75.55 \\
\italian & 96.40 & 65.20 & 57.80 & 89.00 & 95.40 & 49.00 & 52.20 & 58.40 & 82.00 & 83.60 & 87.20 & 92.80 &  \decrease{64.54}{0.15} &  \increase{91.44}{0.68} &  \increase{75.75}{0.20} \\
\chinese & 95.80 & 65.00 & 58.40 & 89.00 & 92.40 & 50.80 & 51.40 & 58.60 & 83.40 & 82.00 & 89.80 & 94.60 &  \increase{65.34}{0.65} & $90.76_{0.00-}$ &  \increase{75.93}{0.38} \\
\multilingual & 97.00 & 65.40 & 56.80 & 91.80 & 94.00 & 49.20 & 49.60 & 59.20 & 83.60 & 84.00 & 88.60 & 92.40 &  \decrease{64.63}{0.06} &  \increase{91.84}{1.08} &  \increase{75.97}{0.42} \\
\native & 97.40 & 69.60 & 62.80 & 91.40 & 95.40 & 50.60 & 54.00 & 61.40 & 85.40 & 84.40 & 90.00 & 94.60 &  \increase{67.69}{3.00} &  \increase{92.64}{1.88} &  \increase{78.08}{2.53} \\
\transEn & 97.40	&87.80	&80.60	&91.20	&91.20	&75.00	&77.40	&80.60	&74.20	&89.80	&87.80	&88.00	&\increase{80.49}{15.80}	&\increase{91.52}{0.76}	&\increase{85.08}{9.53} \\
\transSource & 97.40	&67.40	&61.20	&91.00	&96.00	&46.80	&53.60	&60.40	&85.40	&82.80	&88.80	&95.20	&\increase{66.23}{1.54}	&\increase{92.48}{1.72}	&\increase{77.17}{1.62} \\
\midrule

\multicolumn{16}{l}{\textbf{\mistral}} \\
\english      & 96.60 & 57.60 & 58.00 & 83.00 & 93.20 & 50.60 & 51.60 & 73.40 & 64.60 & 73.60 & 80.00 & 91.20 & 62.26 & 87.52 & 72.78 \\
\italian & 96.00 & 57.40 & 55.40 & 82.00 & 95.80 & 48.60 & 53.80 & 73.00 & 64.80 & 73.00 & 81.60 & 90.00 &  \decrease{62.09}{0.17} &  \decrease{87.36}{0.16} &  \decrease{72.62}{0.16} \\
\chinese & 96.40 & 58.20 & 52.00 & 83.00 & 92.60 & 49.00 & 53.80 & 70.20 & 66.80 & 71.00 & 81.00 & 91.60 &  \decrease{61.57}{0.69} &  \decrease{86.92}{0.60} &  \decrease{72.13}{0.65} \\
\multilingual & 95.80 & 60.20 & 54.00 & 85.80 & 94.40 & 48.80 & 56.00 & 76.20 & 62.80 & 72.60 & 82.60 & 92.20 &  \increase{62.94}{0.68} &  \increase{88.16}{0.64} &  \increase{73.45}{0.67} \\
\native & 96.60 & 74.00 & 64.00 & 87.40 & 95.80 & 48.80 & 62.20 & 82.20 & 79.00 & 80.00 & 86.20 & 91.60 &  \increase{70.91}{8.65} &  \increase{90.28}{2.76} &  \increase{78.98}{6.20} \\
\transEn & 96.60	&88.00	&75.40	&87.60	&89.40	&70.00	&74.20	&77.40	&74.00	&88.20	&85.80	&84.80	&\increase{77.83}{15.57}	&\increase{89.32}{1.80}	&\increase{82.62}{9.84} \\
\transSource & 96.60	&71.40	&63.00	&88.00	&94.40	&49.20	&60.60	&76.40	&73.20	&80.60	&85.40	&93.00	&\increase{68.46}{6.20}	&\increase{90.52}{3.00}	&\increase{77.65}{4.87} \\
\midrule

\multicolumn{16}{l}{\textbf{\aya}} \\
\english      & 95.40 & 28.20 & 17.60 & 83.80 & 84.20 & 0.00  & 11.20 & 45.20 & 10.00 & 80.40 & 73.60 & 77.60 & 26.54 & 84.28 & 50.60 \\
\italian & 92.00 & 15.40 & 18.40 & 86.40 & 91.40 & 0.20 & 11.20 & 45.80 & 34.60 & 82.20 & 79.60 & 85.60 &  \increase{29.31}{2.77} &  \increase{87.52}{3.24} &  \increase{53.57}{2.97} \\
\chinese & 90.60 & 39.40 & 34.80 & 81.00 & 82.80 & 0.00 & 17.60 & 38.60 & 23.00 & 79.00 & 78.40 & 92.20 &  \increase{33.11}{6.57} &  \increase{85.12}{0.84} &  \increase{54.78}{4.18} \\
\multilingual & 93.40 & 52.20 & 52.00 & 88.00 & 90.00 & 5.60 & 46.00 & 58.20 & 43.60 & 84.00 & 83.20 & 90.20 &  \increase{48.69}{22.15} &  \increase{89.12}{4.84} &  \increase{65.53}{14.93} \\
\native & 95.40 & 54.00 & 56.20 & 88.80 & 91.40 & 53.40 & 53.40 & 69.40 & 62.40 & 85.60 & 86.20 & 92.20 &  \increase{62.14}{35.60} &  \increase{90.68}{6.40} &  \increase{74.03}{23.43} \\
\transEn & 95.40	&84.40	&75.20	&86.60	&86.80	&67.40	&74.40	&74.60	&67.80	&86.40	&84.20	&85.60	&\increase{75.43}{48.89}	&\increase{88.16}{3.88}	&\increase{80.73}{30.13} \\
\transSource & 95.40	&54.20	&54.40	&89.40	&90.80	&45.80	&51.00	&65.80	&60.00	&83.80	&86.00	&90.60	&\increase{59.60}{33.06}	&\increase{90.00}{5.72}	&\increase{72.27}{21.67} \\
\midrule

\multicolumn{16}{l}{\textbf{\gptThreeFive}} \\
\english      & 96.00 & 77.20 & 56.80 & 83.00 & 88.80 & 48.40 & 70.80 & 52.00 & 64.80 & 76.20 & 74.00 & 83.80 & 63.43 & 85.56 & 72.65 \\
\italian & 95.00 & 78.60 & 57.20 & 83.20 & 91.20 & 49.20 & 71.60 & 49.40 & 60.40 & 79.40 & 77.40 & 85.60 &  \decrease{63.40}{0.03} &  \increase{86.88}{1.32} &  \increase{73.18}{0.53} \\
\chinese & 94.40 & 77.20 & 58.60 & 84.40 & 86.80 & 48.80 & 69.80 & 50.00 & 62.00 & 76.80 & 75.20 & 86.80 &  \decrease{63.09}{0.34} &  \increase{85.84}{0.28} &  \decrease{72.57}{0.08} \\
\multilingual & 93.80 & 75.00 & 56.80 & 87.40 & 89.60 & 49.20 & 71.40 & 48.00 & 62.80 & 85.80 & 75.80 & 86.80 &  \decrease{62.71}{0.72} &  \increase{88.68}{3.12} &  \increase{73.53}{0.88} \\
\native & 96.00 & 85.20 & 67.60 & 87.20 & 90.80 & 48.00 & 77.20 & 53.20 & 61.40 & 87.60 & 77.60 & 87.60 &  \increase{67.17}{3.74} &  \increase{89.80}{4.24} &  \increase{76.60}{3.95} \\
\midrule

\multicolumn{16}{l}{\textbf{\gptFourO}} \\
\english      & 98.60 & 93.20 & 80.00 & 94.20 & 97.60 & 49.80 & 84.20 & 83.40 & 88.20 & 95.20 & 92.80 & 95.60 & 81.66 & 96.24 & 87.73 \\
\italian & 98.80 & 91.60 & 74.40 & 95.00 & 98.20 & 49.20 & 82.60 & 84.40 & 89.00 & 95.60 & 92.80 & 96.40 &  \decrease{80.57}{1.09} &  \increase{96.80}{0.56} &  \decrease{87.33}{0.40} \\
\chinese & 98.00 & 93.40 & 81.00 & 94.20 & 97.80 & 49.80 & 84.00 & 84.40 & 88.80 & 94.80 & 93.40 & 95.00 &  \increase{82.11}{0.45} &  \decrease{95.96}{0.28} &  \increase{87.88}{0.15} \\
\multilingual & 98.60 & 93.80 & 80.00 & 96.00 & 98.20 & 50.20 & 85.00 & 86.20 & 92.40 & 96.20 & 94.40 & 95.20 &  \increase{83.14}{1.48} &  \increase{96.84}{0.60} &  \increase{88.85}{1.12} \\
\native & 98.60 & 94.80 & 89.60 & 95.20 & 98.20 & 52.20 & 87.60 & 88.80 & 93.80 & 95.20 & 95.00 & 95.20 &  \increase{85.97}{4.31} &  \increase{96.48}{0.24} &  \increase{90.35}{2.62} \\
\bottomrule

    \end{tabular}
    \caption{Accuracies ($\%$) of \english, \multilingual, \native, both \monolingual ICL modes (\italian and \chinese) and two translation strategies (\transEn, \transSource) across $12$ languages of the \xcopa dataset. AVG represents the average accuracy of the language set (LRLs, HRLs or All languages). The \underline{underlined languages} in the table header are \underline{LRLs}, otherwise HRLs. The subscript indicates the performance \textcolor{ForestGreen}{increase$\uparrow$} (or \textcolor{OrangeRed}{decrease$\downarrow$}) of all other modes compared to the \english ICL mode.}
    % \vspace{-.2cm}
    \label{tab:vanilla_eval:xcopa}
\end{table*}

\begin{table*}[!htbp]
    \setlength{\tabcolsep}{2.5pt}
    \scriptsize
    \centering
    \alternaterowcolors
\begin{tabular}{l|ccccccccccccc|lll}
\toprule
\textbf{\xlwic} &
  \multicolumn{1}{c}{\textbf{\underline{bg}}} &
  \multicolumn{1}{c}{\textbf{da}} &
  \multicolumn{1}{c}{\textbf{de}} &
  \multicolumn{1}{c}{\textbf{en}} &
  \multicolumn{1}{c}{\textbf{\underline{et}}} &
  \multicolumn{1}{c}{\textbf{\underline{fa}}} &
  \multicolumn{1}{c}{\textbf{fr}} &
  \multicolumn{1}{c}{\textbf{\underline{hr}}} &
  \multicolumn{1}{c}{\textbf{it}} &
  \multicolumn{1}{c}{\textbf{ja}} &
  \multicolumn{1}{c}{\textbf{ko}} &
  \multicolumn{1}{c}{\textbf{nl}} &
  \textbf{zh} &
  \textbf{\underline{LRL AVG}} &
  \textbf{HRL AVG} &
  \textbf{ALL AVG} \\
  \midrule

\multicolumn{17}{l}{\textbf{\llamaThree}} \\
\english      & 55.13 & 66.15 & 59.49 & 67.44 & 55.38 & 63.33 & 59.49 & 55.64 & 53.85 & 54.62 & 56.67 & 55.90 & 64.10 & 57.37 & 59.74 & 59.01 \\
\french & 57.95 & 63.08 & 66.67 & 64.62 & 52.82 & 68.21 & 66.15 & 57.18 & 56.15 & 55.13 & 51.79 & 62.82 & 65.38 &  \increase{59.04}{1.67} &  \increase{61.31}{1.57} &  \increase{60.61}{1.60} \\
\chinese & 56.67 & 64.10 & 64.87 & 63.59 & 55.38 & 63.08 & 58.97 & 54.87 & 62.31 & 57.18 & 48.21 & 63.85 & 63.33 &  \increase{57.50}{0.13} &  \increase{60.71}{0.97} &  \increase{59.72}{0.71} \\
\japanese & 59.74 & 61.79 & 66.15 & 65.38 & 52.56 & 70.77 & 60.00 & 56.67 & 59.23 & 58.97 & 55.13 & 64.87 & 63.33 &  \increase{59.94}{2.57} &  \increase{61.65}{1.91} &  \increase{61.12}{2.11} \\
\multilingual & 57.18 & 60.00 & 62.82 & 64.10 & 53.85 & 69.23 & 60.00 & 56.92 & 58.97 & 58.46 & 57.44 & 62.82 & 58.72 &  \increase{59.29}{1.92} &  \increase{60.37}{0.63} &  \increase{60.04}{1.03} \\
\native & 63.59 & 55.90 & 69.23 & 67.44 & 62.05 & 68.72 & 66.15 & 58.21 & 59.49 & 58.97 & 66.67 & 66.41 & 63.33 &  \increase{63.14}{5.77} &  \increase{63.73}{3.99} &  \increase{63.55}{4.54} \\
\midrule

\multicolumn{17}{l}{\textbf{\llamaThreeOne}} \\
\english      & 51.54 & 50.77 & 61.28 & 66.92 & 44.62 & 29.74 & 56.67 & 51.79 & 15.90 & 48.46 & 33.85 & 51.28 & 57.44 & 44.42 & 49.17 & 47.71 \\
\french & 54.62 & 60.00 & 66.41 & 64.62 & 49.74 & 22.56 & 62.31 & 57.44 & 52.56 & 49.49 & 53.33 & 65.13 & 54.10 &  \increase{46.09}{1.67} &  \increase{58.66}{9.49} &  \increase{54.79}{7.08} \\
\chinese & 54.87 & 62.82 & 62.31 & 61.79 & 53.08 & 55.64 & 56.15 & 57.95 & 56.67 & 49.74 & 48.21 & 62.31 & 62.05 &  \increase{55.38}{10.96} &  \increase{58.01}{8.84} &  \increase{57.20}{9.49} \\
\japanese & 54.36 & 60.00 & 63.59 & 60.00 & 47.69 & 63.33 & 58.72 & 52.05 & 58.21 & 53.59 & 48.97 & 58.21 & 56.41 &  \increase{54.36}{9.94} &  \increase{57.52}{8.35} &  \increase{56.55}{8.84} \\
\multilingual & 57.69 & 60.00 & 65.38 & 62.56 & 55.13 & 58.46 & 57.95 & 56.92 & 55.38 & 54.36 & 52.31 & 64.62 & 58.46 &  \increase{57.05}{12.63} &  \increase{59.00}{9.83} &  \increase{58.40}{10.69} \\
\native & 61.79 & 64.10 & 68.46 & 66.92 & 62.05 & 70.51 & 62.31 & 57.18 & 56.15 & 53.59 & 63.08 & 69.49 & 62.05 &  \increase{62.88}{18.46} &  \increase{62.91}{13.74} &  \increase{62.90}{15.19} \\
\midrule

\multicolumn{17}{l}{\textbf{\qwenTwo}} \\
\english      & 28.21 & 38.46 & 66.67 & 68.46 & 56.92 & 53.85 & 63.33 & 54.87 & 39.23 & 1.28  & 15.90 & 64.36 & 0.77  & 48.46 & 39.83 & 42.49 \\
\french & 53.59 & 51.54 & 64.87 & 66.92 & 53.85 & 58.21 & 63.08 & 56.41 & 49.74 & 25.90 & 33.08 & 62.82 & 14.10 &  \increase{55.51}{7.05} &  \increase{48.01}{8.18} &  \increase{50.32}{7.83} \\
\chinese & 47.18 & 27.44 & 68.46 & 67.44 & 57.44 & 58.97 & 61.79 & 55.13 & 40.51 & 56.41 & 37.69 & 60.00 & 68.72 &  \increase{54.68}{6.22} &  \increase{54.27}{14.44} &  \increase{54.40}{11.91} \\
\japanese & 56.15 & 30.26 & 66.41 & 66.92 & 56.15 & 60.26 & 61.79 & 56.67 & 47.95 & 60.77 & 59.49 & 66.41 & 40.26 &  \increase{57.31}{8.85} &  \increase{55.58}{15.75} &  \increase{56.11}{13.62} \\
\multilingual & 53.59 & 60.00 & 65.64 & 65.64 & 58.46 & 58.72 & 60.26 & 54.36 & 56.67 & 58.72 & 66.92 & 67.18 & 57.95 &  \increase{56.28}{7.82} &  \increase{62.11}{22.28} &  \increase{60.32}{17.83} \\
\native & 55.13 & 62.05 & 69.49 & 68.46 & 59.23 & 62.05 & 63.08 & 54.62 & 60.51 & 60.77 & 68.21 & 64.87 & 68.72 &  \increase{57.76}{9.30} &  \increase{65.13}{25.30} &  \increase{62.86}{20.37} \\
\midrule

\multicolumn{17}{l}{\textbf{\qwenTwoFive}} \\
\english      & 58.97 & 61.28 & 73.33 & 70.51 & 55.64 & 53.59 & 65.90 & 63.08 & 57.95 & 55.64 & 64.10 & 67.69 & 59.74 & 57.82 & 64.02 & 62.11 \\
\french & 55.13 & 57.18 & 68.97 & 65.13 & 55.38 & 54.10 & 64.87 & 57.18 & 62.05 & 57.95 & 62.82 & 65.64 & 62.56 &  \decrease{55.45}{2.37} &  \decrease{63.02}{1.00} &  \decrease{60.69}{1.42} \\
\chinese & 53.33 & 54.87 & 67.18 & 67.18 & 54.62 & 51.28 & 63.85 & 57.44 & 57.95 & 64.87 & 62.82 & 61.79 & 66.67 &  \decrease{54.17}{3.65} &  \decrease{63.02}{1.00} &  \decrease{60.30}{1.81} \\
\japanese & 53.59 & 54.62 & 67.44 & 67.18 & 54.10 & 49.74 & 63.08 & 57.18 & 56.41 & 64.10 & 63.33 & 65.13 & 60.00 &  \decrease{53.65}{4.17} &  \decrease{62.36}{1.66} &  \decrease{59.68}{2.43} \\
\multilingual & 57.44 & 60.51 & 69.23 & 68.21 & 54.62 & 55.38 & 61.28 & 56.41 & 57.18 & 64.36 & 66.67 & 67.69 & 64.62 &  \decrease{55.96}{1.86} &  \increase{64.42}{0.40} &  \decrease{61.81}{0.30} \\
\native & 60.00 & 63.33 & 72.56 & 70.51 & 55.90 & 63.33 & 64.87 & 55.64 & 62.31 & 64.10 & 70.77 & 72.82 & 66.67 &  \increase{58.72}{0.90} &  \increase{67.55}{3.53} &  \increase{64.83}{2.72} \\
\midrule

\multicolumn{17}{l}{\textbf{\mistral}} \\
\english      & 52.56 & 63.85 & 70.51 & 65.64 & 51.79 & 49.23 & 61.03 & 52.56 & 38.97 & 23.08 & 53.59 & 67.44 & 54.10 & 51.54 & 55.36 & 54.18 \\
\french & 52.05 & 57.69 & 66.41 & 64.36 & 51.54 & 48.46 & 60.77 & 57.18 & 57.18 & 56.67 & 55.13 & 59.74 & 54.10 &  \increase{52.31}{0.77} &  \increase{59.12}{3.76} &  \increase{57.02}{2.84} \\
\chinese & 47.95 & 57.95 & 63.33 & 63.85 & 50.51 & 47.18 & 59.49 & 52.31 & 56.41 & 57.18 & 54.62 & 61.28 & 62.82 &  \decrease{49.49}{2.05} &  \increase{59.66}{4.30} &  \increase{56.53}{2.35} \\
\japanese & 48.21 & 53.08 & 60.77 & 64.10 & 51.28 & 47.44 & 58.46 & 52.31 & 56.92 & 64.36 & 56.67 & 58.72 & 53.59 &  \decrease{49.81}{1.73} &  \increase{58.52}{3.16} &  \increase{55.84}{1.66} \\
\multilingual & 51.54 & 56.67 & 67.18 & 61.28 & 52.05 & 49.74 & 61.28 & 51.28 & 54.36 & 57.69 & 56.67 & 58.72 & 59.49 &  \decrease{51.15}{0.39} &  \increase{59.26}{3.90} &  \increase{56.77}{2.59} \\
\native & 57.44 & 57.69 & 70.00 & 65.64 & 57.69 & 66.92 & 60.77 & 58.72 & 55.13 & 64.36 & 64.87 & 65.90 & 62.82 &  \increase{60.19}{8.65} &  \increase{63.02}{7.66} &  \increase{62.15}{7.97} \\
\midrule

\multicolumn{17}{l}{\textbf{\aya}} \\
\english      & 53.08 & 57.44 & 60.51 & 66.41 & 58.46 & 66.41 & 57.44 & 57.18 & 26.41 & 44.62 & 59.74 & 61.79 & 60.51 & 58.78 & 54.99 & 56.15 \\
\french & 55.64 & 59.23 & 71.79 & 63.85 & 57.69 & 72.31 & 64.87 & 54.10 & 61.79 & 65.38 & 65.64 & 70.77 & 65.64 &  \increase{59.94}{1.16} &  \increase{65.44}{10.45} &  \increase{63.75}{7.60} \\
\chinese & 55.90 & 64.62 & 68.97 & 63.85 & 56.15 & 73.59 & 61.28 & 53.08 & 58.46 & 59.49 & 67.69 & 66.41 & 61.79 &  \increase{59.68}{0.90} &  \increase{63.62}{8.63} &  \increase{62.41}{6.26} \\
\japanese & 59.49 & 58.72 & 70.00 & 66.67 & 58.21 & 71.54 & 63.59 & 55.13 & 54.62 & 61.03 & 68.46 & 70.00 & 64.36 &  \increase{61.09}{2.31} &  \increase{64.16}{9.17} &  \increase{63.21}{7.06} \\
\multilingual & 58.46 & 58.46 & 69.74 & 63.33 & 56.41 & 69.23 & 66.15 & 55.38 & 61.28 & 65.64 & 68.21 & 71.28 & 63.08 &  \increase{59.87}{1.09} &  \increase{65.24}{10.25} &  \increase{63.59}{7.44} \\
\native & 51.54 & 63.08 & 71.54 & 66.41 & 56.92 & 78.97 & 64.87 & 59.49 & 61.03 & 61.03 & 67.95 & 70.51 & 61.79 &  \increase{61.73}{2.95} &  \increase{65.36}{10.37} &  \increase{64.24}{8.09} \\
\midrule

\multicolumn{17}{l}{\textbf{\gptThreeFive}} \\
\english      & 54.36 & 50.77 & 62.31 & 63.59 & 54.62 & 54.10 & 58.46 & 51.79 & 32.82 & 30.77 & 56.67 & 65.13 & 21.03 & 53.72 & 49.06 & 50.49 \\
\french & 55.13 & 56.67 & 64.62 & 62.56 & 58.97 & 52.05 & 58.72 & 56.41 & 56.41 & 58.46 & 59.23 & 61.79 & 58.97 &  \increase{55.64}{1.92} &  \increase{59.72}{10.66} &  \increase{58.46}{7.97} \\
\chinese & 53.59 & 61.28 & 62.56 & 59.74 & 55.13 & 54.36 & 56.41 & 56.92 & 55.90 & 55.64 & 56.41 & 59.23 & 53.85 &  \increase{55.00}{1.28} &  \increase{57.89}{8.83} &  \increase{57.00}{6.51} \\
\japanese & 53.33 & 55.38 & 65.90 & 63.08 & 58.97 & 53.08 & 58.97 & 55.90 & 56.15 & 56.41 & 55.90 & 64.36 & 54.87 &  \increase{55.32}{1.60} &  \increase{59.00}{9.94} &  \increase{57.87}{7.38} \\
\multilingual & 52.82 & 60.00 & 66.92 & 59.23 & 60.00 & 54.36 & 60.77 & 56.41 & 53.59 & 56.67 & 61.54 & 63.33 & 59.49 &  \increase{55.90}{2.18} &  \increase{60.17}{11.11} &  \increase{58.86}{8.37} \\
\native & 54.62 & 62.56 & 64.87 & 63.59 & 59.49 & 58.46 & 58.21 & 60.26 & 54.36 & 57.95 & 59.74 & 64.87 & 53.85 &  \increase{58.21}{4.49} &  \increase{60.11}{11.05} &  \increase{59.53}{9.04} \\
\midrule

\multicolumn{17}{l}{\textbf{\gptFourO}} \\
\english      & 68.72 & 27.95 & 74.36 & 73.33 & 62.56 & 28.46 & 71.79 & 65.64 & 38.21 & 4.10  & 53.59 & 75.64 & 5.38  & 56.35 & 47.15 & 49.98 \\
\french & 66.41 & 58.97 & 72.82 & 67.95 & 60.51 & 39.23 & 71.28 & 68.97 & 61.28 & 63.08 & 72.05 & 71.03 & 70.00 &  \increase{58.78}{2.43} &  \increase{67.61}{20.46} &  \increase{64.89}{14.91} \\
\chinese & 67.44 & 58.21 & 73.08 & 69.23 & 58.72 & 50.26 & 67.69 & 68.97 & 58.72 & 66.15 & 70.77 & 71.79 & 73.59 &  \increase{61.35}{5.00} &  \increase{67.69}{20.54} &  \increase{65.74}{15.76} \\
\japanese & 66.92 & 52.82 & 72.56 & 67.69 & 60.00 & 39.23 & 66.67 & 65.13 & 57.95 & 66.67 & 68.46 & 73.85 & 67.95 &  \increase{57.82}{1.47} &  \increase{66.07}{18.92} &  \increase{63.53}{13.55} \\
\multilingual & 66.41 & 65.13 & 72.31 & 68.72 & 62.05 & 64.62 & 68.97 & 67.95 & 65.64 & 64.10 & 71.28 & 73.85 & 67.69 &  \increase{65.26}{8.91} &  \increase{68.63}{21.48} &  \increase{67.59}{17.61} \\
\native & 71.54 & 70.26 & 76.15 & 73.33 & 65.13 & 82.82 & 71.79 & 67.95 & 67.95 & 66.15 & 76.41 & 75.64 & 72.05 &  \increase{71.86}{15.51} &  \increase{72.19}{25.04} &  \increase{72.09}{22.11} \\
\bottomrule

    \end{tabular}
    \caption{Accuracies ($\%$) of \english, \multilingual, \native all \monolingual ICL modes (\french, \chinese and \japanese) across $13$ languages of the \xlwic dataset. AVG represents the average accuracy of the language set (LRLs, HRLs or All languages). The \underline{underlined languages} in the table header are \underline{LRLs}, otherwise HRLs. The subscript indicates the performance \textcolor{ForestGreen}{increase$\uparrow$} (or \textcolor{OrangeRed}{decrease$\downarrow$}) of all other modes compared to the \english ICL mode.}
    % \vspace{-.2cm}
    \label{tab:vanilla_eval:xlwic}
\end{table*}

\begin{table*}[!htbp]
    \setlength{\tabcolsep}{2.5pt}
    \scriptsize
    \centering
    \alternaterowcolors
\begin{tabular}{l|llllllllllll|lll}
\toprule
\textbf{\xquad} &
  \multicolumn{1}{c}{\textbf{ar}} &
  \multicolumn{1}{c}{\textbf{de}} &
  \multicolumn{1}{c}{\textbf{\underline{el}}} &
  \multicolumn{1}{c}{\textbf{en}} &
  \multicolumn{1}{c}{\textbf{es}} &
  \multicolumn{1}{c}{\textbf{hi}} &
  \multicolumn{1}{c}{\textbf{\underline{ro}}} &
  \multicolumn{1}{c}{\textbf{ru}} &
  \multicolumn{1}{c}{\textbf{\underline{th}}} &
  \multicolumn{1}{c}{\textbf{tr}} &
  \multicolumn{1}{c}{\textbf{\underline{vi}}} &
  \multicolumn{1}{c|}{\textbf{zh}} &
  \multicolumn{1}{l}{\textbf{LRL Avg}} &
  \multicolumn{1}{l}{\textbf{HRL Avg}} &
  \multicolumn{1}{l}{\textbf{ALL Avg}} \\
  \midrule

\multicolumn{16}{l}{\textbf{\llamaThree}} \\
\english      & 60.90 & 75.80 & 69.10 & 86.80 & 79.10 & 69.90 & 79.20 & 62.20 & 69.90 & 72.20 & 73.90 & 77.40 & 72.02       & 73.54       & 73.03        \\
\chinese      & 65.80 & 76.30 & 68.40 & 85.50 & 79.80 & 71.00 & 79.40 & 65.60 & 75.30 & 74.30 & 75.90 & 81.30 & \increase{73.53}{1.51}  & \increase{75.56}{2.02}  & \increase{74.88}{1.85}  \\
\german       & 63.50 & 75.10 & 69.40 & 87.10 & 79.20 & 70.80 & 79.00 & 62.70 & 72.50 & 73.30 & 76.70 & 77.90 & \increase{72.93}{0.91}  & \increase{74.44}{0.90}  & \increase{73.93}{0.90}  \\
\multilingual & 65.70 & 75.80 & 71.90 & 87.40 & 80.60 & 71.30 & 80.10 & 65.90 & 73.90 & 74.70 & 76.70 & 81.00 & \increase{74.30}{2.28}  & \increase{75.98}{2.44}  & \increase{75.42}{2.39}  \\
\native       & 66.70 & 75.10 & 72.00 & 86.80 & 79.40 & 73.60 & 79.90 & 63.90 & 75.40 & 75.60 & 78.60 & 81.30 & \increase{75.22}{3.20}  & \increase{75.92}{2.38}  & \increase{75.69}{2.66}  \\
\midrule

\multicolumn{16}{l}{\textbf{\llamaThreeOne}} \\
\english      & 60.80 & 73.20 & 65.40 & 85.80 & 77.80 & 66.50 & 73.70 & 53.60 & 67.40 & 66.20 & 71.90 & 73.80 & 68.25 &         70.39 &        69.68        \\
\chinese      & 60.00 & 72.50 & 62.70 & 75.70 & 76.20 & 67.60 & 73.20 & 54.50 & 68.40 & 65.30 & 70.70 & 75.00 & \decrease{67.98}{0.27} & \decrease{68.74}{1.65} & \decrease{68.48}{1.20} \\
\german       & 59.60 & 72.80 & 66.60 & 66.80 & 78.30 & 67.40 & 75.10 & 55.70 & 67.70 & 66.90 & 73.40 & 74.50 & \increase{69.20}{0.95}  & \decrease{68.50}{1.89} & \decrease{68.73}{0.95} \\
\multilingual & 60.80 & 73.50 & 68.10 & 83.50 & 79.70 & 69.40 & 76.00 & 58.20 & 71.00 & 66.60 & 73.00 & 75.10 & \increase{71.12}{2.87}  & \increase{71.30}{0.91}  & \increase{71.24}{1.56}  \\
\native       & 61.60 & 72.80 & 67.30 & 85.80 & 77.40 & 69.30 & 76.90 & 59.00 & 71.60 & 67.70 & 75.80 & 75.00 & \increase{71.28}{3.03}  & \increase{71.89}{1.50}  & \increase{71.68}{2.00}  \\
\midrule

\multicolumn{16}{l}{\textbf{\qwenTwo}} \\
\english      & 60.40 & 69.60 & 39.10 & 80.70 & 77.20 & 50.80 & 69.60 & 54.90 & 54.10 & 61.00 & 74.40 & 80.10 & 53.40        & 69.79        & 64.32        \\
\chinese      & 58.00 & 68.40 & 36.80 & 83.10 & 74.60 & 48.80 & 67.40 & 53.80 & 56.60 & 61.40 & 72.80 & 83.70 & \decrease{52.40}{1.00} & \decrease{69.48}{0.31} & \decrease{63.78}{0.54} \\
\german       & 61.00 & 70.40 & 49.20 & 82.40 & 77.10 & 54.10 & 72.30 & 59.60 & 63.10 & 65.80 & 73.60 & 82.20 & \increase{59.67}{6.27}  & \increase{71.51}{1.72}  & \increase{67.57}{3.25}  \\
\multilingual & 60.60 & 71.50 & 48.50 & 83.10 & 78.10 & 55.00 & 72.80 & 60.80 & 64.40 & 64.90 & 74.60 & 82.10 & \increase{60.18}{6.78}  & \increase{71.96}{2.17}  & \increase{68.03}{3.71}  \\
\native       & 62.80 & 70.40 & 55.70 & 80.70 & 77.20 & 58.60 & 72.90 & 65.20 & 72.40 & 65.70 & 75.70 & 83.70 & \increase{64.90}{11.50} & \increase{72.68}{2.89}  & \increase{70.08}{5.76}  \\
\midrule

\multicolumn{16}{l}{\textbf{\qwenTwoFive}} \\
\english      & 63.70 & 74.30 & 57.00 & 85.50 & 80.60 & 62.20 & 80.20 & 67.00 & 76.50 & 69.90 & 75.10 & 85.10 & 68.98        & 75.15        & 73.09        \\
\chinese      & 57.80 & 67.30 & 54.10 & 69.60 & 67.40 & 60.90 & 77.10 & 61.30 & 76.90 & 70.10 & 65.30 & 87.20 & \decrease{67.25}{1.73} & \decrease{68.25}{6.90} & \decrease{67.92}{5.17} \\
\german       & 63.00 & 75.90 & 57.30 & 82.70 & 80.50 & 62.50 & 79.80 & 67.80 & 76.20 & 71.60 & 76.00 & 85.20 & \decrease{68.95}{0.03} & \increase{75.34}{0.19}  & \increase{73.21}{0.12}  \\
\multilingual & 63.70 & 75.50 & 58.40 & 81.20 & 80.20 & 62.30 & 80.40 & 67.90 & 77.90 & 70.80 & 76.30 & 86.20 & \increase{69.75}{0.77}  & \increase{75.23}{0.08}  & \increase{73.40}{0.31}  \\
\native       & 67.90 & 75.90 & 61.00 & 85.50 & 79.80 & 64.90 & 79.50 & 69.30 & 78.40 & 73.10 & 78.10 & 87.20 & \increase{70.95}{1.97}  & \increase{77.10}{1.95}  & \increase{75.05}{1.96}  \\
\midrule

\multicolumn{16}{l}{\textbf{\mistral}} \\
\english      & 56.50 & 70.30 & 52.90 & 83.90 & 78.70 & 51.50 & 73.60 & 56.50 & 65.70 & 63.50 & 65.90 & 51.30 & 60.93        & 65.82        & 64.19        \\
\chinese      & 60.40 & 71.80 & 61.30 & 84.30 & 78.40 & 69.90 & 73.70 & 60.80 & 70.60 & 65.60 &  75.50      &  76.2     & \increase{68.80}{7.87}  & \increase{71.63}{5.81}  & \increase{70.71}{6.52}  \\
\german       & 60.00 & 70.70 & 61.30 & 84.50 & 79.10 & 69.40 & 74.40 & 60.70 & 69.40 & 65.90 & 74.80 & 72.40 & \increase{68.62}{7.69}  & \increase{71.01}{5.19}  & \increase{70.22}{6.03}  \\
\multilingual & 61.30 & 72.30 & 62.60 & 84.80 & 79.60 & 70.00 & 75.40 & 62.40 & 72.30 & 68.00 & 75.20      & 75.20      & \increase{70.08}{9.15}  & \increase{72.35}{6.53}  & \increase{71.59}{7.40}  \\
\native       & 61.50 & 70.70 & 63.80 & 83.90 & 76.40 & 70.50 & 76.00 & 63.00 & 72.10 & 68.20 & 74.10 & 76.20 & \increase{70.60}{9.67}  & \increase{71.75}{5.93}  & \increase{71.37}{7.18}  \\
\midrule

\multicolumn{16}{l}{\textbf{\aya}} \\
\english      & 71.60 & 76.70 & 74.00 & 87.60 & 83.70 & 70.30 & 83.00 & 64.70 & 54.60 & 68.40 & 80.20 & 80.60 & 70.48        & 76.69        & 74.62        \\
\chinese      & 72.00 & 72.20 & 72.00 & 66.10 & 82.50 & 69.20 & 81.60 & 63.70 & 54.40 & 63.50 & 79.50 & 82.10 & \decrease{69.30}{1.18} & \decrease{72.70}{3.99} & \decrease{71.57}{3.05} \\
\german       & 71.60 & 74.70 & 68.40 & 63.40 & 78.20 & 69.50 & 76.70 & 50.60 & 55.10 & 59.50 & 78.20 & 68.70 & \decrease{67.42}{3.06} & \decrease{68.11}{8.58} & \decrease{67.88}{6.74} \\
\multilingual & 72.50 & 74.20 & 72.30 & 78.20 & 83.10 & 71.20 & 82.40 & 64.10 & 55.10 & 67.90 & 79.80 & 78.20 & \decrease{70.25}{0.23} & \decrease{74.75}{1.94} & \decrease{73.25}{1.37} \\
\native       & 72.00 & 74.70 & 72.00 & 87.60 & 82.80 & 67.70 & 81.30 & 64.60 & 51.50 & 67.60 & 79.60 & 82.10 & \decrease{68.12}{2.36} & \decrease{76.38}{0.31} & \decrease{73.62}{1.00} \\
\bottomrule

    \end{tabular}
    \caption{Accuracies ($\%$) of \english, \multilingual, \native all \monolingual ICL modes (\german and \chinese) across $12$ languages of the \xquad dataset. AVG represents the average accuracy of the language set (LRLs, HRLs or All languages). The \underline{underlined languages} in the table header are \underline{LRLs}, otherwise HRLs. The subscript indicates the performance \textcolor{ForestGreen}{increase$\uparrow$} (or \textcolor{OrangeRed}{decrease$\downarrow$}) of all other modes compared to the \english ICL mode.}
    % \vspace{-.2cm}
    \label{tab:vanilla_eval:xquad}
\end{table*}

\begin{table*}[!htbp]
    \setlength{\tabcolsep}{2pt}
    \scriptsize
    \centering
    \alternaterowcolors[5]
\begin{tabular}{l|llccccc|llccccc}
\toprule
\multicolumn{1}{l|}{\textbf{McNemar's Test}} & \multicolumn{7}{c|}{\textbf{Low-Resource Languages}} & \multicolumn{7}{c}{\textbf{High-Resource   Languages}} \\
\multicolumn{1}{l|}{\textbf{\mgsm ICL Mode}} &
  \multicolumn{1}{c|}{\textbf{$\chi^2$}} &
  \multicolumn{1}{c|}{\textbf{$p$-value}} &
  \multicolumn{1}{c|}{\textbf{Sig.}} &
  \multicolumn{1}{c|}{\textbf{\#Both}} &
  \multicolumn{1}{c|}{\textbf{\#M1 \faTimes}} &
  \multicolumn{1}{c|}{\textbf{\#M1 \faCheck}} &
  \multicolumn{1}{c|}{\textbf{\#Both}} &
  \multicolumn{1}{c|}{\textbf{$\chi^2$}} &
  \multicolumn{1}{c|}{\textbf{$p$-value}} &
  \multicolumn{1}{c|}{\textbf{Sig.}} &
  \multicolumn{1}{c|}{\textbf{\#Both}} &
  \multicolumn{1}{c|}{\textbf{\#M1 \faTimes}} &
  \multicolumn{1}{c|}{\textbf{\#M1 \faCheck}} &
  \multicolumn{1}{c}{\textbf{\#Both}} \\
\multicolumn{1}{l|}{\textbf{M1 vs M2 Comparison}} &
  \multicolumn{1}{c|}{\textbf{}} &
  \multicolumn{1}{c|}{\textbf{}} &
  \multicolumn{1}{c|}{\textbf{Level}} &
  \multicolumn{1}{c|}{\textbf{\faTimes}} &
  \multicolumn{1}{c|}{\textbf{M2 \faCheck}} &
  \multicolumn{1}{c|}{\textbf{M2 \faTimes}} &
  \multicolumn{1}{c|}{\textbf{\faCheck}} &
  \multicolumn{1}{c|}{\textbf{}} &
  \multicolumn{1}{c|}{\textbf{}} &
  \multicolumn{1}{c|}{\textbf{Level}} &
  \multicolumn{1}{c|}{\textbf{\faTimes}} &
  \multicolumn{1}{c|}{\textbf{M2 \faCheck}} &
  \multicolumn{1}{c|}{\textbf{M2 \faTimes}} &
  \multicolumn{1}{c}{\textbf{\faCheck}} \\
  \midrule

\multicolumn{15}{l}{\textbf{\llamaThree}}                                                                                                  \\
\english vs \french          & 0.85   & $3.56\times10^{-1}$  &      & 280  & 78   & 91  & 551 & 0.08   & $7.84\times10^{-1}$  &      & 300  & 104  & 109  & 1237  \\
\english vs \chinese         & 0.65   & $4.20\times10^{-1}$  &      & 271  & 87   & 99  & 543 & 1.09   & $2.97\times10^{-1}$  &      & 295  & 109  & 126  & 1220  \\
\bottomrule
\end{tabular}

    \caption{McNemar's test results of ICL modes on LRL and HRL splits of \mgsm dataset. Baseline is the \english mode, compared with other \monolingual, \multilingual, and \native modes.}
    % \vspace{-.2cm}
    \label{tab:hyp_test:vanilla_eval:mgsm}
\end{table*}

\begin{table*}[!htbp]
    \setlength{\tabcolsep}{1.5pt}
    \scriptsize
    \centering
    \alternaterowcolors[5]
\begin{tabular}{l|llccccc|llccccc}
\toprule
\multicolumn{1}{l|}{\textbf{McNemar's Test}} & \multicolumn{7}{c|}{\textbf{Low-Resource Languages}} & \multicolumn{7}{c}{\textbf{High-Resource   Languages}} \\
\multicolumn{1}{l|}{\textbf{\xcopa ICL Mode}} &
  \multicolumn{1}{c|}{\textbf{$\chi^2$ }} &
  \multicolumn{1}{c|}{\textbf{$p$-value}} &
  \multicolumn{1}{c|}{\textbf{Sig.}} &
  \multicolumn{1}{c|}{\textbf{\#Both}} &
  \multicolumn{1}{c|}{\textbf{\#M1 \faTimes}} &
  \multicolumn{1}{c|}{\textbf{\#M1 \faCheck}} &
  \multicolumn{1}{c|}{\textbf{\#Both}} &
  \multicolumn{1}{c|}{\textbf{$\chi^2$ }} &
  \multicolumn{1}{c|}{\textbf{$p$-value}} &
  \multicolumn{1}{c|}{\textbf{Sig.}} &
  \multicolumn{1}{c|}{\textbf{\#Both}} &
  \multicolumn{1}{c|}{\textbf{\#M1 \faTimes}} &
  \multicolumn{1}{c|}{\textbf{\#M1 \faCheck}} &
  \multicolumn{1}{c}{\textbf{\#Both}} \\
\multicolumn{1}{l|}{\textbf{M1 vs M2 Comparison}} &
  \multicolumn{1}{c|}{\textbf{}} &
  \multicolumn{1}{c|}{\textbf{}} &
  \multicolumn{1}{c|}{\textbf{Level}} &
  \multicolumn{1}{c|}{\textbf{\faTimes}} &
  \multicolumn{1}{c|}{\textbf{M2 \faCheck}} &
  \multicolumn{1}{c|}{\textbf{M2 \faTimes}} &
  \multicolumn{1}{c|}{\textbf{\faCheck}} &
  \multicolumn{1}{c|}{\textbf{}} &
  \multicolumn{1}{c|}{\textbf{}} &
  \multicolumn{1}{c|}{\textbf{Level}} &
  \multicolumn{1}{c|}{\textbf{\faTimes}} &
  \multicolumn{1}{c|}{\textbf{M2 \faCheck}} &
  \multicolumn{1}{c|}{\textbf{M2 \faTimes}} &
  \multicolumn{1}{c}{\textbf{\faCheck}} \\
  \midrule

\multicolumn{15}{l}{\textbf{\llamaThree}}                                                                                                   \\
\english vs \italian         & 274.27  & $1.33\times10^{-61}$  & *** & 1246 & 636  & 166 & 1452 & 8.84    & $2.95\times10^{-3}$  & **   & 287  & 111  & 70   & 2032 \\
\english vs \chinese         & 347.11  & $1.80\times10^{-77}$  & *** & 1177 & 705  & 157 & 1461 & 5.85    & $1.55\times10^{-2}$  & *    & 288  & 110  & 76   & 2026 \\
\bottomrule
\end{tabular}

    \caption{McNemar's test results of ICL modes on LRL and HRL splits of \xcopa dataset. Baseline is the \english mode, compared with other \monolingual, \multilingual, and \native modes.}
    % \vspace{-.2cm}
    \label{tab:hyp_test:vanilla_eval:xcopa}
\end{table*}

\begin{table*}[!htbp]
    \setlength{\tabcolsep}{2pt}
    \scriptsize
    \centering
    \alternaterowcolors[5]
\begin{tabular}{l|llccccc|llccccc}
\toprule
\multicolumn{1}{l|}{\textbf{McNemar's Test}} & \multicolumn{7}{c|}{\textbf{Low-Resource Languages}} & \multicolumn{7}{c}{\textbf{High-Resource   Languages}} \\
\multicolumn{1}{l|}{\textbf{\xlwic ICL Mode}} &
  \multicolumn{1}{c|}{\textbf{$\chi^2$}} &
  \multicolumn{1}{c|}{\textbf{$p$-value}} &
  \multicolumn{1}{c|}{\textbf{Sig.}} &
  \multicolumn{1}{c|}{\textbf{\#Both}} &
  \multicolumn{1}{c|}{\textbf{\#M1 \faTimes}} &
  \multicolumn{1}{c|}{\textbf{\#M1 \faCheck}} &
  \multicolumn{1}{c|}{\textbf{\#Both}} &
  \multicolumn{1}{c|}{\textbf{$\chi^2$ }} &
  \multicolumn{1}{c|}{\textbf{$p$-value}} &
  \multicolumn{1}{c|}{\textbf{Sig.}} &
  \multicolumn{1}{c|}{\textbf{\#Both}} &
  \multicolumn{1}{c|}{\textbf{\#M1 \faTimes}} &
  \multicolumn{1}{c|}{\textbf{\#M1 \faCheck}} &
  \multicolumn{1}{c}{\textbf{\#Both}} \\
\multicolumn{1}{l|}{\textbf{M1 vs M2 Comparison}} &
  \multicolumn{1}{c|}{\textbf{}} &
  \multicolumn{1}{c|}{\textbf{}} &
  \multicolumn{1}{c|}{\textbf{Level}} &
  \multicolumn{1}{c|}{\textbf{\faTimes}} &
  \multicolumn{1}{c|}{\textbf{M2 \faCheck}} &
  \multicolumn{1}{c|}{\textbf{M2 \faTimes}} &
  \multicolumn{1}{c|}{\textbf{\faCheck}} &
  \multicolumn{1}{c|}{\textbf{}} &
  \multicolumn{1}{c|}{\textbf{}} &
  \multicolumn{1}{c|}{\textbf{Level}} &
  \multicolumn{1}{c|}{\textbf{\faTimes}} &
  \multicolumn{1}{c|}{\textbf{M2 \faCheck}} &
  \multicolumn{1}{c|}{\textbf{M2 \faTimes}} &
  \multicolumn{1}{c}{\textbf{\faCheck}} \\
  \midrule

\multicolumn{15}{l}{\textbf{\llamaThree}}                                                                                                  \\
\english vs \french          & 1.85    & $1.74\times10^{-1}$  &      & 483  & 182 & 156 & 739 & 3.70    & $5.45\times10^{-2}$   &      & 991  & 422  & 367 & 1730 \\
\english vs \chinese         & 0.00    & $9.60\times10^{-1}$  &      & 469  & 196 & 194 & 701 & 1.38    & $2.41\times10^{-1}$   &      & 1000 & 413  & 379 & 1718 \\
\bottomrule
\end{tabular}

    \caption{McNemar's test results of ICL modes on LRL and HRL splits of \xlwic dataset. Baseline is the \english mode, compared with other \monolingual, \multilingual, and \native modes.}
    % \vspace{-.2cm}
    \label{tab:hyp_test:vanilla_eval:xlwic}
\end{table*}

\begin{table*}[!htbp]
    \setlength{\tabcolsep}{2pt}
    \scriptsize
    \centering
    \alternaterowcolors[5]
\begin{tabular}{l|llccccc|llccccc}
\toprule
\multicolumn{1}{l|}{\textbf{McNemar's Test}} & \multicolumn{7}{c|}{\textbf{Low-Resource Languages}} & \multicolumn{7}{c}{\textbf{High-Resource   Languages}} \\
\multicolumn{1}{l|}{\textbf{\xquad ICL Mode}} &
  \multicolumn{1}{c|}{\textbf{$\chi^2$}} &
  \multicolumn{1}{c|}{\textbf{$p$-value}} &
  \multicolumn{1}{c|}{\textbf{Sig.}} &
  \multicolumn{1}{c|}{\textbf{\#Both}} &
  \multicolumn{1}{c|}{\textbf{\#M1 \faTimes}} &
  \multicolumn{1}{c|}{\textbf{\#M1 \faCheck}} &
  \multicolumn{1}{c|}{\textbf{\#Both}} &
  \multicolumn{1}{c|}{\textbf{$\chi^2$ }} &
  \multicolumn{1}{c|}{\textbf{$p$-value}} &
  \multicolumn{1}{c|}{\textbf{Sig.}} &
  \multicolumn{1}{c|}{\textbf{\#Both}} &
  \multicolumn{1}{c|}{\textbf{\#M1 \faTimes}} &
  \multicolumn{1}{c|}{\textbf{\#M1 \faCheck}} &
  \multicolumn{1}{c}{\textbf{\#Both}} \\
\multicolumn{1}{l|}{\textbf{M1 vs M2 Comparison}} &
  \multicolumn{1}{c|}{\textbf{}} &
  \multicolumn{1}{c|}{\textbf{}} &
  \multicolumn{1}{c|}{\textbf{Level}} &
  \multicolumn{1}{c|}{\textbf{\faTimes}} &
  \multicolumn{1}{c|}{\textbf{M2 \faCheck}} &
  \multicolumn{1}{c|}{\textbf{M2 \faTimes}} &
  \multicolumn{1}{c|}{\textbf{\faCheck}} &
  \multicolumn{1}{c|}{\textbf{}} &
  \multicolumn{1}{c|}{\textbf{}} &
  \multicolumn{1}{c|}{\textbf{Level}} &
  \multicolumn{1}{c|}{\textbf{\faTimes}} &
  \multicolumn{1}{c|}{\textbf{M2 \faCheck}} &
  \multicolumn{1}{c|}{\textbf{M2 \faTimes}} &
  \multicolumn{1}{c}{\textbf{\faCheck}} \\
  \midrule

\multicolumn{15}{l}{\textbf{\llamaThree}}                                                                                                 \\
\english vs \german         & 5.02    & $2.50\times10^{-2}$  & *   & 979  & 140 & 104 & 2777 & 10.55   & $1.16\times10^{-3}$   & **   & 1842  & 275 & 203 & 5680 \\
\english vs \chinese        & 10.61   & $1.12\times10^{-3}$  & **  & 925  & 194 & 134 & 2747 & 42.92   & $5.72\times10^{-11}$   & ***  & 1734  & 383 & 221 & 5662 \\
\bottomrule
\end{tabular}

    \caption{McNemar's test results of ICL modes on LRL and HRL splits of \xquad dataset. Baseline is the \english mode, compared with other \monolingual, \multilingual, and \native modes.}
    % \vspace{-.2cm}
    \label{tab:hyp_test:vanilla_eval:xquad}
\end{table*}

\begin{table*}[!htbp]
    \setlength{\tabcolsep}{4pt}
    \scriptsize
    \centering
    \alternaterowcolors
\begin{tabular}{l|ccccccccccc|lll}
\toprule
    \textbf{\mgsm \cis} &
  \textbf{\underline{bn}} &
  \textbf{de} &
  \textbf{en} &
  \textbf{es} &
  \textbf{fr} &
  \textbf{ja} &
  \textbf{ru} &
  \textbf{\underline{sw}} &
  \textbf{\underline{te}} &
  \textbf{\underline{th}} &
  \textbf{zh} &
  \textbf{\underline{LRL AVG}} &
  \textbf{HRL AVG} &
  \textbf{ALL AVG} \\ 
  \midrule

\multicolumn{15}{l}{\textbf{\llamaThree}} \\
\english$+\ $\cisEn           & 64.40 & 75.20 & 84.80 & 78.40 & 74.00 & 69.20 & 74.00 & 56.80 & 49.20 & 73.60 & 72.40 & 61.00 & 75.43 & 70.18 \\
\english$+\ $\cisFr           & 66.00 & 77.20 & 85.20 & 80.40 & 76.40 & 67.20 & 75.60 & 54.00 & 58.40 & 68.00 & 70.80 & 61.60 & 76.11 & 70.84 \\
\english$+\ $\cisJa          & 66.40 & 75.60 & 83.20 & 78.80 & 74.80 & 66.80 & 77.20 & 53.20 & 53.60 & 72.00 & 70.80 & 61.30 & 75.31 & 70.22 \\
\english$+\ $\cisZh           & 63.60 & 76.80 & 83.20 & 77.20 & 76.40 & 69.20 & 77.60 & 52.40 & 52.40 & 68.80 & 71.60 & 59.30 & 76.00 & 69.93 \\
\english$+\ $\cisMulti & 65.60 & 75.20 & 83.20 & 78.40 & 77.20 & 66.40 & 76.40 & 51.60 & 55.60 & 69.20 & 70.40 & 60.50 & 75.31 & 69.93 \\
\midrule

\multicolumn{15}{l}{\textbf{\llamaThreeOne}}                                                                                          \\
\bottomrule

    \end{tabular}
    \caption{Accuracies ($\%$) of CIS modes across $11$ languages of the \mgsm dataset. AVG represents the average accuracy of the language set (LRLs, HRLs or All languages). The \underline{underlined languages} in the table header are \underline{LRLs}, otherwise HRLs.}
    % \vspace{-.2cm}
    \label{tab:cis:mgsm}
\end{table*}

\begin{table*}[!htbp]
    \setlength{\tabcolsep}{3pt}
    \scriptsize
    \centering
    \alternaterowcolors
\begin{tabular}{l|cccccccccccc|lll}
\toprule
\textbf{\xcopa \cis} &
  \multicolumn{1}{c}{\textbf{en}} &
  \multicolumn{1}{c}{{\textbf{\underline{et}}}} &
  \multicolumn{1}{c}{{\textbf{\underline{ht}}}} &
  \multicolumn{1}{c}{\textbf{id}} &
  \multicolumn{1}{c}{\textbf{it}} &
  \multicolumn{1}{c}{{\textbf{\underline{qu}}}} &
  \multicolumn{1}{c}{{\textbf{\underline{sw}}}} &
  \multicolumn{1}{c}{{\textbf{\underline{ta}}}} &
  \multicolumn{1}{c}{{\textbf{\underline{th}}}} &
  \multicolumn{1}{c}{\textbf{tr}} &
  \multicolumn{1}{c}{\textbf{\underline{vi}}} &
  \textbf{zh} &
  \textbf{\underline{LRL AVG}} &
  \textbf{HRL AVG} &
  \textbf{ALL AVG} \\
\midrule

\multicolumn{16}{l}{\textbf{\llamaThree}}                                                                                                             \\
\english$+\ $\cisEn           & 95.40 & 56.60 & 16.00 & 80.20 & 84.80 & 18.60 & 39.00 & 58.40 & 69.80 & 70.80 & 80.20 & 87.40 & 48.37 & 83.72 & 63.10 \\
\english$+\ $\cisFr           & 95.40 & 57.00 & 47.00 & 80.40 & 85.80 & 42.20 & 52.40 & 57.80 & 72.40 & 74.20 & 80.60 & 86.60 & 58.49 & 84.48 & 69.32 \\
\english$+\ $\cisJa          & 95.00 & 57.20 & 39.60 & 82.20 & 86.60 & 46.40 & 56.00 & 57.00 & 73.20 & 74.00 & 81.40 & 85.60 & 58.69 & 84.68 & 69.52 \\
\english$+\ $\cisZh           & 95.00 & 58.40 & 35.80 & 82.00 & 86.00 & 43.00 & 56.80 & 57.00 & 73.80 & 73.60 & 81.40 & 87.60 & 58.03 & 84.84 & 69.20 \\
\english$+\ $\cisMulti & 95.60 & 58.40 & 51.00 & 81.60 & 86.80 & 49.00 & 59.20 & 56.40 & 73.60 & 74.60 & 80.40 & 88.00 & 61.14 & 85.32 & 71.22 \\
\midrule

\multicolumn{16}{l}{\textbf{\llamaThreeOne}}                                                                                                          \\
\bottomrule

    \end{tabular}
    \caption{Accuracies ($\%$) of CIS modes across $12$ languages of the \xcopa dataset. AVG represents the average accuracy of the language set (LRLs, HRLs or All languages). The \underline{underlined languages} in the table header are \underline{LRLs}, otherwise HRLs. }
    % \vspace{-.2cm}
    \label{tab:cis:xcopa}
\end{table*}

\begin{table*}[!htbp]
    \setlength{\tabcolsep}{2.5pt}
    \scriptsize
    \centering
    \alternaterowcolors
\begin{tabular}{l|ccccccccccccc|lll}
\toprule
\textbf{\xlwic \cis} &
  \multicolumn{1}{c}{\textbf{\underline{bg}}} &
  \multicolumn{1}{c}{\textbf{da}} &
  \multicolumn{1}{c}{\textbf{de}} &
  \multicolumn{1}{c}{\textbf{en}} &
  \multicolumn{1}{c}{\textbf{\underline{et}}} &
  \multicolumn{1}{c}{\textbf{\underline{fa}}} &
  \multicolumn{1}{c}{\textbf{fr}} &
  \multicolumn{1}{c}{\textbf{\underline{hr}}} &
  \multicolumn{1}{c}{\textbf{it}} &
  \multicolumn{1}{c}{\textbf{ja}} &
  \multicolumn{1}{c}{\textbf{ko}} &
  \multicolumn{1}{c}{\textbf{nl}} &
  \textbf{zh} &
  \textbf{\underline{LRL AVG}} &
  \textbf{HRL AVG} &
  \textbf{ALL AVG} \\
  \midrule

\multicolumn{17}{l}{\textbf{\llamaThree}}                                                                                                                             \\
\english$+\ $\cisEn           & 55.13 & 65.90 & 61.54 & 65.64 & 53.85 & 64.36 & 59.23 & 53.59 & 52.05 & 52.82 & 54.62 & 56.92 & 63.59 & 56.73 & 59.15 & 58.40 \\
\english$+\ $\cisFr           & 56.15 & 67.18 & 63.08 & 64.62 & 54.62 & 65.13 & 60.00 & 54.10 & 53.85 & 52.82 & 55.64 & 60.00 & 60.26 & 57.50 & 59.72 & 59.03 \\
\english$+\ $\cisJa          & 56.67 & 67.95 & 62.05 & 64.87 & 54.10 & 66.67 & 60.77 & 55.13 & 56.67 & 53.85 & 53.85 & 59.23 & 62.05 & 58.14 & 60.14 & 59.53 \\
\english$+\ $\cisZh           & 57.44 & 68.72 & 63.33 & 66.15 & 54.36 & 65.64 & 58.97 & 55.64 & 54.10 & 53.33 & 54.87 & 60.26 & 58.97 & 58.27 & 59.86 & 59.37 \\
\english$+\ $\cisMulti & 56.67 & 65.90 & 62.31 & 64.36 & 54.62 & 65.90 & 60.77 & 54.10 & 54.62 & 51.79 & 52.31 & 59.49 & 61.03 & 57.82 & 59.17 & 58.76 \\
\midrule

\multicolumn{17}{l}{\textbf{\llamaThreeOne}}                                                                                                                          \\
\bottomrule

    \end{tabular}
    \caption{Accuracies ($\%$) of CIS modes across $13$ languages of the \xlwic dataset. AVG represents the average accuracy of the language set (LRLs, HRLs or All languages). The \underline{underlined languages} in the table header are \underline{LRLs}, otherwise HRLs. }
    % \vspace{-.2cm}
    \label{tab:cis:xlwic}
\end{table*}

\begin{table*}[!htbp]
    \setlength{\tabcolsep}{2.5pt}
    \scriptsize
    \centering
    \alternaterowcolors
\begin{tabular}{l|cccccccccccc|lll}
\toprule
\textbf{\xquad \cis} &
  \textbf{ar} &
  \textbf{de} &
  \underline{ \textbf{el}} &
  \textbf{en} &
  \textbf{es} &
  \underline{ \textbf{hi}} &
  \underline{ \textbf{ro}} &
  \textbf{ru} &
  \underline{\textbf{th}} &
  \textbf{tr} &
  \textbf{vi} &
  \textbf{zh} &
  \underline{\textbf{LRL AVG}} &
  \textbf{HRL AVG} &
  \textbf{ALL AVG} \\
  \midrule

\multicolumn{16}{l}{\textbf{\llamaThree}}                                                                                                      \\
\english$+\ $\cisEn    & 60.70 & 75.00 & 68.30 & 86.70 & 79.30 & 68.70 & 79.10 & 62.20 & 69.60 & 72.40 & 74.40 & 77.80 & 71.43 & 73.56 & 72.85 \\
\english$+\ $\cisFr    & 58.80 & 75.30 & 68.30 & 86.70 & 79.50 & 68.00 & 79.00 & 62.00 & 68.60 & 71.30 & 73.60 & 76.70 & 70.98 & 72.99 & 72.32 \\
\english$+\ $\cisJa    & 58.60 & 75.00 & 67.70 & 86.80 & 79.70 & 66.90 & 79.20 & 61.70 & 69.60 & 71.50 & 73.00 & 77.80 & 70.85 & 73.01 & 72.29 \\
\english$+\ $\cisZh    & 58.30 & 75.30 & 68.40 & 87.00 & 79.40 & 68.20 & 78.80 & 61.90 & 69.70 & 71.40 & 73.20 & 77.90 & 71.28 & 73.05 & 72.46 \\
\english$+\ $\cisMulti & 58.50 & 75.30 & 68.70 & 87.30 & 79.80 & 66.40 & 79.20 & 62.30 & 69.30 & 71.60 & 74.10 & 76.70 & 70.90 & 73.20 & 72.43 \\

\midrule

\multicolumn{16}{l}{\textbf{\llamaThreeOne}}                                                                                                   \\
\bottomrule

    \end{tabular}
    \caption{Accuracies ($\%$) of \cis modes across $12$ languages of the \xquad dataset. AVG represents the average accuracy of the language set (LRLs, HRLs or All languages). The \underline{underlined languages} in the table header are \underline{LRLs}, otherwise HRLs. }
    % \vspace{-.2cm}
    \label{tab:cis:xquad}
\end{table*}

\begin{table*}[!htbp]
    \setlength{\tabcolsep}{0.5pt}
    \scriptsize
    \centering
    \alternaterowcolors[5]
\begin{tabular}{l|llccccc|llccccc}
\toprule
\multicolumn{1}{l|}{\textbf{McNemar's Test}} & \multicolumn{7}{c|}{\textbf{Low-Resource Languages}} & \multicolumn{7}{c}{\textbf{High-Resource Languages}} \\
\multicolumn{1}{l|}{\textbf{\mgsm \cis Mode}} &
  \multicolumn{1}{c|}{\textbf{$\chi^2$}} &
  \multicolumn{1}{c|}{\textbf{$p$-value}} &
  \multicolumn{1}{c|}{\textbf{Sig.}} &
  \multicolumn{1}{c|}{\textbf{\#Both}} &
  \multicolumn{1}{c|}{\textbf{\#M1 \faTimes}} &
  \multicolumn{1}{c|}{\textbf{\#M1 \faCheck}} &
  \multicolumn{1}{c|}{\textbf{\#Both}} &
  \multicolumn{1}{c|}{\textbf{$\chi^2$}} &
  \multicolumn{1}{c|}{\textbf{$p$-value}} &
  \multicolumn{1}{c|}{\textbf{Sig.}} &
  \multicolumn{1}{c|}{\textbf{\#Both}} &
  \multicolumn{1}{c|}{\textbf{\#M1 \faTimes}} &
  \multicolumn{1}{c|}{\textbf{\#M1 \faCheck}} &
  \multicolumn{1}{c}{\textbf{\#Both}} \\
\multicolumn{1}{l|}{\textbf{M1 vs M2 Comparison}} &
  \multicolumn{1}{c|}{\textbf{}} &
  \multicolumn{1}{c|}{\textbf{}} &
  \multicolumn{1}{c|}{\textbf{Level}} &
  \multicolumn{1}{c|}{\textbf{\faTimes}} &
  \multicolumn{1}{c|}{\textbf{M2 \faCheck}} &
  \multicolumn{1}{c|}{\textbf{M2 \faTimes}} &
  \multicolumn{1}{c|}{\textbf{\faCheck}} &
  \multicolumn{1}{c|}{\textbf{}} &
  \multicolumn{1}{c|}{\textbf{}} &
  \multicolumn{1}{c|}{\textbf{Level}} &
  \multicolumn{1}{c|}{\textbf{\faTimes}} &
  \multicolumn{1}{c|}{\textbf{M2 \faCheck}} &
  \multicolumn{1}{c|}{\textbf{M2 \faTimes}} &
  \multicolumn{1}{c}{\textbf{\faCheck}} \\
  \midrule

\multicolumn{15}{l}{\textbf{\llamaThree}}                                                                                                                         \\
\tiny{\english$+\ $\cisEn vs   \english$+\ $\cisZh}            & 1.57  & 2.10E-01 &     & 317 & 73  & 90  & 520 & 0.47  & 4.93E-01 &     & 339 & 91  & 81  & 1239 \\
\tiny{\english$+\ $\cisEn vs   \english$+\ $\cisFr}            & 0.16  & 6.85E-01 &     & 311 & 79  & 73  & 537 & 0.60  & 4.39E-01 &     & 323 & 107 & 95  & 1225 \\
\bottomrule
\end{tabular}

    \caption{McNemar's test results of \cis modes comparison on LRL and HRL splits of \mgsm dataset across $6$ MLLMs. Sig. -- Significance. \faTimes -- the model response is incorrect; \faCheck -- correct.}
    % \vspace{-.2cm}
    \label{tab:hyp_test:cis:mgsm}
\end{table*}

\begin{table*}[!htbp]
    \setlength{\tabcolsep}{0.5pt}
    \scriptsize
    \centering
    \alternaterowcolors[5]
\begin{tabular}{l|llccccc|llccccc}
\toprule
\multicolumn{1}{l|}{\textbf{McNemar's Test}} & \multicolumn{7}{c|}{\textbf{Low-Resource Languages}} & \multicolumn{7}{c}{\textbf{High-Resource Languages}} \\
\multicolumn{1}{l|}{\textbf{\xcopa \cis Mode}} &
  \multicolumn{1}{c|}{\textbf{$\chi^2$}} &
  \multicolumn{1}{c|}{\textbf{$p$-value}} &
  \multicolumn{1}{c|}{\textbf{Sig.}} &
  \multicolumn{1}{c|}{\textbf{\#Both}} &
  \multicolumn{1}{c|}{\textbf{\#M1 \faTimes}} &
  \multicolumn{1}{c|}{\textbf{\#M1 \faCheck}} &
  \multicolumn{1}{c|}{\textbf{\#Both}} &
  \multicolumn{1}{c|}{\textbf{$\chi^2$}} &
  \multicolumn{1}{c|}{\textbf{$p$-value}} &
  \multicolumn{1}{c|}{\textbf{Sig.}} &
  \multicolumn{1}{c|}{\textbf{\#Both}} &
  \multicolumn{1}{c|}{\textbf{\#M1 \faTimes}} &
  \multicolumn{1}{c|}{\textbf{\#M1 \faCheck}} &
  \multicolumn{1}{c}{\textbf{\#Both}} \\
\multicolumn{1}{l|}{\textbf{M1 vs M2 Comparison}} &
  \multicolumn{1}{c|}{\textbf{}} &
  \multicolumn{1}{c|}{\textbf{}} &
  \multicolumn{1}{c|}{\textbf{Level}} &
  \multicolumn{1}{c|}{\textbf{\faTimes}} &
  \multicolumn{1}{c|}{\textbf{M2 \faCheck}} &
  \multicolumn{1}{c|}{\textbf{M2 \faTimes}} &
  \multicolumn{1}{c|}{\textbf{\faCheck}} &
  \multicolumn{1}{c|}{\textbf{}} &
  \multicolumn{1}{c|}{\textbf{}} &
  \multicolumn{1}{c|}{\textbf{Level}} &
  \multicolumn{1}{c|}{\textbf{\faTimes}} &
  \multicolumn{1}{c|}{\textbf{M2 \faCheck}} &
  \multicolumn{1}{c|}{\textbf{M2 \faTimes}} &
  \multicolumn{1}{c}{\textbf{\faCheck}} \\
  \midrule

\multicolumn{15}{l}{\textbf{\llamaThree}}                                                                                                                           \\
\tiny{\english$+\ $\cisEn vs   \english$+\ $\cisZh}            & 199.24 & 3.05E-45 & *** & 1353 & 454 & 116 & 1577 & 8.28  & 4.00E-03 & **  & 349 & 58  & 30 & 2063 \\
\tiny{\english$+\ $\cisEn vs   \english$+\ $\cisFr}            & 214.84 & 1.21E-48 & *** & 1340 & 467 & 113 & 1580 & 3.34  & 6.76E-02 &     & 349 & 58  & 39 & 2054 \\
\bottomrule
\end{tabular}

    \caption{McNemar's test results of \cis modes comparison on LRL and HRL splits of \xcopa dataset across $6$ MLLMs. Sig. -- Significance. \faTimes -- the model response is incorrect; \faCheck -- correct.}
    % \vspace{-.2cm}
    \label{tab:hyp_test:cis:xcopa}
\end{table*}

\begin{table*}[!htbp]
    \setlength{\tabcolsep}{0.5pt}
    \scriptsize
    \centering
    \alternaterowcolors[5]
\begin{tabular}{l|llccccc|llccccc}
\toprule
\multicolumn{1}{l|}{\textbf{McNemar's Test}} & \multicolumn{7}{c|}{\textbf{Low-Resource Languages}} & \multicolumn{7}{c}{\textbf{High-Resource Languages}} \\
\multicolumn{1}{l|}{\textbf{\xlwic \cis Mode}} &
  \multicolumn{1}{c|}{\textbf{$\chi^2$}} &
  \multicolumn{1}{c|}{\textbf{$p$-value}} &
  \multicolumn{1}{c|}{\textbf{Sig.}} &
  \multicolumn{1}{c|}{\textbf{\#Both}} &
  \multicolumn{1}{c|}{\textbf{\#M1 \faTimes}} &
  \multicolumn{1}{c|}{\textbf{\#M1 \faCheck}} &
  \multicolumn{1}{c|}{\textbf{\#Both}} &
  \multicolumn{1}{c|}{\textbf{$\chi^2$}} &
  \multicolumn{1}{c|}{\textbf{$p$-value}} &
  \multicolumn{1}{c|}{\textbf{Sig.}} &
  \multicolumn{1}{c|}{\textbf{\#Both}} &
  \multicolumn{1}{c|}{\textbf{\#M1 \faTimes}} &
  \multicolumn{1}{c|}{\textbf{\#M1 \faCheck}} &
  \multicolumn{1}{c}{\textbf{\#Both}} \\
\multicolumn{1}{l|}{\textbf{M1 vs M2 Comparison}} &
  \multicolumn{1}{c|}{\textbf{}} &
  \multicolumn{1}{c|}{\textbf{}} &
  \multicolumn{1}{c|}{\textbf{Level}} &
  \multicolumn{1}{c|}{\textbf{\faTimes}} &
  \multicolumn{1}{c|}{\textbf{M2 \faCheck}} &
  \multicolumn{1}{c|}{\textbf{M2 \faTimes}} &
  \multicolumn{1}{c|}{\textbf{\faCheck}} &
  \multicolumn{1}{c|}{\textbf{}} &
  \multicolumn{1}{c|}{\textbf{}} &
  \multicolumn{1}{c|}{\textbf{Level}} &
  \multicolumn{1}{c|}{\textbf{\faTimes}} &
  \multicolumn{1}{c|}{\textbf{M2 \faCheck}} &
  \multicolumn{1}{c|}{\textbf{M2 \faTimes}} &
  \multicolumn{1}{c}{\textbf{\faCheck}} \\
  \midrule

  \multicolumn{15}{l}{\textbf{\llamaThree}}                                                                                                                           \\
\tiny{\english$+\ $\cisEn vs   \english$+\ $\cisZh}            & 3.31  & 6.90E-02 &     & 583 & 92  & 68  & 817 & 1.69   & 1.94E-01 &     & 1251 & 183 & 158 & 1918 \\
\tiny{\english$+\ $\cisEn vs   \english$+\ $\cisFr}            & 0.81  & 3.69E-01 &     & 594 & 81  & 69  & 816 & 1.16   & 2.81E-01 &     & 1269 & 165 & 145 & 1931 \\
\bottomrule
\end{tabular}

    \caption{McNemar's test results of \cis modes comparison on LRL and HRL splits of \xlwic dataset across $6$ MLLMs. Sig. -- Significance. \faTimes -- the model response is incorrect; \faCheck -- correct.}
    % \vspace{-.2cm}
    \label{tab:hyp_test:cis:xlwic}
\end{table*}

\begin{table*}[!htbp]
    \setlength{\tabcolsep}{0.5pt}
    \scriptsize
    \centering
    \alternaterowcolors[5]
\begin{tabular}{l|llccccc|llccccc}
\toprule
\multicolumn{1}{l|}{\textbf{McNemar's Test}} & \multicolumn{7}{c|}{\textbf{Low-Resource Languages}} & \multicolumn{7}{c}{\textbf{High-Resource Languages}} \\
\multicolumn{1}{l|}{\textbf{\xquad \cis Mode}} &
  \multicolumn{1}{c|}{\textbf{$\chi^2$}} &
  \multicolumn{1}{c|}{\textbf{$p$-value}} &
  \multicolumn{1}{c|}{\textbf{Sig.}} &
  \multicolumn{1}{c|}{\textbf{\#Both}} &
  \multicolumn{1}{c|}{\textbf{\#M1 \faTimes}} &
  \multicolumn{1}{c|}{\textbf{\#M1 \faCheck}} &
  \multicolumn{1}{c|}{\textbf{\#Both}} &
  \multicolumn{1}{c|}{\textbf{$\chi^2$}} &
  \multicolumn{1}{c|}{\textbf{$p$-value}} &
  \multicolumn{1}{c|}{\textbf{Sig.}} &
  \multicolumn{1}{c|}{\textbf{\#Both}} &
  \multicolumn{1}{c|}{\textbf{\#M1 \faTimes}} &
  \multicolumn{1}{c|}{\textbf{\#M1 \faCheck}} &
  \multicolumn{1}{c}{\textbf{\#Both}} \\
\multicolumn{1}{l|}{\textbf{M1 vs M2 Comparison}} &
  \multicolumn{1}{c|}{\textbf{}} &
  \multicolumn{1}{c|}{\textbf{}} &
  \multicolumn{1}{c|}{\textbf{Level}} &
  \multicolumn{1}{c|}{\textbf{\faTimes}} &
  \multicolumn{1}{c|}{\textbf{M2 \faCheck}} &
  \multicolumn{1}{c|}{\textbf{M2 \faTimes}} &
  \multicolumn{1}{c|}{\textbf{\faCheck}} &
  \multicolumn{1}{c|}{\textbf{}} &
  \multicolumn{1}{c|}{\textbf{}} &
  \multicolumn{1}{c|}{\textbf{Level}} &
  \multicolumn{1}{c|}{\textbf{\faTimes}} &
  \multicolumn{1}{c|}{\textbf{M2 \faCheck}} &
  \multicolumn{1}{c|}{\textbf{M2 \faTimes}} &
  \multicolumn{1}{c}{\textbf{\faCheck}} \\
  \midrule

\multicolumn{15}{l}{\textbf{\llamaThree}}                                                                                                                               \\
\tiny{\english$+\ $\cisEn vs   \english$+\ $\cisZh}            & 0.19   & 6.66E-01 &     & 1079 & 64  & 70  & 2787 & 6.53   & 1.06E-02  & *   & 2013 & 102 & 143 & 5742 \\
\tiny{\english$+\ $\cisEn vs   \english$+\ $\cisFr}            & 2.37   & 1.24E-01 &     & 1091 & 52  & 70  & 2787 & 8.80   & 3.01E-03  & **  & 2023 & 92  & 138 & 5747 \\
\bottomrule
\end{tabular}

    \caption{McNemar's test results of \cis modes comparison on LRL and HRL splits of \xquad dataset across $6$ MLLMs. Sig. -- Significance. \faTimes -- the model response is incorrect; \faCheck -- correct.}
    % \vspace{-.2cm}
    \label{tab:hyp_test:cis:xquad}
\end{table*}

\end{document}